\documentclass[10pt,journal,compsoc]{IEEEtran}
\usepackage{epsfig}
\usepackage{graphicx}
\usepackage{amsmath}
\usepackage{amssymb}

\usepackage{mathtools}
\usepackage{multirow}
\usepackage{booktabs} 

\usepackage{color}

\usepackage{algorithm}
\usepackage{algpseudocode}

\usepackage{url}
\usepackage[dvipsnames]{xcolor}
\usepackage[pagebackref=true,breaklinks=true,letterpaper=true,colorlinks,bookmarks=false]{hyperref}
\usepackage{array}
\usepackage{subfigure}
\usepackage{multirow}
\usepackage{tabularx}
\usepackage{makecell}
\usepackage{adjustbox}

\newcolumntype{L}[1]{>{\raggedright\arraybackslash}p{#1}}
\newcolumntype{C}[1]{>{\centering\arraybackslash}p{#1}}
\newcolumntype{R}[1]{>{\raggedleft\arraybackslash}p{#1}}

\newcommand{\widthscalefive}{0.16}
\newcommand{\etal}{\textit{et al}.}
\newcommand{\ie}{\textit{i}.\textit{e}., }

\ifCLASSOPTIONcompsoc
  \usepackage[nocompress]{cite}
\else
  \usepackage{cite}
\fi
\ifCLASSINFOpdf
\else
\fi
\begin{document}
%
\title{Structure-Preserving Image Super-Resolution}
%
%
%
%

\author{Cheng~Ma,
        Yongming~Rao, 
        Jiwen~Lu\IEEEauthorrefmark{1},~\IEEEmembership{Senior Member,~IEEE,}
        and~Jie~Zhou\IEEEauthorrefmark{1},~\IEEEmembership{Senior Member,~IEEE}
\IEEEcompsocitemizethanks{\IEEEcompsocthanksitem
* Corresponding authors.

The authors are with Beijing National Research Center for Information Science and Technology (BNRist) and the Department of Automation, Tsinghua University, Beijing, 100084, China. Email:
macheng17@mails.tsinghua.edu.cn; raoyongmimg95@gmail.com; lujiwen@tsinghua.edu.cn; jzhou@tsinghua.edu.cn. 
}
}

%
%

\markboth{IEEE TRANSACTIONS ON PATTERN ANALYSIS AND MACHINE INTELLIGENCE}%
{Shell \MakeLowercase{\textit{et al.}}: Bare Demo of IEEEtran.cls for Computer Society Journals}
%



\IEEEtitleabstractindextext{%
\begin{abstract}
Structures matter in single image super-resolution (SISR).
Benefiting from generative adversarial networks (GANs), recent studies have promoted the development of SISR by recovering photo-realistic images. However, there are still undesired structural distortions in the recovered images. 
In this paper, we propose a structure-preserving super-resolution (SPSR) method to alleviate the above issue while maintaining the merits of GAN-based methods to generate perceptual-pleasant details. Firstly, we propose SPSR with gradient guidance (SPSR-G) by exploiting gradient maps of images to guide the recovery in two aspects. 
On the one hand, we restore high-resolution gradient maps by a gradient branch to provide additional structure priors for the SR process. 
On the other hand, we propose a gradient loss to impose a second-order restriction on the super-resolved images, which helps generative networks concentrate more on geometric structures. 
Secondly, since the gradient maps are handcrafted and may only be able to capture limited aspects of structural information, we further extend SPSR-G by introducing a learnable neural structure extractor (NSE) to unearth richer local structures and provide stronger supervision for SR. 
We propose two self-supervised structure learning methods, contrastive prediction and solving jigsaw puzzles, to train the NSEs. 
Our methods are model-agnostic, which can be potentially used for off-the-shelf SR networks.
Experimental results on five benchmark datasets show that the proposed methods outperform state-of-the-art  perceptual-driven SR methods under LPIPS, PSNR, and SSIM metrics. 
Visual results demonstrate the superiority of our methods in restoring structures while generating natural SR images. Code is available at~\url{https://github.com/Maclory/SPSR}.
\end{abstract}

}

\maketitle

\IEEEdisplaynontitleabstractindextext

%
\IEEEpeerreviewmaketitle

\section{Introduction}
\IEEEPARstart{S}{ingle} image super-resolution (SISR) aims to recover high-resolution (HR) images from their low-resolution (LR) counterparts. SISR is a fundamental problem in the community of computer vision and can be applied in many image analysis tasks including surveillance~\cite{zhang2010super,rasti2016convolutional} and satellite image~\cite{thornton2006sub,tatem2002super}. Super-resolution is a widely known ill-posed problem since each LR input may have multiple HR solutions. With the development of deep learning, a number of deep SISR methods~\cite{dong2014learning,shi2016real} have been proposed and have largely boosted the performance of super-resolution. Most of these methods are optimized by the objectives of L1 or mean squared error (MSE) which measure the pixel-wise distances between SR images and the HR ones. However, such optimizing objectives impel a deep model to produce an image which may be a statistical average of possible HR solutions to the one-to-many problem. As a result, such methods usually generate blurry images with high peak signal-to-noise ratio (PSNR). 

\begin{figure}[htbp]
\centering

\subfigure[HR]{
\begin{minipage}[b]{0.48\linewidth}
\includegraphics[width=1 \linewidth]{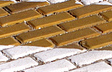}
\end{minipage}%
}%
\subfigure[RCAN~\cite{RCAN}]{
\begin{minipage}[b]{0.48\linewidth}
\includegraphics[width=1 \linewidth]{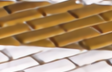}
\end{minipage}
}
\vspace{-2mm}

\subfigure[SRGAN~\cite{ledig2017photo}]{
\begin{minipage}[b]{0.48\linewidth}
\includegraphics[width=1 \linewidth]{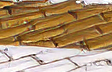}
\end{minipage}%
}%
\subfigure[ESRGAN~\cite{wang2018esrgan}]{
\begin{minipage}[b]{0.48\linewidth}
\includegraphics[width=1 \linewidth]{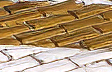}
\end{minipage}
}
\vspace{-2mm}

\subfigure[NatSR~\cite{soh2019natural}]{
\begin{minipage}[b]{0.48\linewidth}
\includegraphics[width=1 \linewidth]{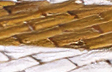}
\end{minipage}%
}%
\subfigure[SPSR (Ours)]{
\begin{minipage}[b]{0.48\linewidth}
\includegraphics[width=1 \linewidth]{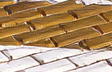}
\end{minipage}
}
\vspace{-5mm}

\centering
\caption{SR results of different methods. RCAN represents PSNR-oriented methods, typically generating straight but blurry edges for the bricks. Perceptual-driven methods including SRGAN, ESRGAN, and NatSR commonly recover sharper but geometric-inconsistent textures. Our SPSR result is sharper than that of RCAN, and preserves finer geometric structures compared with perceptual-driven methods. Best viewed on screen. }
\label{fig:head1}
\end{figure}

Hence, several methods aiming to recover photo-realistic images have been proposed recently by utilizing generative adversarial networks (GANs)~\cite{goodfellow2014generative}, such as SRGAN~\cite{ledig2017photo}, EnhanceNet~\cite{EnhanceNet}, ESRGAN~\cite{wang2018esrgan} and NatSR~\cite{soh2019natural}. 
While GAN-based methods can generate high-fidelity SR results, there are always geometric distortions along with sharp edges and fine textures. Some SR examples are presented in Fig.~\ref{fig:head1}. We can see PSNR-oriented methods like RCAN~\cite{RCAN} recover blurry but straight edges for the bricks, while edges restored by perceptual-driven methods are sharper but twisted. 
GAN-based methods generally suffer from structural inconsistency since the discriminators may introduce unstable factors to the optimization procedure. Some methods have been proposed to balance the trade-off between the merits of two kinds of SR methods. For example, Controllable Feature Space Network (CFSNet)~\cite{wang2019cfsnet} designs an interactive framework to transfer continuously between two objectives of perceptual quality and distortion reduction. Nevertheless, the intrinsic problem is not mitigated since the two goals cannot be achieved simultaneously. Hence it is necessary to explicitly guide perceptual-driven SR methods to preserve structures for further enhancing the SR performance. 

In this paper, we propose a structure-preserving super-resolution with gradient guidance  (SPSR-G) method to alleviate the above-mentioned issue. Since the gradient map reveals the sharpness of each local region in an image, we exploit this powerful tool to guide image recovery. 
On the one hand, we design a gradient branch which converts the gradient maps of LR images to the HR ones as an auxiliary SR problem. The recovered gradients are also integrated into the SR branch to provide structure priors for SR. Besides, the gradients can highlight the regions where sharpness and structures should be paid more attention to, so as to guide the high-quality generation implicitly. 
On the other hand, we propose a gradient loss to explicitly supervise the gradient maps of recovered images. Together with the original image-space objectives, the gradient loss restricts the second-order relations of neighboring pixels. By introducing the gradient branch and the gradient loss, structural configurations of images can be better retained and we are able to obtain SR results with high perceptual quality and fewer geometric distortions. 
To the best of our knowledge, we are the first to explicitly consider preserving geometric structures in GAN-based SR methods.

\begin{figure}[t]
\centering

\subfigure[SPSR with gradient guidance]{
\begin{minipage}[b]{0.95\linewidth}
\includegraphics[width=1 \linewidth]{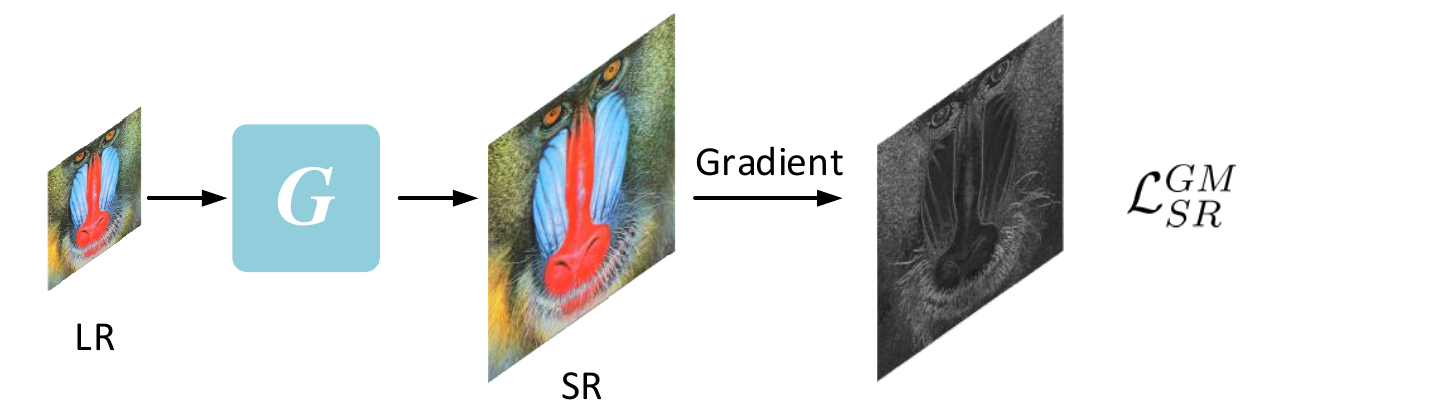}
\end{minipage}%
}%
\vspace{-2mm}

\subfigure[SPSR with self-supervised neural structure extractor (NSE)]{
\begin{minipage}[b]{0.95\linewidth}
\includegraphics[width=1 \linewidth]{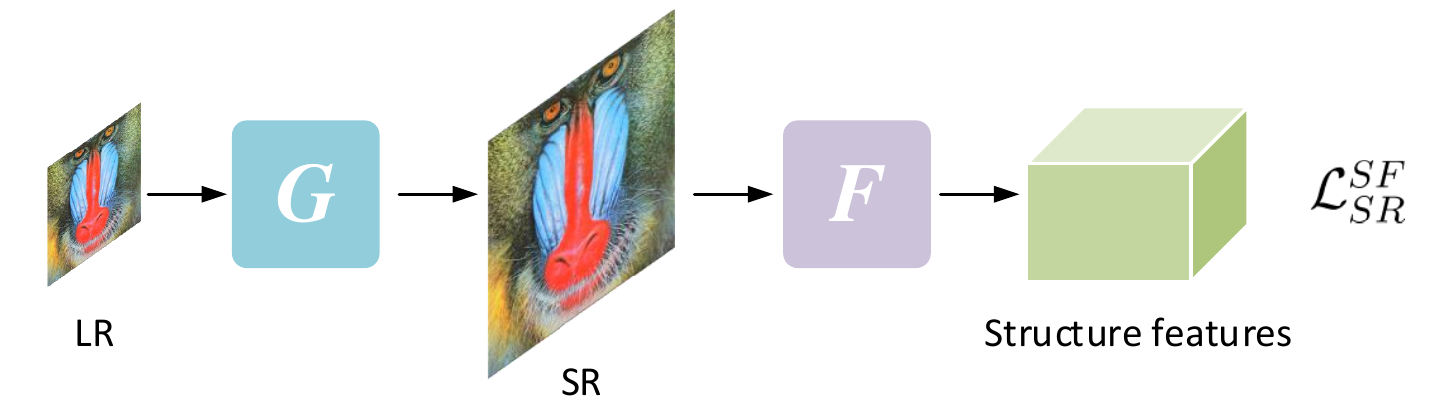}
\end{minipage}%
}%
\vspace{-2mm}

\centering
\caption{Comparison of SPSR with gradient guidance and SPSR with self-supervised neural structure extractor. The former one imposes constraints on the gradient maps (GM) of super-resolved images by hand-crafted operators. The latter one learns a powerful neural structure extractor by self-supervised structure learning methods. Additional supervision is provided by reducing the distances of the extracted structure features (SF) for SR results and the corresponding HR images. }
\label{fig:head2}
\end{figure}

While gradient maps can be exploited to preserve geometric structures of SR images, they are still extracted by hand-crafted operators and can only reflect one aspect of structures, \ie the value difference of adjacent pixels. They are limited in capturing richer structural information, such as more complex texture patterns. As a result, the supervision provided by gradient maps may not be powerful enough. To address this issue, we extend SPSR-G by introducing a learnable neural structure extractor (NSE) to capture more generic structural information for local image patches. We also design a structure loss to provide stronger supervision for the generator by constraining the distance of structure features extracted by NSE.  
By doing so, the generator can learn more knowledge on local structures. Fig.~\ref{fig:head2} illustrates the difference between the two proposed methods. In order to train the NSE, we investigate the ability of self-supervised learning methods~\cite{oord2018representation,noroozi2016unsupervised} to unearth local structures. Specifically, we design two structure learning schemes, contrastive prediction and jigsaw puzzles, for the two corresponding SPSR models, SPSR-P and SPSR-J, respectively. Different from most previous works on self-supervised learning, our method only focuses on exploring local structures instead of semantic information. 
Moreover, our methods are model-agnostic, which can be potentially used for off-the-shelf SR networks.

We conduct comprehensive experiments on five widely-used benchmark datasets. 
Experimental results on SR show that our methods succeed in enhancing SR fidelity by reducing structural distortions and are superior to state-of-the-art perceptual-driven SR methods quantitatively and qualitatively. We find the learned NSEs perform well in capturing structure information and boosting SR performance. We show that the proposed structure loss can replace the original gradient-space losses and image-space adversarial loss, which demonstrates the efficacy of the proposed method.

This paper is an extended version of our conference paper~\cite{ma2020structure}. We make the following new contributions:
\begin{enumerate}
\item We further propose a neural structural extractor to unearth more generic and complex structure information. Based on the extracted structure features from SR and HR images, we design a new structure loss to provide stronger supervision for the SR training process. 

\item We investigate two self-supervised structure learning methods for training the neural structure extractors based on contrastive prediction and jigsaw puzzles. To our best knowledge, we are the first to explore local structure information of images via self-supervised learning to boost SR performance. 

\item We conduct extensive experiments and show the excellent capacity of the proposed SR methods with the neural structure extractors.  Moreover, we have implemented more experiments on user studies, ablation studies, and parameter analysis. 

\end{enumerate}

\section{Related Work}

In this section, we briefly review the topics of single image super-resolution (SISR) 
and self-supervised learning. 

\subsection{Single Image Super-Resolution}

\textbf{PSNR-Oriented Methods}: Most previous approaches target high PSNR. As a pioneer, Dong \etal~\cite{dong2014learning} propose SRCNN, which firstly maps LR images to HR ones by a three-layer CNN. Since then, researchers in this field have been focusing on designing effective neural architectures to improve SR performance. DRCN~\cite{DRCN} and VDSR~\cite{VDSR} are proposed by Kim \etal~by implementing very deep convolutional networks and recursive convolutional networks. 
Ledig \etal~\cite{ledig2017photo} propose SRResNet by employing the idea of skip connections~\cite{he2016deep}. 
Zhang \etal~\cite{RDN} propose RDN by utilizing residual dense blocks for SR and further introduce RCAN~\cite{RCAN}
to achieve superior performance on PSNR. 
Guo \etal~\cite{guo2020closed} add an additional constraint on LR images to improve the prediction accuracy. 
Besides, recent SR methods pay attention to the relationship between intermediate features. 
Mei \etal~\cite{mei2020image} propose a cross-scale attention module to exploit the inherent image properties. 
Niu \etal~\cite{niu2020single} design a holistic attention network to establish the holistic dependencies among layers, channels, and positions. 

\textbf{Perceptual-Driven Methods}: The methods mentioned above mainly focus on achieving high PSNR and thus use the MSE or L1 loss as the objective. 
However, these methods usually produce blurry images.
Johnson \etal~\cite{johnson2016perceptual} propose perceptual loss to improve the visual quality of recovered images. 
Ledig \etal~\cite{ledig2017photo} utilize adversarial loss~\cite{goodfellow2014generative} to construct SRGAN, which becomes the first framework able to generate photo-realistic HR images. Furthermore, Sajjadi \etal~\cite{EnhanceNet} restore high-fidelity textures by texture loss. Wang \etal~\cite{wang2018esrgan} enhance the previous frameworks by introducing Residual-in-Residual Dense Block (RRDB) to the proposed ESRGAN. 
Wei \etal~\cite{wei2020component} propose a component divide-and-conquer model to recover flat regions, edges and corner regions separately. 
Recently, stochastic SR also attracts the community's attention, which injects stochasticity to the generation instead of only modeling deterministic mappings. Bahat \etal~\cite{bahat2020explorable} explore the multiple different HR explanations to the LR inputs by editing the SR outputs whose downsampling versions match the LR inputs. Besides, various generative models are exploited for this task, such as conditional generative adversarial networks~\cite{shaham2019singan}, conditional normalizing flow~\cite{lugmayr2020srflow} and conditional variational autoencoder~\cite{hyun2020varsr}. 
Although these existing perceptual-driven methods improve the overall visual quality of super-resolved images, they sometimes generate unnatural artifacts and geometric distortions when recovering details.

\textbf{Gradient-Relevant Methods}: Gradient information has been utilized in previous work~\cite{luan2017deep,anoosheh2019night}.
For SR methods, Fattal~\cite{fattal2007image} proposes a method based on edge statistics of image gradients by learning
the prior dependency of different resolutions. 
Sun \etal~\cite{sun2010gradient} propose a gradient profile prior to represent image gradients and a gradient field transformation to enhance sharpness of super-resolved images.  
Yan \etal~\cite{yan2015single} propose a SR method based on gradient profile sharpness which is extracted from gradient description models. 
In these methods, statistical dependencies are modeled by estimating HR edge-related parameters according to those observed in LR images. However, the modeling procedure is accomplished point by point, which is complex and inflexible.  
In fact, deep learning is outstanding in handling probability transformation over the distribution of pixels. However, few methods have utilized its powerful abilities in gradient-relevant SR methods. 
Zhu \etal~\cite{zhu2015modeling} propose a gradient-based SR method by collecting a dictionary of gradient patterns and modeling deformable gradient compositions. Yang \etal~\cite{yang2017deep} propose a recurrent residual network to reconstruct fine details guided by the edges which are extracted by off-the-shelf edge detector. 
While edge reconstruction and gradient field constraint have been utilized in some methods, their purposes are mainly to recover high-frequency components for PSNR-orientated SR methods. 
Similarly, SR methods based on Laplacian pyramid~\cite{lai2017deep} and wavelet transform~\cite{liu2018multi,huang2017wavelet,bae2017beyond} decompose images into multiple sub-band components for the simplification of image reconstruction. They also focus  on recovering high-frequency information of images.  
Different from these methods, we aim to reduce geometric distortions produced by GAN-based methods and exploit gradient maps as structure guidance for SR. For deep adversarial networks, gradient-space constraint may provide additional supervision for better image reconstruction. 
To the best of our knowledge, no GAN-based SR method has exploited gradient-space guidance for preserving texture structures. In this work, we aim to leverage gradient information to further improve the GAN-based SR methods.

\subsection{Self-Supervised Learning}

Self-supervised learning~\cite{jing2020self} is a subfield of unsupervised learning and has drawn an amount of attention. No annotations labeled by humans are needed for it and the cost to collect large-scale unlabeled datasets is low. Hence compared to supervised learning, self-supervised learning methods are more efficient and may have stronger generalizing ability. Besides, self-supervised models can be used as pre-trained models to provide good initializations and rich prior knowledge for the training of other tasks. Existing self-supervised methods for images can mainly be classified into generation-based methods and context-based methods. 
For the first category, image generation methods (including generative adversarial networks~\cite{goodfellow2014generative}, autoencoders~\cite{hinton2006reducing} and autoregressive models~\cite{oord2016pixel}), image inpainting methods~\cite{pathak2016context} and image super-resolution methods~\cite{maeda2020unpaired} have been proposed to estimate the distribution of training data and reconstruct realistic images. For these methods, there is a lack of further study on the performance of downstream tasks. Differently, image colorization whose goal is to colorize each pixel of a gray-scale input has been utilized as a pretext task for self-supervised representation learning~\cite{zhang2016colorful,larsson2017colorization,larsson2016learning}. 
In the second category, some methods explore the similarity of context by contrasting~\cite{chen2020simple,tian2019contrastive,oord2018representation}, which reduces the distances of positive samples while enlarging those of negative samples. 
Some other methods leverage the relationships of spatial context structures and design jigsaw puzzles as pretext tasks~\cite{noroozi2016unsupervised,kim2018learning}. 
While the learned representations of these methods are tested in downstream tasks, they mainly capture high-level semantic information. The tasks also focus on semantic understanding, such as classification, segmentation, detection, etc. In this paper,  we aim to extract low-level structure features by neural structure extractors which are trained by solving contrastive prediction and jigsaw puzzles. The learned structure features are utilized as a regularization to boost the performance of super-resolution.

\begin{figure*}
\begin{center}
\includegraphics[width=\linewidth]{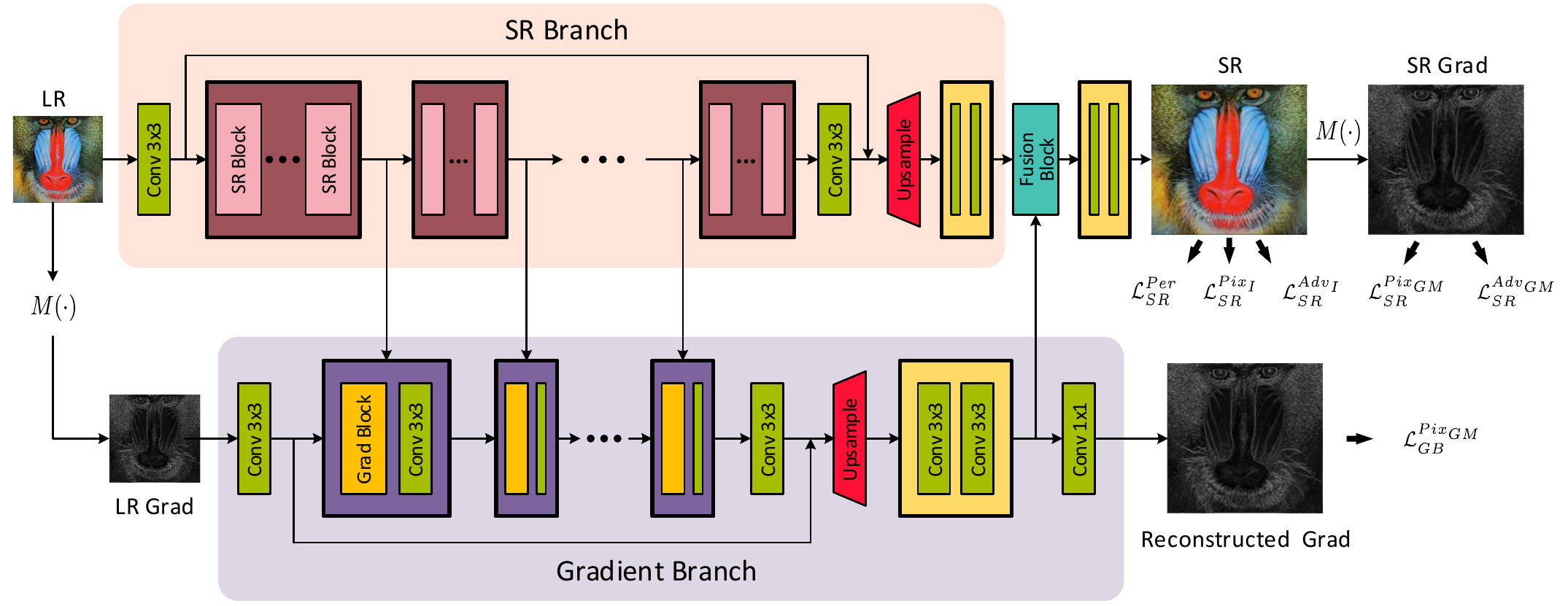}
\vspace{-9mm}
\end{center}
   \caption{Overall framework of our SPSR-G method. Our architecture consists of two branches, the SR branch and the gradient branch. The gradient branch aims to super-resolve LR gradient maps to the HR counterparts. It incorporates multi-level representations from the SR branch to reduce parameters and outputs gradient information to guide the SR process by a fusion block in return. The final SR outputs are optimized by not only conventional image-space losses, but also the proposed gradient-space objectives. }
\label{fig:framework}
\end{figure*}

\section{Structure-Preserving Super Resolution with Gradient Guidance}

In this section, we first introduce the overall framework of our structure-preserving super-resolution method with gradient guidance (SPSR-G). Then we present the details of gradient branch, structure-preserving SR branch, and final objective functions accordingly.

\subsection{Overview}

In SISR, we aim to take LR images $I^{LR}$ as inputs and generate SR images $I^{SR}$ given their HR counterparts $I^{HR}$ as ground-truth. We denote the generator as $G$ with parameters $\theta_G$. Then we have $I^{SR} = G(I^{LR}; \theta_G)$. $I^{SR}$ should be as similar to $I^{HR}$ as possible. If the parameters are optimized by an loss function $\mathcal{L}$, we have the following formulation: 
\begin{eqnarray}
\theta_G^* &=& \arg \mathop{\min}_{\theta_G} \mathbb{E}_{I^{SR}}\mathcal{L}(G(I^{LR}; \theta_G), I^{HR}). 
\end{eqnarray}

The overall framework is depicted as Fig.~\ref{fig:framework}. The generator is composed of two branches, one of which is a structure-preserving SR branch and the other is a gradient branch. The SR branch takes $I^{LR}$ as input and aims to recover the SR output $I^{SR}$ with the guidance provided by the recovered SR gradient map from the gradient branch. 

\subsection{Network Architecture}

\textbf{Gradient Branch}: 
The target of the gradient branch is to estimate the translation of gradient maps from the LR modality to the HR one. The gradient map for an image $I$ is obtained by computing the difference between adjacent pixels:
\begin{eqnarray}
I_x(\mathbf{x}) &=& I(x+1,y)-I(x-1, y), \nonumber \\
I_y(\mathbf{x}) &=& I(x, y+1)-I(x, y-1), \nonumber \\
\nabla I(\mathbf{x}) &=& (I_x(\mathbf{x}), I_y(\mathbf{x})), \nonumber \\
M(I) &=& \|\nabla I\|_2,
\end{eqnarray}
where $M(\cdot)$ stands for the operation to extract gradient map whose elements are gradient lengths for pixels with coordinates $\mathbf{x}=(x,y)$. The operation to get the gradients can be easily achieved by a convolution layer with a fixed kernel. In fact, we do not consider gradient direction information since gradient intensity is adequate to reveal the sharpness of local regions in recovered images. Hence we adopt the intensity maps as the gradient maps. 
Such gradient maps can be regarded as another kind of image, so that techniques for image-to-image translation can be utilized to learn the mapping between two modalities. The translation process is equivalent to the spatial distribution translation from LR edge sharpness to HR edge sharpness. Since most areas of the gradient map are close to zero, the convolutional neural network can concentrate more on the spatial relationship of outlines. Therefore, it may be easier for the network to capture structure dependency and consequently produce approximate gradient maps for SR images. 

As shown in Fig.~\ref{fig:framework}, the gradient branch incorporates several intermediate-level representations from the SR branch. The motivation of such a scheme is that the well-designed SR branch is capable of carrying rich structural information which is pivotal to the recovery of gradient maps. Hence we utilize the features as a strong prior to promote the performance of the gradient branch, whose parameters can be largely reduced in this case. 
Between every two intermediate features, we employ an Residual in Residual Dense Block (RRDB)~\cite{wang2018esrgan}, termed a gradient block, to extract features for the generation of gradient maps.
Once we get the SR gradient maps by the gradient branch, we can integrate the obtained gradient features into the SR branch to guide SR reconstruction in return. The magnitude of gradient maps can implicitly reflect whether a recovered region should be sharp or smooth. In practice, we feed the feature maps produced by the next-to-last layer of the gradient branch to the SR branch. Meanwhile, we generate output gradient maps by a $1\times1$ convolution layer with these feature maps as inputs. 

\textbf{Structure-Preserving SR Branch}:
We design a structure-preserving SR branch to get the final SR outputs. This branch constitutes two parts. The first part is a regular SR network comprising of multiple generative blocks which can be any architecture. Here we introduce the backbone of ESRGAN~\cite{wang2018esrgan}. There are 23 RRDBs in the original model. Therefore, we incorporate the feature maps from the 5th, 10th, 15th, 20th blocks to the gradient branch. Since regular SR models produce images with only 3 channels, we remove the last convolutional reconstruction layer and feed the output feature to the consecutive part. 
The second part of the SR branch wires the SR gradient feature maps obtained from the gradient branch as mentioned above. 
We fuse the structure information by a fusion block which concatenates the features from two branches together. Then we use another RRDB block and convolutional layers to reconstruct the final SR features and produce SR results.

\subsection{Objective Functions}

\textbf{Conventional Loss}:
Most SR methods optimize the elaborately designed networks by a common pixelwise loss, which is efficient for the task of super-resolution measured by PSNR. This metric can reduce the average pixel difference between recovered images and ground-truths but the results may be too smooth to maintain sharp edges for visual effects. However, this loss is still widely used to accelerate convergence and improve SR performance: 
\begin{eqnarray}
\mathcal{L}^{Pix_I}_{SR} &=& \mathbb{E}_{I^{SR}} \|G(I^{LR})-I^{HR}\|_1.
\end{eqnarray}

Perceptual loss has been proposed in~\cite{johnson2016perceptual} to improve perceptual quality of recovered images. Features containing semantic information are extracted by a pre-trained VGG network~\cite{simonyan2014very}. The Euclidean distances between the features of HR images and SR ones are minimized in perceptual loss:
\begin{eqnarray}
\mathcal{L}^{Per}_{SR} &=& \mathbb{E}_{I^{SR}} \|\phi_i(G(I^{LR}))-\phi_i(I^{HR})\|_1, 
\end{eqnarray}
where $\phi_i(.)$ denotes the $i$th layer output of the VGG model.

Methods~\cite{ledig2017photo,wang2018esrgan} based on GANs~\cite{goodfellow2014generative,berthelot2017began,RaGAN,radford2015unsupervised,arjovsky2017wasserstein,gulrajani2017improved} also play an important role in the SR problem. The discriminator $D_I$ and the generator $G$ are optimized by a two-player game as follows: 
\begin{eqnarray}
\mathcal{L}^{{Dis}_I}_{SR} &=&  -\mathbb{E}_{I^{SR}}[\log (1-D_I(I^{SR}))] \nonumber \\
&&-\mathbb{E}_{I^{HR}}[\log D_I(I^{HR})], \\
\mathcal{L}^{{Adv}_I}_{SR} &=& -\mathbb{E}_{I^{SR}}[\log D_I(G(I^{LR}))].
\end{eqnarray}

Following~\cite{RaGAN,wang2018esrgan} we conduct relativistic average GAN (RaGAN) to achieve better optimization in practice. 
For convenience, we define $\mathcal{L}^{I}_{SR}$ as the summation of the losses imposed on the image space, $\mathcal{L}^{Per}_{SR}$, $\mathcal{L}^{Pix_I}_{SR}$ and $\mathcal{L}^{{Adv}_I}_{SR}$. 
Models supervised by the above objective functions merely consider the image-space constraint for images but neglect the semantically structural information provided by the gradient space. While the generated results look  photo-realistic, there are also a number of undesired geometric distortions. Thus we introduce the gradient loss to alleviate this issue. 

\begin{figure}[t]
\centering

\subfigure[HR]{
\begin{minipage}[b]{0.3\linewidth}
\includegraphics[width=1 \linewidth]{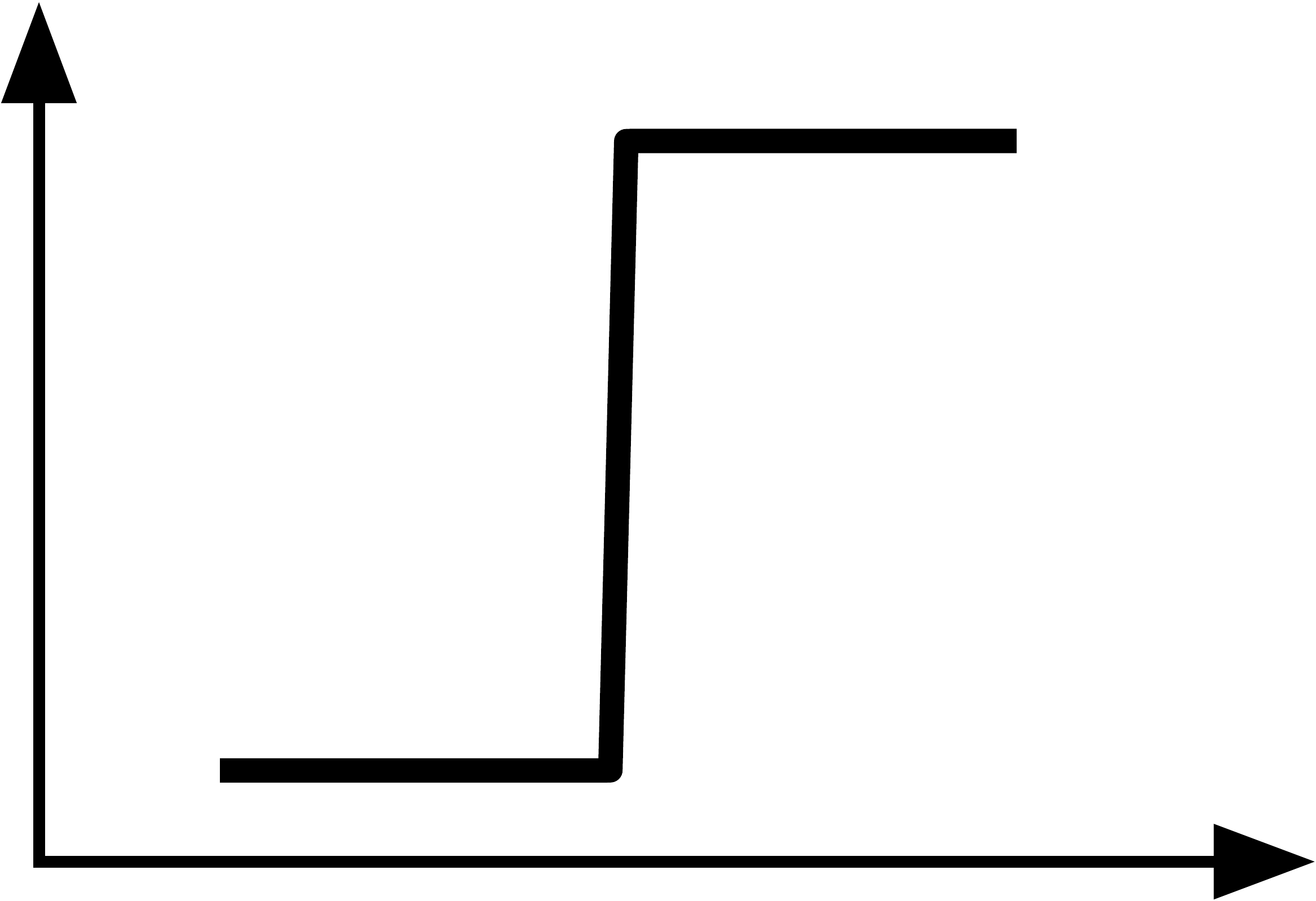}
\end{minipage}%
}\hspace{1pt}
\subfigure[Blurry SR]{
\begin{minipage}[b]{0.3\linewidth}
\includegraphics[width=1 \linewidth]{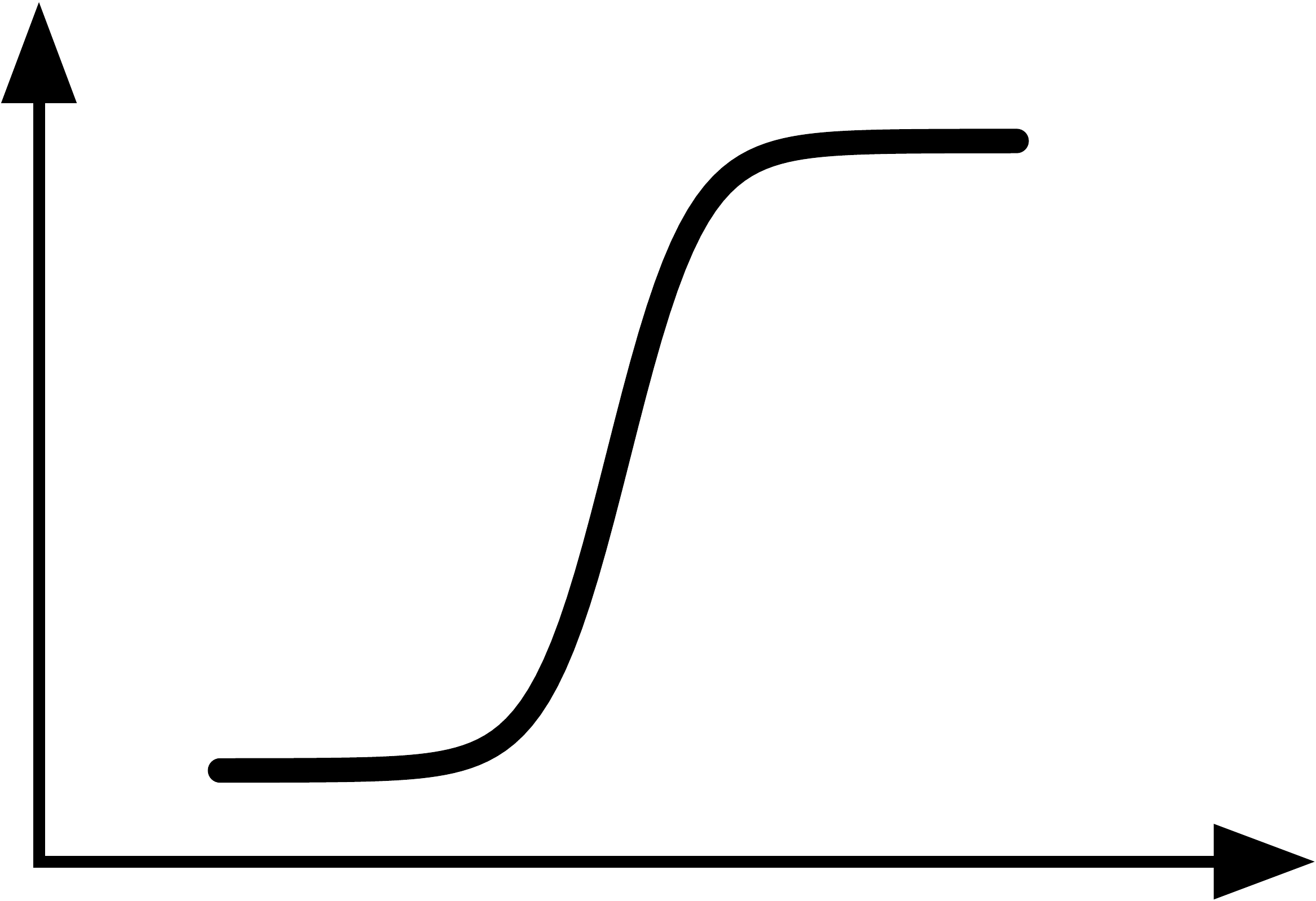}
\end{minipage}%
}\hspace{1pt}
\subfigure[Sharp SR]{
\begin{minipage}[b]{0.3\linewidth}
\includegraphics[width=1 \linewidth]{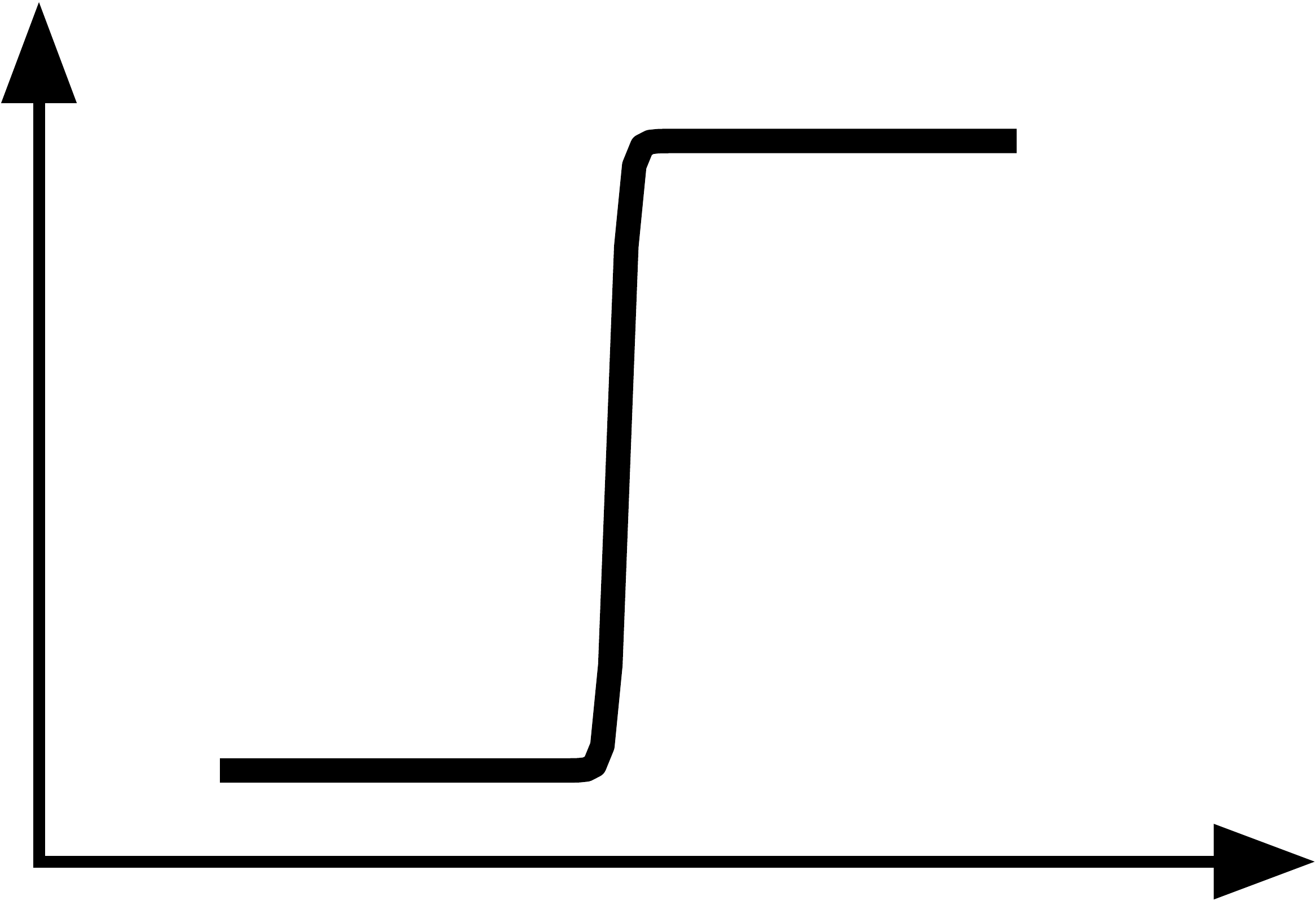}
\end{minipage}%
}
\vspace{-2mm}

\subfigure[HR Gradiant]{
\begin{minipage}[b]{0.3\linewidth}
\includegraphics[width=1 \linewidth]{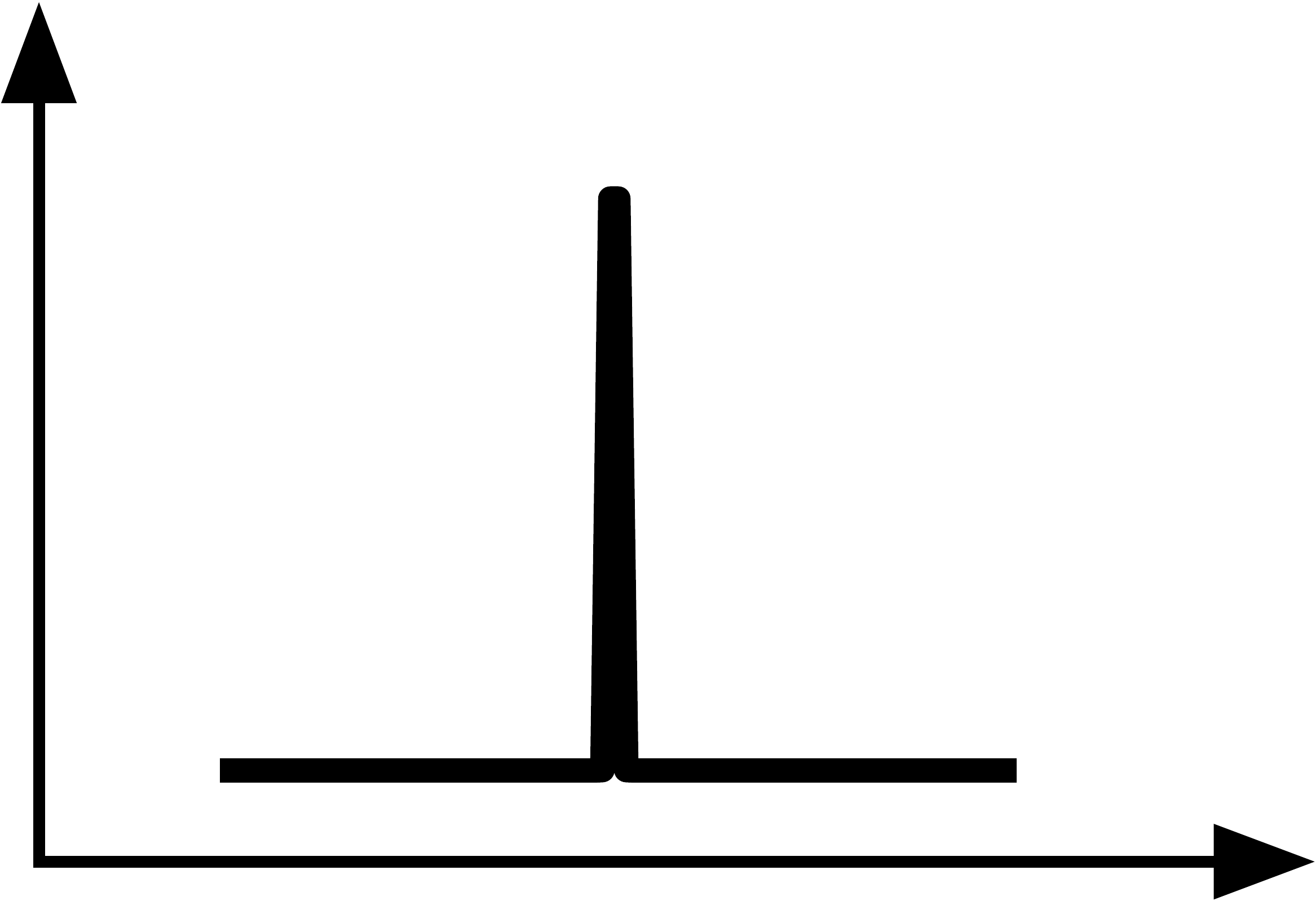}
\end{minipage}%
}\hspace{1pt}
\subfigure[Blurry Gradiant]{
\begin{minipage}[b]{0.3\linewidth}
\includegraphics[width=1 \linewidth]{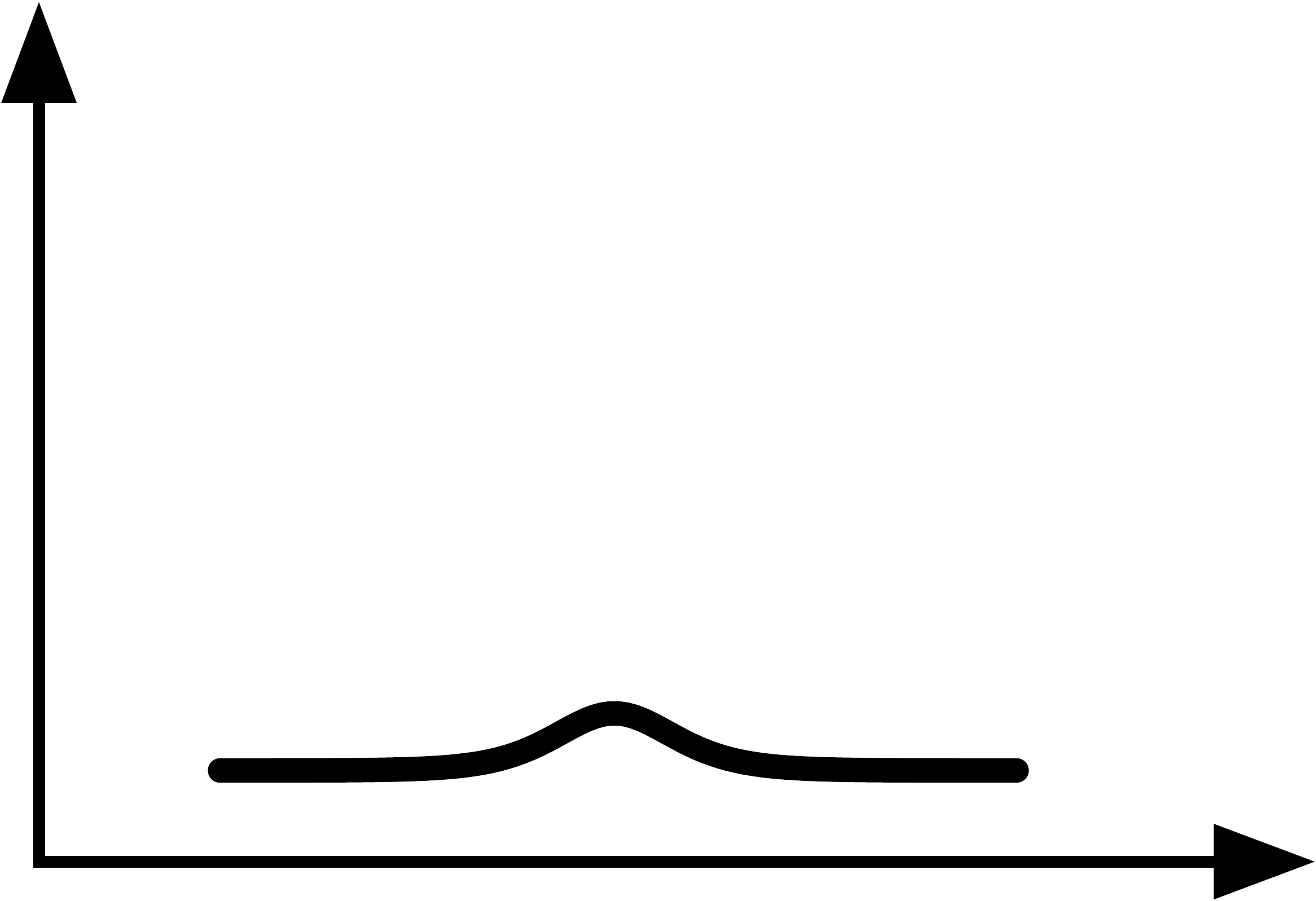}
\end{minipage}%
}\hspace{1pt}
\subfigure[Sharp Gradiant]{
\begin{minipage}[b]{0.3\linewidth}
\includegraphics[width=1 \linewidth]{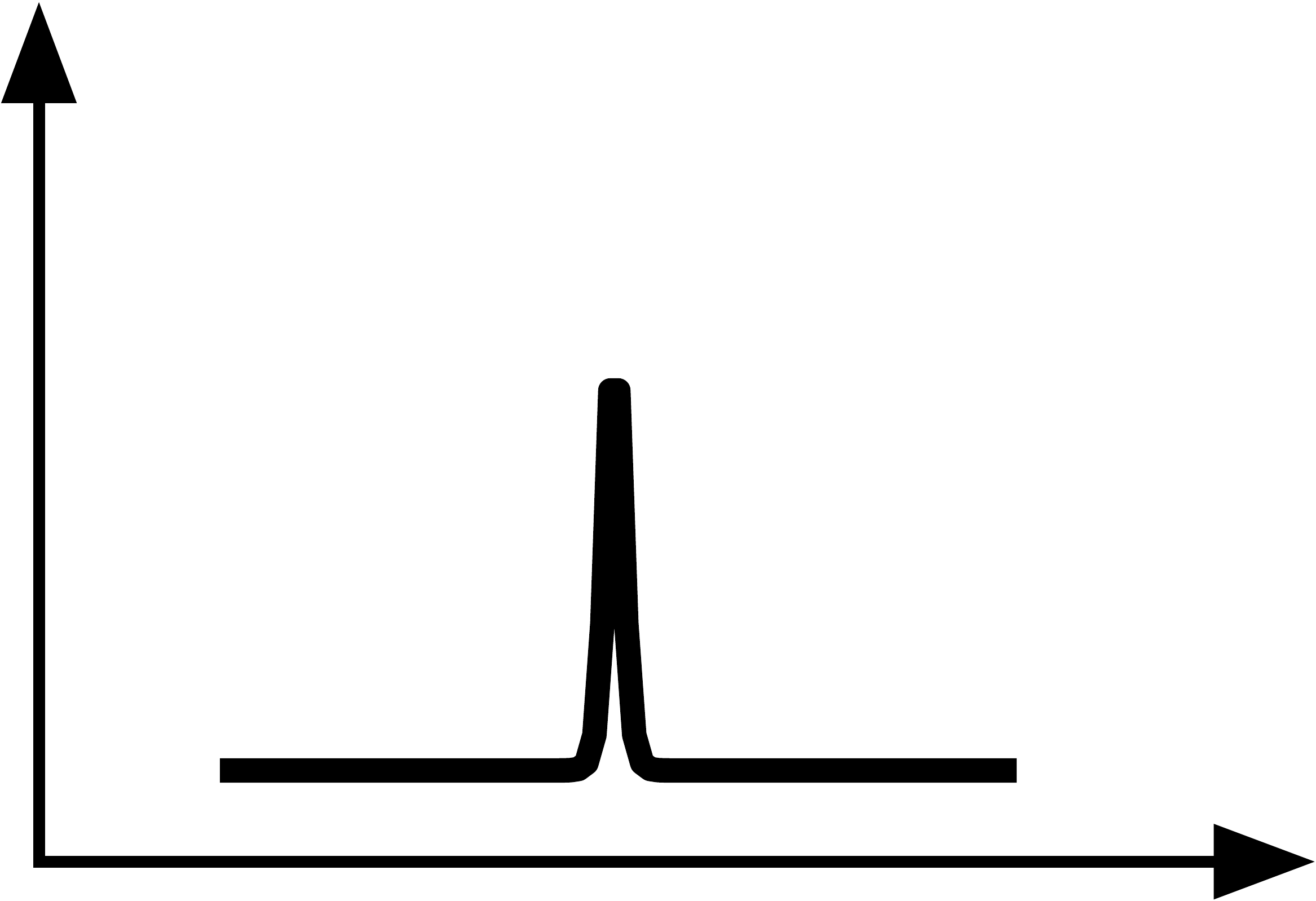}
\end{minipage}%
}
\vspace{-2mm}

\centering
\caption{Illumination of a simple 1-D case. The first row shows the pixel sequences and the second row shows the corresponding gradient maps. }
\label{fig:method_graph}
\end{figure} 

\textbf{Gradient Loss}: 
Our motivation can be illustrated clearly by Fig.~\ref{fig:method_graph}. Here we only consider a simple 1-dimensional case. If the model is only optimized in image space by the L1 loss, we usually get an SR sequence as Fig.~\ref{fig:method_graph} (b) given an input testing sequence whose ground-truth is a sharp edge as Fig.~\ref{fig:method_graph} (a). The model fails to recover sharp edges for the reason that the model tends to give a statistical average of possible HR solutions from training data. 
In this case, if we compute and show the gradient magnitudes of two sequences, it can be observed that the SR gradient is flat with low values while the HR gradient is a spike with high values. They are far from each other. This inspires us that if we add a second-order gradient constraint to the optimization objective, the model may learn more from the gradient space. It helps the model
focus on neighboring configuration so that the local intensity of sharpness can be inferred more appropriately. Therefore, if the gradient information as Fig.~\ref{fig:method_graph} (f) is captured, the probability of recovering Fig.~\ref{fig:method_graph} (c) is increased significantly. SR methods can benefit from such guidance to avoid over-smooth or over-sharpening restoration. Moreover, it is easier to extract geometric characteristics in the gradient space. Hence geometric structures can be also preserved well, resulting in more photo-realistic SR images.

Here we propose a gradient loss to achieve the above goals. Since we have mentioned the gradient map is an ideal tool to reflect structural information of an image, it can also be utilized as a second-order constraint to provide supervision to the generator.  We formulate the gradient loss by diminishing the distance between the gradient map extracted from the SR image and the one from the corresponding HR image. 
With the supervision in both image and gradient domains, the generator can not only learn fine appearance but also attach importance to avoiding geometric distortions. We design two terms of loss to penalize the difference in the gradient maps (GM) of the SR and HR images. One is based on the pixelwise loss as follows:
\begin{eqnarray}
\mathcal{L}^{{Pix}_{GM}}_{SR} = \mathbb{E}_{I^{SR}} \|M(G(I^{LR}))-M(I^{HR})\|_1. 
\end{eqnarray}
The other is to discriminate whether a gradient patch is from the HR gradient map. We design another gradient discriminator network to achieve this goal: 
\begin{eqnarray}
\mathcal{L}^{{Dis}_{GM}}_{SR} &=&  -\mathbb{E}_{I^{SR}}[\log (1-D_{GM}(M(I^{SR})))] \nonumber \\
&& -\mathbb{E}_{I^{HR}}[\log D_{GM}(M(I^{HR}))].
\end{eqnarray}
The gradient discriminator can also supervise the generation of SR results by adversarial learning:
\begin{eqnarray}
\mathcal{L}^{{Adv}_{GM}}_{SR} &=& -\mathbb{E}_{I^{SR}}[\log D_{GM}(M(G(I^{LR})))].
\end{eqnarray}

Further, we define the losses imposed on the gradient space as $\mathcal{L}^{GM}_{SR}$, which is the summation of $\mathcal{L}^{{Pix}_{GM}}_{SR}$ and $\mathcal{L}^{{Adv}_{GM}}_{SR}$.
Note that each step in $M(\cdot)$ is differentiable. Hence the model with gradient loss can be trained in an end-to-end manner. Furthermore, it is convenient to adopt gradient loss as additional guidance in any generative model due to the concise formulation and strong transferability.

\textbf{Overall Objective}: 
In conclusion, we have two discriminators $D_I$ and $D_{GM}$ which are optimized by $\mathcal{L}^{{Dis}_I}_{SR}$ and $\mathcal{L}^{{Dis}_{GM}}_{SR}$, respectively. For the generator, two terms of loss are used to provide supervision signals simultaneously. One is imposed on the structure-preserving SR branch while the other is to reconstruct high-quality gradient maps by minimizing the pixelwise loss $\mathcal{L}^{Pix_{GM}}_{GB}$ in the gradient branch (GB). 
The overall objective is defined as follows: 
\begin{eqnarray}
\mathcal{L}^{G} &=& \mathcal{L}^{G}_{SR}+\mathcal{L}^{G}_{GB} \notag \\ 
&=&\mathcal{L}^{I}_{SR}+\mathcal{L}^{GM}_{SR}+\mathcal{L}^{G}_{GB} \notag \\
&=&\mathcal{L}^{Per}_{SR}+\beta^I_{SR}\mathcal{L}^{Pix_I}_{SR}+\gamma^I_{SR}\mathcal{L}^{{Adv}_I}_{SR}+\beta^{GM}_{SR}\mathcal{L}^{Pix_{GM}}_{SR} \notag \\ 
&&+\gamma^{GM}_{SR}\mathcal{L}^{Adv_{GM}}_{SR}+\beta^{GM}_{GB}\mathcal{L}^{Pix_{GM}}_{GB}. 
\end{eqnarray}
$\beta^I_{SR}$, $\gamma^I_{SR}$, $\beta^{GM}_{SR}$, $\gamma^{GM}_{SR}$ and $\beta^{GM}_{GB}$ denote the trade-off parameters of different losses. Among these, $\beta^I_{SR}$, $\beta^{GM}_{SR}$ and $\beta^{GM}_{GB}$ are the weights of the pixel losses for SR images, gradient maps of SR images and SR gradient maps respectively. $\gamma^I_{SR}$ and $\gamma^{GM}_{SR}$ are the weights of the adversarial losses for SR images and their gradient maps. 

\textbf{Discussion}:
Here we discuss whether images with few geometric structures affect the SR capacity of our method.
Intuitively, we have an observation on SR: due to the ill-posed nature of SR, slight discrepancies in textures are more inevitable and acceptable than those in structures. Hence edges may be more important than textures for SR images. 
Besides, images with no or few geometric structures usually have relatively low gradient values and small variations in pixel values. Therefore the missing pixels are easier to infer given the LR inputs. Excellent solutions have been provided for such cases by existing methods~\cite{ledig2017photo,wang2018esrgan}. 
Methodologically, our goal is to take a step forward by preserving structures of SR images based on previous work. 
Therefore we integrate the information from the gradient branch to the SR branch by an adaptive fusion module composed of a concatenation layer and learnable convolutions. Even if the gradient branch may produce unhelpful or inaccurate details, they can be filtered out by the fusion module and hence are less vulnerable to the final SR performance. 

\section{Structure-Preserving Super Resolution with Neural Structure Extractor}

While gradient maps can provide additional structural supervision to the SR training process, there may still be some limitations with gradient maps. 
Gradients can only reflect the difference between adjacent pixels but neglects other aspects of  structures, such as more complex texture patterns. Hence the extracted structural information may not be powerful enough. 
The exploration of richer local structures needs larger receptive fields than gradient maps. 

In order to address these limitations, we propose a neural structure extractor (NSE) to extract local structural features for images. Different from previous handcrafted operators, the NSE is a learnable small neural network and is able to encode neural features which have stronger representing abilities to capture structural information. The extracted features can be utilized to design structural criteria, which provides powerful supervision on whether and how the generator should modify its recovery. We propose two self-supervised structure learning methods,  contrastive prediction and solving jigsaw puzzles, to train the NSEs. We name the corresponding SR models SPSR-P and SPSR-J, respectively. Different from existing self-supervised learning methods including~\cite{oord2018representation} and~\cite{noroozi2016unsupervised}, our methods focus on exploring local structures instead of handling holistic semantics. The learned representations contain more comprehensive local structural information than the original handcrafted gradient operator.

\begin{figure*}[htbp]
\centering

\subfigure[Learning neural structure extractor $F_P$ by contrastive prediction]{
\begin{minipage}[b]{0.9\linewidth}
\includegraphics[width=1 \linewidth]{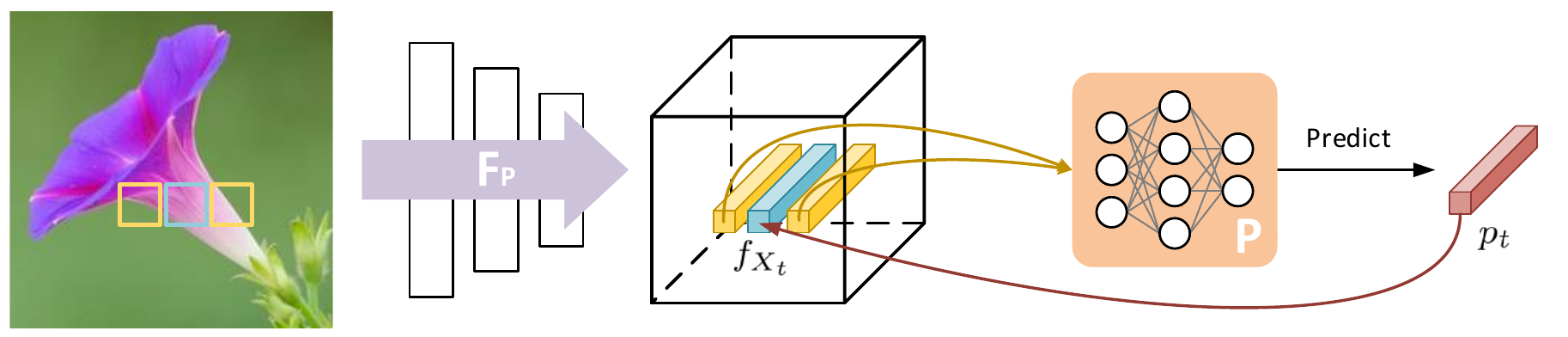}
\end{minipage}%
}%
\vspace{-3mm}

\subfigure[Learning neural structure extractor $F_J$ by solving jigsaw puzzles]{
\begin{minipage}[b]{0.9\linewidth}
\includegraphics[width=1 \linewidth]{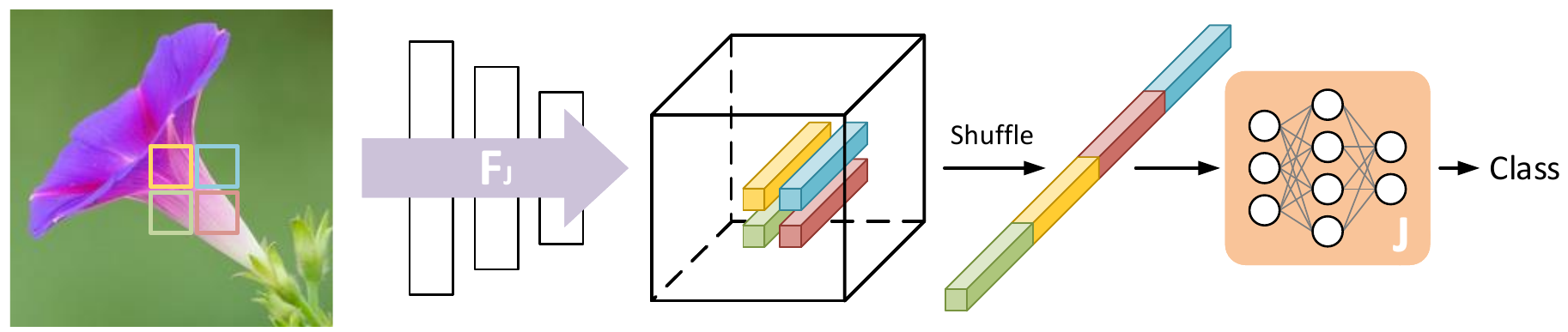}
\end{minipage}%
}%
\vspace{-3mm}

\centering
\caption{Illustration of the self-supervised structure learning schemes for training neural structure extractors. The top figure shows the flowchart of learning by contrastive prediction while the bottom one is by solving jigsaw puzzles. }
\label{fig:ss}
\end{figure*}

\subsection{Neural Structure Extractor Architecture}

We denote the neural structure extractor as $F$ with parameters $\theta_F$. Given an image $I$, we get the structure feature by $f=F(I)$. Instead of using a feature vector to represent the whole image like the practice of high-level semantic understanding tasks, we let $F(I)$ be a feature map where the channel-wise vector of each spatial position models the corresponding neighboring structures on the image. We use several residual blocks~\cite{he2016deep} to build our model, whose detailed architecture can be found in the supplementary material. 
It is unnecessary to contain a number of convolutional layers in the model since the receptive field of each obtained feature vector should be relatively small. Otherwise, the feature may cover large image regions and tend to learn holistic semantic information of objects instead of geometric structures. Hence our model extracts structure features with a receptive field of $31\times31$. Such a scale is enough for modeling local geometric appearance. Moreover, the spatial size of the extracted feature map is downsampled by a factor of 4 compared to the input image size in order to compress redundant information and reduce computational cost. As a result, the receptive fields of adjacent features have a stride of 4 on spatial dimensions.

\subsection{Self-Supervised Structure Learning}

In this section, we describe the training strategy for neural structure extractors and explain how the extracted features are able to represent local structures of image patches. We investigate two methods for structure learning by solving contrastive prediction and jigsaw puzzles, respectively. 

\textbf{Contrastive Prediction}: 
Due to the associations of contextual image structures, we are able to predict local structures for a target location $X_t$ given the structural information of a set of neighboring locations $X_1, X_2, ..., X_M$. Based on this observation, by employing contrastive learning which reduces the feature distance of related locations and enlarges the feature distance of unrelated locations, we induce the structure features to encode underlying contextual structures. 
Specifically, we define the NSE for contrastive prediction as $F_P$ and the structure feature for a location $X_i$ as $f_{X_i}$. Then we can predict the structure representation for $X_t$ by $p_t=P(f_{X_1}, f_{X_2}, ... , f_{X_M})$, where $P$ is an autoregressive model composed of several fully connected layers and ReLU~\cite{nair2010rectified} activation layers. $P$ takes the concatenation of $f_{X_1}, f_{X_2}, ... , f_{X_M}$ as input and outputs $p_t$. 
If $P$ is capable of successfully predicting $f_{X_t}$ given the neighboring structure features $f_{X_1}, f_{X_2}, ..., f_{X_M}$ as inputs, the ability of $F_P$ to extract structural information is verified. Hence $p_t$ should be close to the positive feature $f_{X_t}$ and far from negative features from unrelated locations. The negative features are from the same structure feature map as the positive feature,  but their receptive fields have no overlap with $f_{X_t}$ and $f_{X_1}, f_{X_2}, ..., f_{X_M}$. 
To achieve this goal, we implement the InfoNCE loss~\cite{oord2018representation} to distinguish $f_{X_t}$ from the feature sets $\mathcal{S}=\{f_{X_t}, f_1, f_2, ..., f_{N_I}\}$, where $f_1, f_2, ..., f_{N_I}$ are the negative features from $N_I$ irrelevant locations on $F_P(I)$. The objective can be formulated as:
\begin{eqnarray}
\mathcal{L}^{Pred}_{F}=-\mathbb{E}_{\mathcal{S}}\left[\log\frac{h(f_{X_t}, p_t)}{h(f_{X_t}, p_t)+\sum_{n=1}^{N_{I}} h(f_n, p_t)} \right], 
\end{eqnarray}
where $h(f, p_t)$ calculates the similarity between a feature sample $f$ and the predicted feature $p_t$. $h(f, p_t)$ should has high values for positive features and low values for negative samples. We use cosine distance as the similarity metric:
\begin{eqnarray}
h(f, p_t) &=& \exp\left(\tau\cdot\frac{f}{\|f\|}\cdot\frac{p_t}{\|p_t\|}\right), 
\end{eqnarray}
where $\tau$ is a hyper-parameter to control the dynamic range. 

Note that the receptive fields of $f_{X_1}, f_{X_2}, ..., f_{X_M}$ have no overlap with those of the features from $\mathcal{S}$. The reason is that the latent representation of the target position should be inferred by the consistency of local geometric structures, instead of shared information of the overlap regions. Similarly, the discrimination of negative features should also depend on the inconsistency of local structures. In practice, we can design different sampling strategies to specify $X_t$ and $f_{X_1}, f_{X_2}, ..., f_{X_M}$. An example of horizontal sampling is shown in Fig.~\ref{fig:ss} (a). The relative locations of these points are fixed during training and testing. 

\textbf{Jigsaw Puzzles}: Here we describe the details of training the NSE by solving jigsaw puzzles. Similar to the contrastive prediction mentioned above, we do not aim to let the network handle semantic information by understanding what parts an object is made of like previous methods~\cite{noroozi2016unsupervised}. We just concentrate on the structures. Therefore, our goal to solve jigsaw puzzles is to make the extracted features capture local structural patterns, such as color distribution and edge continuity. Hence we select several neighboring features from $F_J(I)$ as $f_{X_i}$, where $X_i$ is the selected coordinates among $X_1, X_2, ..., X_M$ and $F_J$ is the NSE trained by solving jigsaw puzzles. The receptive fields of these features also have no overlaps since we want the jigsaw puzzle to be solved by the inference of structures, instead of information sharing. In our implementation, $M$ is set to 4 and the selected vectors are located in a $2\times2$ grid with fixed relative positions. As a result, there are $4!=24$ permutations to rearrange the features and the patches with respect to their corresponding receptive fields. $f_{X_1}, f_{X_2}, ..., f_{X_M}$ are concatenated by a random permutation as the input to the classifier $J$, which has three fully connected layers and outputs the possibility of each permutation. By forcing the network to classify the correct permutation labels, the learned features possess enough representation abilities for local structures. We use the cross-entropy loss as the objective function for the classification problem: 
\begin{eqnarray}
\mathcal{L}^{Jig}_{F} &=& -\mathbb{E}_c\log\left(\frac{\exp(c_y)}{\sum_{j=1}^{M!}\exp{c_j}} \right), 
\end{eqnarray}
where $c$ represents the output activation vector of the classifier $J$. 
$c_j$ and $c_y$ are the $j$th and $y$th elements of $c$, respectively, where $y$ is the index of the groundtruth permutation. 

\subsection{Structure Loss for Super-Resolution}

After obtaining the neural structure extractor, we design structure criteria based on it to train the SPSR-P and SPSR-J models. Similar to the gradient loss described above, we propose a structure loss for structure-preserving super-resolution. The structure loss is also composed of two terms of loss functions, the pixelwise loss and the adversarial loss. The pixelwise loss is utilized to constrain the similarity between the structure features (SF) extracted from the HR images and the super-resolved ones. Such loss can encourage the generator to capture hierarchical structural information from the HR images and modify the generation of SR images accordingly. This term of loss can be formulated as:
\begin{eqnarray}
\mathcal{L}^{{Pix}_{SF}}_{SR} = \mathbb{E}_{I^{SR}} \|F(G(I^{LR}))-F(I^{HR})\|_1, 
\end{eqnarray}
where  $F$ represents the learned NSE. Moreover, we design another discriminator $D_{SF}$ for the adversarial loss. The discriminator is similar to that used for the gradient loss, but has small differences in the network architecture since the structure features have different scales from the gradient maps. 
By adversarial learning, $D_{SF}$ provides training signals to the SR generator:
\begin{eqnarray}
\mathcal{L}^{{Adv}_{SF}}_{SR} &=& -\mathbb{E}_{I^{SR}}[\log D_{SF}(F(G(I^{LR})))].
\end{eqnarray} 

We denote $\mathcal{L}^{SF}_{SR}$ as the combination of these two losses, $\mathcal{L}^{{Pix}_{SF}}_{SR}$ and $\mathcal{L}^{{Adv}_{SF}}_{SR}$, as follows:
\begin{eqnarray}
\mathcal{L}^{SF}_{SR} &=& \beta_{SR}^{SF}\mathcal{L}^{{Pix}_{SF}}_{SR}+\gamma^{SF}_{SR}\mathcal{L}^{{Adv}_{SF}}_{SR},
\end{eqnarray} 
where $\beta_{SR}^{SF}$ and $\gamma^{SF}_{SR}$ are the trade-off parameters for the two terms. Different from SPSR-G, we remove the adversarial loss of the image space $\mathcal{L}^{{Adv}_I}_{SR}$ and the losses of the gradient space, $\mathcal{L}^{Pix_{GM}}_{SR}$ and $\mathcal{L}^{Adv_{GM}}_{SR}$. We replace these losses with the structure losses. In Section~\ref{sec:experiment}, analysis on experimental results show that the proposed structure loss has stronger capacity in supervising the SR process than the gradient loss and the original image-space adversarial loss. The objective function is summarized as:  

\begin{eqnarray}
\mathcal{L}^{G}_{SR} &=&\mathcal{L}^{Per}_{SR}+\beta^I_{SR}\mathcal{L}^{Pix_I}_{SR}+\mathcal{L}^{SF}_{SR}.
\end{eqnarray}

\section{Experiments}
\label{sec:experiment}

\begin{figure}
    \centering
    \subfigure{
    \begin{minipage}[b]{\linewidth}
    \includegraphics[width=\linewidth]{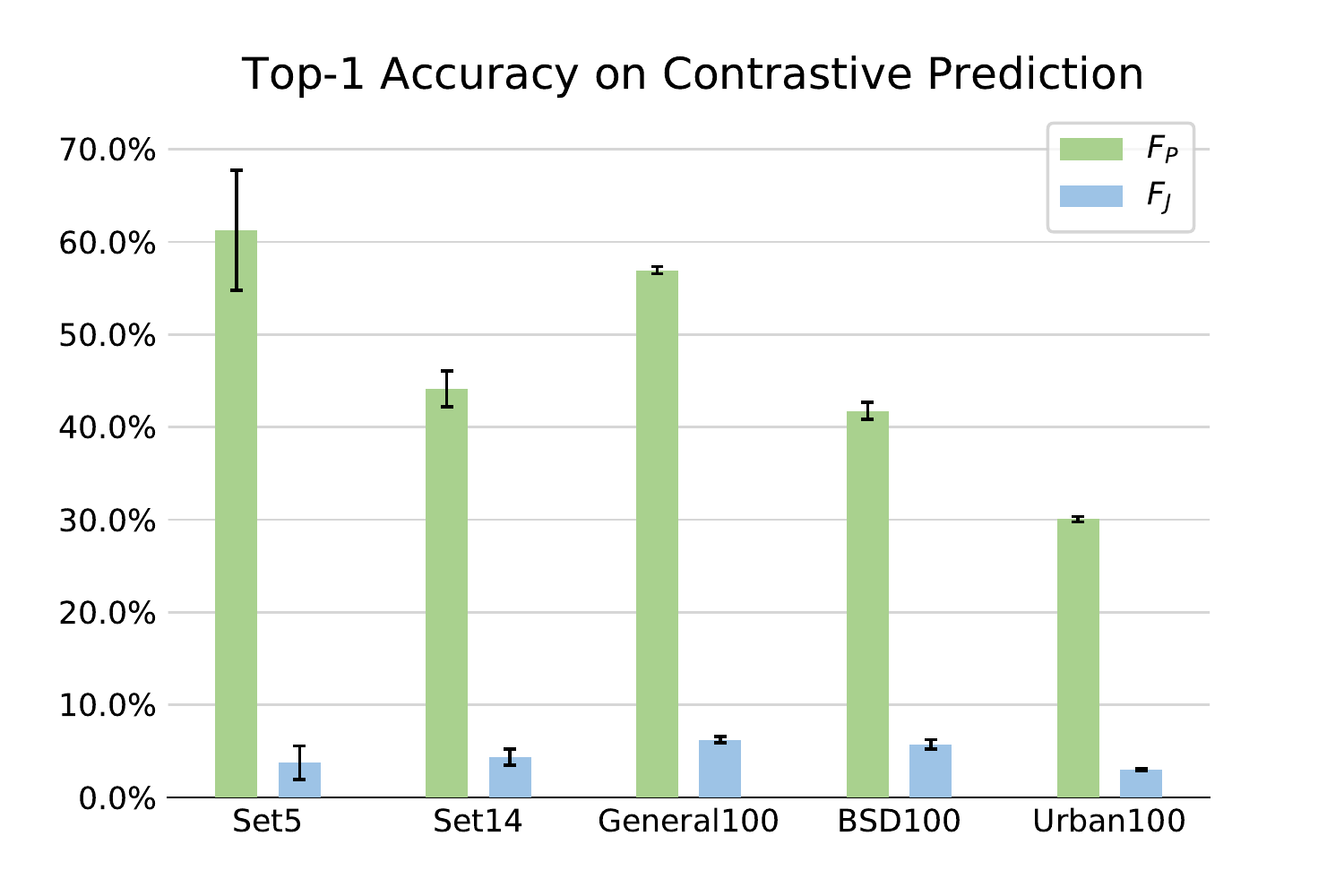}
    \end{minipage}
    }
    \vspace{-5mm}

    \centering
    \subfigure{
    \begin{minipage}[b]{\linewidth}
    \includegraphics[width=\linewidth]{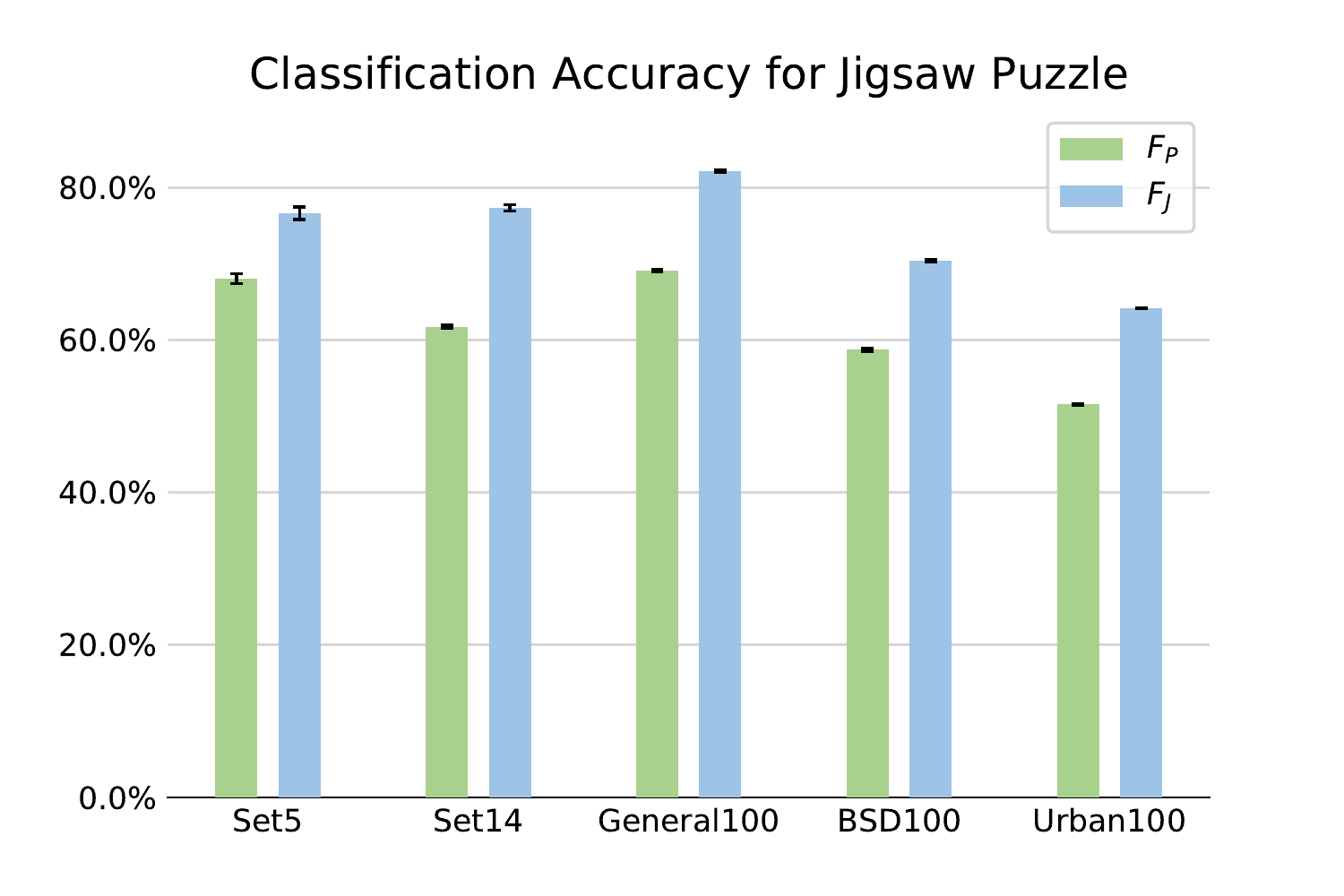}
    \end{minipage}
    }
    \vspace{-6mm}
    \caption{Validation results of the obtained NSEs on the two self-supervised structure learning tasks.  }
    \label{fig:jigsaw}
\end{figure}

In this section, we conduct experiments to evaluate the proposed methods of SPSR-G, SPSR-P, and SPSR-J. First, we introduce the implementation details. Second, we investigate the structure-capturing abilities of two self-supervised learning methods. Third, we compare our proposed methods with state-of-the-art perceptual-driven SR methods and analyze the influence of components of SPSR-G and SPSR-P. 

\subsection{Implementation Details}

\textbf{Datasets and Evaluation Metrics}: For evaluating the performance of the proposed SPSR methods, We utilize DIV2K~\cite{agustsson2017ntire} as the training dataset and five commonly used benchmarks for testing: Set5~\cite{bevilacqua2012low}, Set14~\cite{zeyde2010single}, BSD100~\cite{BSD100}, Urban100~\cite{huang2015single} and General100~\cite{dong2016accelerating}. Following~\cite{ledig2017photo,wang2018esrgan}, we downsample HR images by bicubic interpolation to get LR inputs with a scaling factor of $4\times$. We choose Learned Perceptual Image Patch Similarity (LPIPS)~\cite{zhang2018unreasonable}, PSNR and Structure Similarity (SSIM)~\cite{wang2004image} as the evaluation metrics. Lower LPIPS values indicate higher perceptual quality. 

\begin{figure*}[htbp]
	\newlength\fsdttwofig
	\setlength{\fsdttwofig}{-1.5mm}
	\scriptsize
	\centering
	
	\begin{tabular}{cc}
		
		\begin{adjustbox}{valign=t}
		\tiny
			\begin{tabular}{c}
				\includegraphics[width=0.25\textwidth]{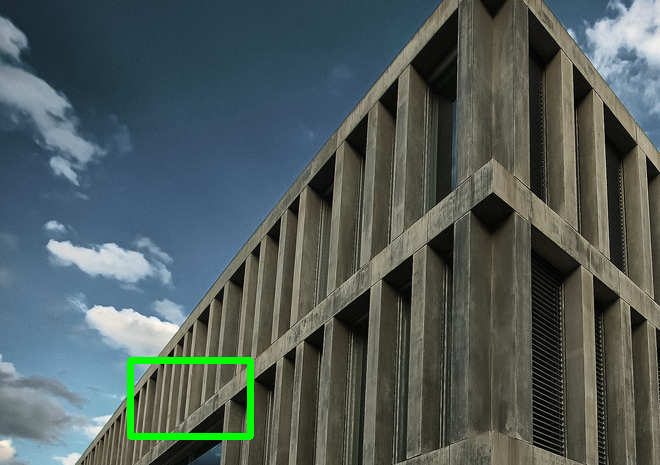}
				\\
				 `img\_027' from Urban100
				
			\end{tabular}
		\end{adjustbox}
		\hspace{-2.3mm}
		\begin{adjustbox}{valign=t}
		\tiny
			\begin{tabular}{ccccc}
				\includegraphics[width=\widthscalefive \textwidth]{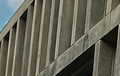} \hspace{\fsdttwofig} &
				\includegraphics[width=\widthscalefive \textwidth]{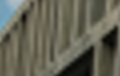} \hspace{\fsdttwofig} &
				\includegraphics[width=\widthscalefive \textwidth]{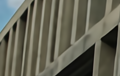} \hspace{\fsdttwofig} &
				\includegraphics[width=\widthscalefive \textwidth]{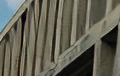} 
				\\
				HR \hspace{\fsdttwofig} &
				Bicubic \hspace{\fsdttwofig} &
				RRDB \hspace{\fsdttwofig} &
				SRGAN
				\\
				\includegraphics[width=\widthscalefive \textwidth]{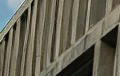} \hspace{\fsdttwofig} &
				\includegraphics[width=\widthscalefive \textwidth]{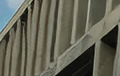} \hspace{\fsdttwofig} &
				\includegraphics[width=\widthscalefive \textwidth]{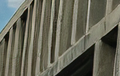} \hspace{\fsdttwofig} &
				\includegraphics[width=\widthscalefive \textwidth]{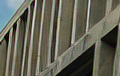} 
				\\ 
				ESRGAN \hspace{\fsdttwofig} &
				NatSR \hspace{\fsdttwofig} &
				SPSR-G \hspace{\fsdttwofig} &
				SPSR-P
				\\
			\end{tabular}
		\end{adjustbox}
		\vspace{0.5mm}
		\\

		\begin{adjustbox}{valign=t}
		\tiny
			\begin{tabular}{c}
				\includegraphics[width=0.25\textwidth]{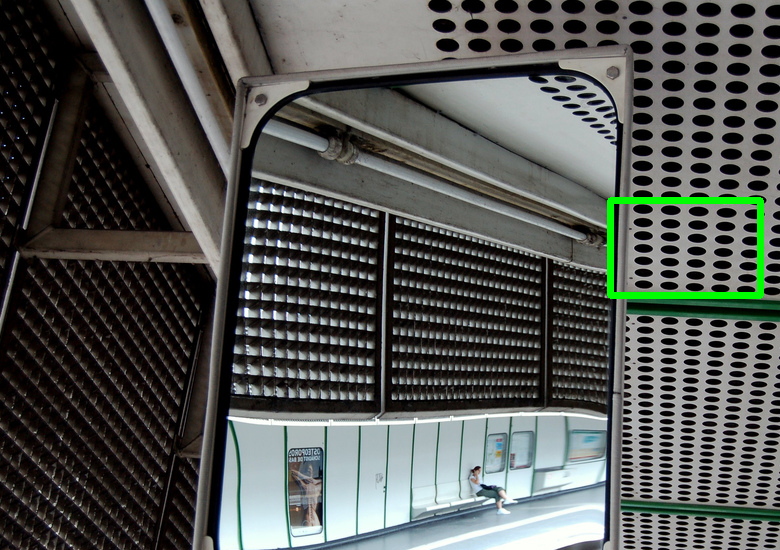}
				\\
				 `img\_004' from Urban100
				
			\end{tabular}
		\end{adjustbox}
		\hspace{-2.3mm}
		\begin{adjustbox}{valign=t}
		\tiny
			\begin{tabular}{ccccc}
				\includegraphics[width=\widthscalefive \textwidth]{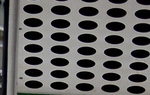} \hspace{\fsdttwofig} &
				\includegraphics[width=\widthscalefive \textwidth]{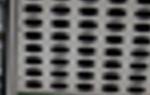} \hspace{\fsdttwofig} &
				\includegraphics[width=\widthscalefive \textwidth]{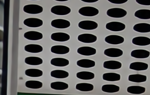} \hspace{\fsdttwofig} &
				\includegraphics[width=\widthscalefive \textwidth]{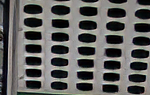} 
				\\
				HR \hspace{\fsdttwofig} &
				Bicubic \hspace{\fsdttwofig} &
				RRDB \hspace{\fsdttwofig} &
				SRGAN
				\\
				\includegraphics[width=\widthscalefive \textwidth]{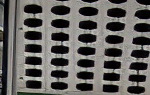} \hspace{\fsdttwofig} &
				\includegraphics[width=\widthscalefive \textwidth]{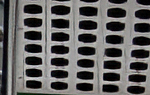} \hspace{\fsdttwofig} &
				\includegraphics[width=\widthscalefive \textwidth]{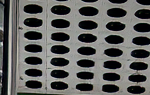} \hspace{\fsdttwofig} &
				\includegraphics[width=\widthscalefive \textwidth]{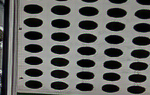} 
				\\ 
				ESRGAN \hspace{\fsdttwofig} &
				NatSR \hspace{\fsdttwofig} &
				SPSR-G \hspace{\fsdttwofig} &
				SPSR-P
				\\
			\end{tabular}
		\end{adjustbox}
		\vspace{0.5mm}
		\\

 		\begin{adjustbox}{valign=t}
		\tiny
			\begin{tabular}{c}
				\includegraphics[width=0.25\textwidth]{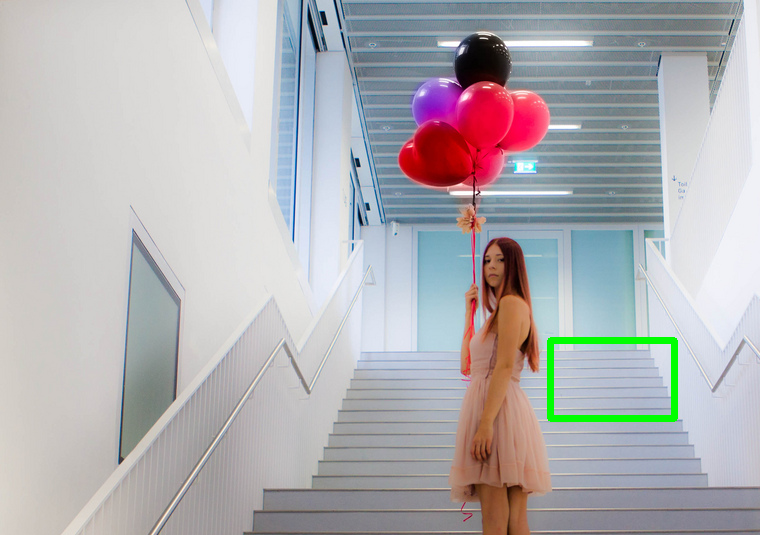}
				\\
				`img\_009' from Urban100
				
			\end{tabular}
		\end{adjustbox}
		\hspace{-2.3mm}
		\begin{adjustbox}{valign=t}
		\tiny
			\begin{tabular}{ccccc}
				\includegraphics[width=\widthscalefive \textwidth]{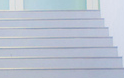} \hspace{\fsdttwofig} &
				\includegraphics[width=\widthscalefive \textwidth]{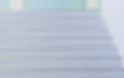} \hspace{\fsdttwofig} &
				\includegraphics[width=\widthscalefive \textwidth]{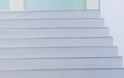} \hspace{\fsdttwofig} &
				\includegraphics[width=\widthscalefive \textwidth]{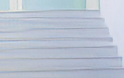} 
				\\
				HR \hspace{\fsdttwofig} &
				Bicubic \hspace{\fsdttwofig} &
				RRDB \hspace{\fsdttwofig} &
				SRGAN
				\\
				\includegraphics[width=\widthscalefive \textwidth]{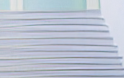} \hspace{\fsdttwofig} &
				\includegraphics[width=\widthscalefive \textwidth]{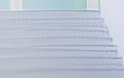} \hspace{\fsdttwofig} &
				\includegraphics[width=\widthscalefive \textwidth]{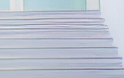} \hspace{\fsdttwofig} &
				\includegraphics[width=\widthscalefive \textwidth]{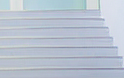} 
				\\ 
				ESRGAN \hspace{\fsdttwofig} &
				NatSR \hspace{\fsdttwofig} &
				SPSR-G \hspace{\fsdttwofig} &
				SPSR-P
				\\
			\end{tabular}
		\end{adjustbox}
 		\vspace{0.5mm}
 		\\
 		
		\begin{adjustbox}{valign=t}
		\tiny
			\begin{tabular}{c}
				\includegraphics[width=0.25\textwidth]{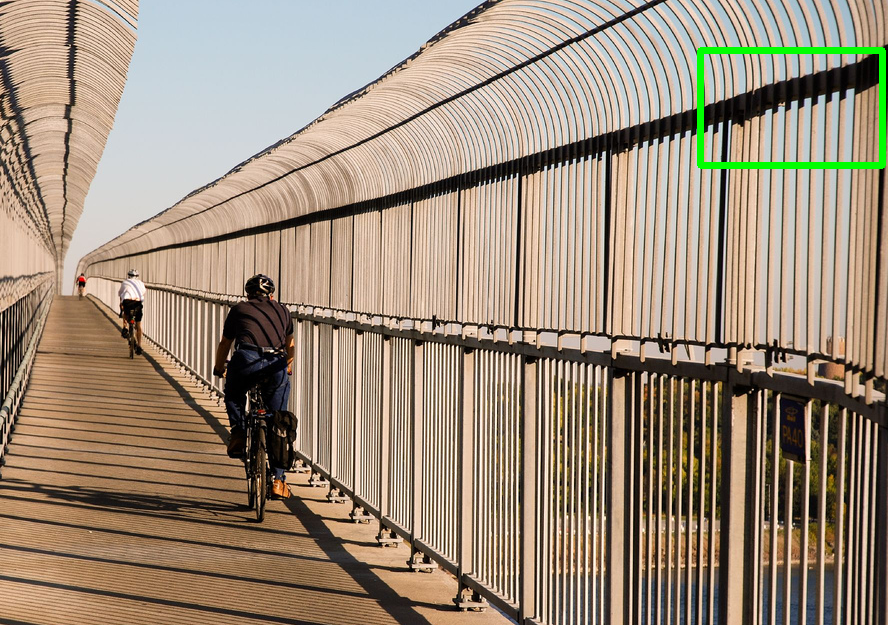}
				\\
				`img\_024' from Urban100
			\end{tabular}
		\end{adjustbox}
		\hspace{-2.3mm}
		\begin{adjustbox}{valign=t}
		\tiny
			\begin{tabular}{ccccc}
				\includegraphics[width=\widthscalefive \textwidth]{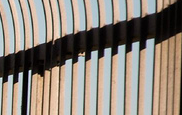} \hspace{\fsdttwofig} &
				\includegraphics[width=\widthscalefive \textwidth]{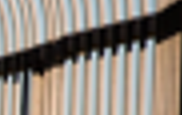} \hspace{\fsdttwofig} &
				\includegraphics[width=\widthscalefive \textwidth]{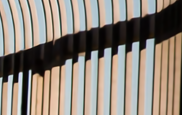} \hspace{\fsdttwofig} &
				\includegraphics[width=\widthscalefive \textwidth]{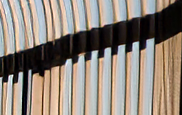} 
				\\
				HR \hspace{\fsdttwofig} &
				Bicubic \hspace{\fsdttwofig} &
				RRDB \hspace{\fsdttwofig} &
				SRGAN
				\\
				\includegraphics[width=\widthscalefive \textwidth]{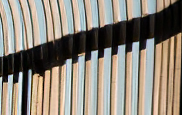} \hspace{\fsdttwofig} &
				\includegraphics[width=\widthscalefive \textwidth]{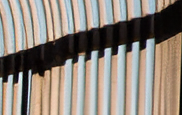} \hspace{\fsdttwofig} &
				\includegraphics[width=\widthscalefive \textwidth]{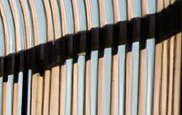} \hspace{\fsdttwofig} &
				\includegraphics[width=\widthscalefive \textwidth]{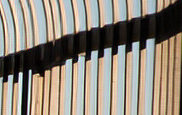} 
				\\ 
				ESRGAN \hspace{\fsdttwofig} &
				NatSR \hspace{\fsdttwofig} &
				SPSR-G \hspace{\fsdttwofig} &
				SPSR-P
				\\
			\end{tabular}
		\end{adjustbox}
		
	\end{tabular}\vspace{-2mm}
	\caption{
		Visual comparison with state-of-the-art SR methods. The results show that our proposed SPSR-G and SPSR-P methods significantly outperform other methods in structure restoration while generating perceptual pleasant SR images. Best viewed on screen.  
	}
\label{fig:best_result_4x}
\end{figure*} 

\begin{table*}
\centering
\caption{Comparison with state-of-the-art SR methods on benchmark datasets. The best performance among perceptual-driven SR methods is \textbf{highlighted} in \textcolor{red}{\textbf{red}} (1st best) and \textcolor{blue}{\textbf{blue}} (2nd best). Our SPSR methods obtain the best LPIPS values and comparable PSNR and SSIM values simultaneously. NatSR is more like a PSNR-oriented method since it has high PSNR and SSIM and relatively poor LPIPS performance.  } \label{quantitative:PSNRPI}
\vspace{-5px}
  \begin{tabular}{C{1.4cm}L{0.9cm}|C{1.2cm}C{1.2cm}|C{1.2cm}C{1.2cm}C{1.2cm}C{1.2cm}C{1.2cm}C{1.2cm}C{1.2cm}}
      \Xhline{1.0pt}
      \textbf{Dataset} & \textbf{Metric} & \textbf{Bicubic} & \textbf{RRDB} & \textbf{SFTGAN} & \textbf{SRGAN} & \textbf{ESRGAN} & \textbf{NatSR} & \textbf{SPSR-G} & \textbf{SPSR-J} & \textbf{SPSR-P} \\
      \Xhline{1.0pt}
      \multirow{3}*{\textbf{Set5}}
      & LPIPS & 0.3407 & 0.1684 & 0.0890  & 0.0882  & 0.0748  & 0.0939  & 0.0644 & \textcolor{blue}{\textbf{0.0614}} &\textcolor{red}{\textbf{0.0591}} \\
      & PSNR & 28.420 & 32.702 & 29.932  & 29.168  & 30.454  & 30.991  & 30.400 & \textcolor{blue}{\textbf{30.995}} & \textcolor{red}{\textbf{31.036}} \\ 
      & SSIM & 0.8245 & 0.9123 & 0.8665  & 0.8613  & 0.8677  & \textcolor{red}{\textbf{0.8800}}  & 0.8627 & \textcolor{blue}{\textbf{0.8773}} & 0.8772 \\  \hline
      \multirow{3}*{\textbf{Set14}}
      & LPIPS & 0.4393 & 0.2710 & 0.1481  & 0.1663  & 0.1329  & 0.1758  & 0.1318 & \textcolor{blue}{\textbf{0.1272}} & \textcolor{red}{\textbf{0.1257}} \\
      & PSNR & 26.100 & 28.953 & 26.223  & 26.171  & 26.276  & \textcolor{red}{\textbf{27.514}}  & 26.640 & 27.027 & \textcolor{blue}{\textbf{27.067}} \\ 
      & SSIM & 0.7850 & 0.8588 & 0.7854  & 0.7841  & 0.7783  & \textcolor{red}{\textbf{0.8140}}  & 0.7930 & 0.8073 & \textcolor{blue}{\textbf{0.8076}} \\ \hline
      \multirow{3}*{\textbf{BSD100}} 
      & LPIPS & 0.5249 & 0.3572 & 0.1769  & 0.1980  & 0.1614  & 0.2114  & 0.1611 & \textcolor{red}{\textbf{0.1544}} & \textcolor{blue}{\textbf{0.1561}} \\
      & PSNR & 25.961 & 27.838  & 25.505  & 25.459  & 25.317  & \textcolor{red}{\textbf{26.445}}  & 25.505 & 25.975 & \textcolor{blue}{\textbf{26.048}} \\ 
      & SSIM & 0.6675 & 0.7448 & 0.6549  & 0.6485  & 0.6506  & \textcolor{red}{\textbf{0.6831}}  & 0.6576 & 0.6788 & \textcolor{blue}{\textbf{0.6818}} \\ \hline
      \multirow{3}*{\textbf{General100}}
      & LPIPS & 0.3528 & 0.1663 & 0.1030  & 0.1055  & 0.0879  & 0.1117  & 0.0863 & \textcolor{blue}{\textbf{0.0830}} & \textcolor{red}{\textbf{0.0820}} \\
      & PSNR & 28.018 & 32.049  & 29.026  & 28.575  & 29.412  & \textcolor{red}{\textbf{30.346}}  & 29.414 & 30.003 & \textcolor{blue}{\textbf{30.101}} \\ 
      & SSIM & 0.8282 & 0.9033 & 0.8508  & 0.8541  & 0.8546  & \textcolor{red}{\textbf{0.8721}}  & 0.8537 & 0.8680 & \textcolor{blue}{\textbf{0.8696}} \\ \hline
      \multirow{3}*{\textbf{Urban100}}
      & LPIPS & 0.4726 & 0.1958 & 0.1433  & 0.1551  & 0.1229  & 0.1500  & 0.1184 & \textcolor{blue}{\textbf{0.1171}} & \textcolor{red}{\textbf{0.1146}} \\
      & PSNR & 23.145 & 27.027 & 24.013  & 24.397  & 24.360  & \textcolor{red}{\textbf{25.464}}  & 24.799 & 25.099 & \textcolor{blue}{\textbf{25.228}} \\ 
      & SSIM & 0.9011 & 0.9700 & 0.9364  & 0.9381  & 0.9453  & 0.9505  & 0.9481 & \textcolor{blue}{\textbf{0.9517}} & \textcolor{red}{\textbf{0.9531}} \\ \hline
      \Xhline{1.0pt}
  \end{tabular}
\end{table*} 

\begin{figure*}[htbp]
   \setlength{\fsdttwofig}{-1.5mm}
   \scriptsize
   \centering
   
   \begin{tabular}{cc}

      \begin{adjustbox}{valign=t}
      \tiny
         \begin{tabular}{c}
            \includegraphics[width=0.25\textwidth]{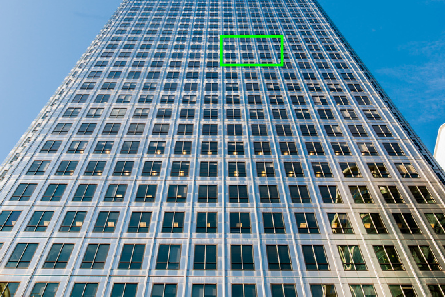}
            \\
            `img\_030' from Urban100
         \end{tabular}
      \end{adjustbox}
      \hspace{-2.3mm}
      \begin{adjustbox}{valign=t}
      \tiny
         \begin{tabular}{ccccc}
            \includegraphics[width=\widthscalefive \textwidth]{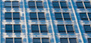} \hspace{\fsdttwofig} &
            \includegraphics[width=\widthscalefive \textwidth]{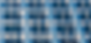} \hspace{\fsdttwofig} &
            \includegraphics[width=\widthscalefive \textwidth]{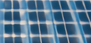} \hspace{\fsdttwofig} &
            \includegraphics[width=\widthscalefive \textwidth]{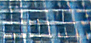}
            \\
            HR \hspace{\fsdttwofig} &
            Bicubic \hspace{\fsdttwofig} &
            RRDB  \hspace{\fsdttwofig} &
            SRGAN
            \\
            \includegraphics[width=\widthscalefive \textwidth]{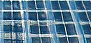} \hspace{\fsdttwofig} &
            \includegraphics[width=\widthscalefive \textwidth]{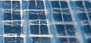} \hspace{\fsdttwofig} &
            \includegraphics[width=\widthscalefive \textwidth]{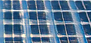} \hspace{\fsdttwofig} &
            \includegraphics[width=\widthscalefive \textwidth]{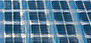} 
            \\ 
            ESRGAN \hspace{\fsdttwofig} &
            NatSR \hspace{\fsdttwofig} &
            SPSR-G \hspace{\fsdttwofig} &
            SPSR-P
            \\
         \end{tabular}
      \end{adjustbox}
      \\
      
      \begin{adjustbox}{valign=t}
      \tiny
         \begin{tabular}{c}
            \includegraphics[width=0.25\textwidth]{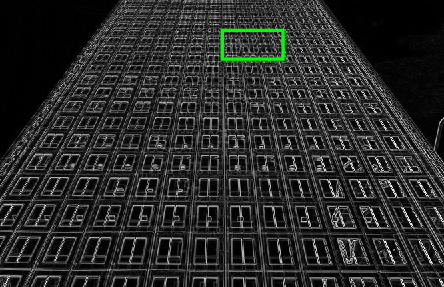}
            \\
             `im\_030' from Urban100
            
         \end{tabular}
      \end{adjustbox}
      \hspace{-2.3mm}
      \begin{adjustbox}{valign=t}
      \tiny
         \begin{tabular}{ccccc}
            \includegraphics[width=\widthscalefive \textwidth]{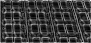} \hspace{\fsdttwofig} &
            \includegraphics[width=\widthscalefive \textwidth]{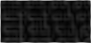} \hspace{\fsdttwofig} &
            \includegraphics[width=\widthscalefive \textwidth]{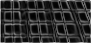} \hspace{\fsdttwofig} &
            \includegraphics[width=\widthscalefive \textwidth]{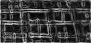}
            \\
            HR \hspace{\fsdttwofig} &
            Bicubic \hspace{\fsdttwofig} &
            RRDB  \hspace{\fsdttwofig} &
            SRGAN
            \\
            \includegraphics[width=\widthscalefive \textwidth]{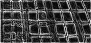} \hspace{\fsdttwofig} &
            \includegraphics[width=\widthscalefive \textwidth]{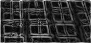} \hspace{\fsdttwofig} &
            \includegraphics[width=\widthscalefive \textwidth]{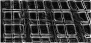} \hspace{\fsdttwofig} &
            \includegraphics[width=\widthscalefive \textwidth]{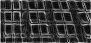}
            \\ 
            ESRGAN \hspace{\fsdttwofig} &
            NatSR \hspace{\fsdttwofig} &
            SPSR-G \hspace{\fsdttwofig} &
            SPSR-P
            \\
         \end{tabular}
      \end{adjustbox}
      
   \end{tabular}\vspace{-2mm}
   \caption{
      Comparison of SR results and gradient maps with state-of-the-art SR methods. The proposed SPSR methods can better preserve gradients and structures. Best viewed on screen. 
   }
\label{fig:best_result_4x_grad}
\end{figure*} 

\textbf{Training Details}: For SPSR-G, We use the architecture of ESRGAN~\cite{wang2018esrgan} as the backbone of our SR branch and the RRDB block~\cite{wang2018esrgan} as the gradient block. We randomly sample 15 $32\times32$ patches from LR images for each input mini-batch. Therefore the ground-truth HR patches have a size of $128\times128$. 
We use a pretrained PSNR-oriented RRDB~\cite{wang2018esrgan} model to initialize the parameters of the shared architectures of our model and RRDB. 
The pixelwise loss, perceptual loss, adversarial loss, and gradient loss are used as the optimizing objectives. A pre-trained 19-layer VGG network~\cite{simonyan2014very} is employed to calculate the feature distances in the perceptual loss. We also use a VGG-style network to perform discrimination. 
ADAM optimizor~\cite{Adam} with $\beta_1=0.9, \beta_2=0.999$ and $\epsilon=1\times10^{-8}$ is used for optimization. We set the learning rates to $1\times10^{-4}$ for both generator and discriminator, and reduce them to half at 50k, 100k, 200k, 300k iterations. As for the trade-off parameters of losses, we follow the settings in~\cite{wang2018esrgan} and set $\beta^I_{SR}$ and $\gamma^I_{SR}$ to $0.01$ and $0.005$, accordingly. Then we set the weights of gradient loss equal to those of image-space loss. Hence $\beta^{GM}_{SR}=0.01$ and $\gamma^{GM}_{SR}=0.005$. In terms of $\beta^{GM}_{GB}$, we set it to $0.5$ for better performance of gradient translation. 

To train the structure extractor by contrastive prediction, we set the initial learning rate to 0.001 and decrease it by a factor of 0.2 every 20 epochs. The hyper-parameter $\tau$ is set to 64 for stable convergence. We sample image patches of $420\times 420$ for training. Thus each positive structure feature on an image patch is corresponding to about 10000 negative features, which improves the representation capacity of the features. To solve jigsaw puzzles, we use a smaller learning rate of 0.0001 as initialization and decrease it every 30 epochs. For both tasks, the batch size is 48 while the channel number of extracted structure features is 32. For training SPSR-P and SPSR-J, the weights for the pixelwise structure loss and the adversarial structure loss are $\beta^{SF}_{SR}=10^{-7}$ and $\gamma^{SF}_{SR}=10^{-1}$, respectively. 
The experiments are implemented on PyTorch~\cite{paszke2019pytorch} with NVIDIA GTX 1080Ti GPUs.

\subsection{Investigation of NSEs}

We first display the experimental results of the structure extractors learned by contrastive prediction and solving jigsaw puzzles, \ie $F_P$ and $F_J$. After training on DIV2K, we get the two extractors and validate their abilities to capture structures on the five benchmark datasets. Specifically, for the contrastive prediction task, we densely sample $200\times 200$ patches from the testing images and extract the corresponding feature maps by two structure extractors. Then we randomly select a target position from each feature map and predict the corresponding feature vector according to its neighboring two horizontal input vectors by autoregressive models $P'_P$ and $P'_J$. These two models are finetuned by the contrastive prediction method on the training set with the parameters of $F_P$ and $F_J$ fixed. For each position, there is only one positive vector and around 2000 negative vectors as disturbances. The top-1 accuracy values of two extractors in 10 repeated testings are shown in the top figure of Fig.~\ref{fig:jigsaw}. Similarly, for the task of solving jigsaw puzzles, We densely sample $84\times 84$ patches and randomly rank the four vectors sampled by $2\times 2$ grids from the extracted structure features. The classification rates of $J'_P$ and $J'_J$ are displayed in the bottom figure of Fig.~\ref{fig:jigsaw}. We see that $F_P$ is able to perform well on not only the contrastive prediction task but also the jigsaw puzzle task. Contrastively, $F_J$ cannot present satisfactory performance on cross-validation of the contrastive prediction task. This indicates that $F_P$ has a stronger structure-capturing ability than $F_J$. 
The analyses of different sampling strategies for contrastive prediction are presented in the supplementary.

\subsection{Results and Analysis of SR}

\begin{table*}
\centering
\caption{Comparison of SPSR-G models with different components. The best results are \textbf{highlighted}. SPSR-G w/o GB has better LPIPS performance than ESRGAN in all benchmark datasets. SPSR-G surpasses ESRGAN on all the measurements in all the testing sets. } \label{table:Comparison}
 \vspace{-5px}
  \begin{tabular}{C{2.5cm}|C{1.1cm}C{1.1cm}C{1.1cm}|C{1.1cm}C{1.1cm}C{1.1cm}|C{1.1cm}C{1.1cm}C{1.1cm}}
  \Xhline{1.0pt}
  \multirow{2}*{\textbf{Method}} & 
  \multicolumn{3}{c|}{\textbf{Set14}} & 
  \multicolumn{3}{c|}{\textbf{BSD100}} & 
  \multicolumn{3}{c}{\textbf{Urban100}} \\
  &
  \textbf{LPIPS} &
  \textbf{PSNR} &
  \textbf{SSIM} &
  \textbf{LPIPS} &
  \textbf{PSNR} &
  \textbf{SSIM} &
  \textbf{LPIPS} &
  \textbf{PSNR} &
  \textbf{SSIM} \\
      \Xhline{1.0pt}
      \multirow{1}*{\textbf{ESRGAN}~\cite{wang2018esrgan}} & 
      \multirow{1}*{0.1329}  & 
      \multirow{1}*{26.276}  & 
      \multirow{1}*{0.778}  & 
      \multirow{1}*{0.1614}  & 
      \multirow{1}*{25.317}  & 
      \multirow{1}*{0.651}  & 
      \multirow{1}*{0.1229}   & 
      \multirow{1}*{24.360}  & 
      \multirow{1}*{0.945}  \\ 
      \hline
      \multirow{1}*{\textbf{SPSR-G w/o GB}} & 
      \multirow{1}*{0.1320} & 
      \multirow{1}*{26.027}  & 
      \multirow{1}*{0.785}  & 
      \multirow{1}*{\textbf{0.1604}}  & 
      \multirow{1}*{25.376}  & 
      \multirow{1}*{\textbf{0.659}}  & 
      \multirow{1}*{0.1188}  &
      \multirow{1}*{23.939}  & 
      \multirow{1}*{0.940}  \\ 
      \hline
      \multirow{1}*{\textbf{SPSR-G w/o GL}} & 
      \multirow{1}*{0.1335}  & 
      \multirow{1}*{26.547}  & 
      \multirow{1}*{\textbf{0.794}}  & 
      \multirow{1}*{0.1611}  & 
      \multirow{1}*{25.214}  & 
      \multirow{1}*{0.647}  & 
      \multirow{1}*{0.1195}  & 
      \multirow{1}*{24.309}  & 
      \multirow{1}*{0.942}  \\ 
      \hline
      \multirow{1}*{\textbf{SPSR-G w/o AGL}} & 
      \multirow{1}*{0.1364}  & 
      \multirow{1}*{26.603}  & 
      \multirow{1}*{0.789}  & 
      \multirow{1}*{0.1624}  & 
      \multirow{1}*{\textbf{25.573}} & 
      \multirow{1}*{0.657}  & 
      \multirow{1}*{0.1194}  &
      \multirow{1}*{\textbf{24.827}}  & 
      \multirow{1}*{0.948} \\ 
      \hline
      \multirow{1}*{\textbf{SPSR-G w/o PGL}} & 
      \multirow{1}*{0.1341}  & 
      \multirow{1}*{26.341}  & 
      \multirow{1}*{0.783}  & 
      \multirow{1}*{0.1619}  & 
      \multirow{1}*{25.449}  & 
      \multirow{1}*{0.654}  & 
      \multirow{1}*{0.1222}  &
      \multirow{1}*{24.515}  & 
      \multirow{1}*{0.943} \\ 
      \hline
      \multirow{1}*{\textbf{SPSR-G}} & 
      \multirow{1}*{\textbf{0.1318}}  & 
      \multirow{1}*{\textbf{26.640}}  & 
      \multirow{1}*{0.793}  & 
      \multirow{1}*{0.1611}  & 
      \multirow{1}*{25.505}  & 
      \multirow{1}*{0.658}  & 
      \multirow{1}*{\textbf{0.1184}}  &
      \multirow{1}*{24.799}  & 
      \multirow{1}*{\textbf{0.948}} \\ 
      \hline
\Xhline{1.0pt}
  \end{tabular}
\vspace{-3mm}
\end{table*} 

\textbf{Quantitative Comparison}: We compare our method quantitatively with state-of-the-art perceptual-driven SR methods including SFTGAN~\cite{SFTGAN}, SRGAN~\cite{ledig2017photo}, ESRGAN~\cite{wang2018esrgan} and NatSR~\cite{soh2019natural}. Results of LPIPS, PSNR, and SSIM values are presented in TABLE~\ref{quantitative:PSNRPI}. In each row, the best result is highlighted in red while the second best is in blue. We can see in all the testing datasets our SPSR methods achieve better LPIPS performance than other methods. Meanwhile, our methods get higher PSNR and SSIM values in most datasets than other methods except NatSR. It is noteworthy that while NatSR gets the highest PSNR and SSIM values in most datasets, our methods surpass NatSR by a large margin in terms of LPIPS. Moreover, NatSR cannot achieve the second-best LPIPS values in any testing set. Thus NatSR is more like a PSNR-oriented SR method, which tends to produce relatively blurry results with high PSNR compared to other perceptual-driven methods. Besides, we get better performance than ESRGAN
which uses the same basic reconstruction block as our models. 
Therefore, the results demonstrate the superior ability of our SPSR method to obtain excellent perceptual quality and minor distortions simultaneously. By comparing SPSR-G, SPSR-J, and SPSR-P, we see that although the original gradient-space losses and image-space adversarial loss are removed, SPSR-J and SPSR-P both present much better performance than SPSR-G, which demonstrates the efficacy of the proposed neural structure extractors and the structure loss. Besides, SPSR-P is more powerful than SPSR-J, which is corresponding to the conclusions obtained by Fig.~\ref{fig:jigsaw}: the extractor trained by contrastive prediction has a stronger ability to preserve geometric structures than that trained by solving jigsaw puzzles. Hence in the following sections, we mainly analyze the method of SPSR-P.

\textbf{Qualitative Comparison}: We also perform visual comparisons to perceptual-driven SR methods. From Fig.~\ref{fig:best_result_4x} we see that our results are more natural and realistic than other methods. For the first image, SPSR-G and SPSR-P infer sharp edges of the ovals properly, indicating that our methods are capable of capturing structural characteristics of objects in images. In other rows, our method also recovers better textures than the compared SR methods. The structures in our results are clear without severe distortions, while other methods fail to show satisfactory appearance for the objects. 
We also visualize some SR results and the corresponding gradient maps, as shown in Fig.~\ref{fig:best_result_4x_grad}. We can see the gradient maps of other methods tend to have small values or contain structure degradation while ours are bold and natural. 
The qualitative comparison proves that our proposed SPSR methods can learn more structure information from the gradient maps and structure features, which helps generate photo-realistic SR images by preserving geometric structures.

\begin{figure}[t]
    \centering
    \includegraphics[width=0.95\linewidth]{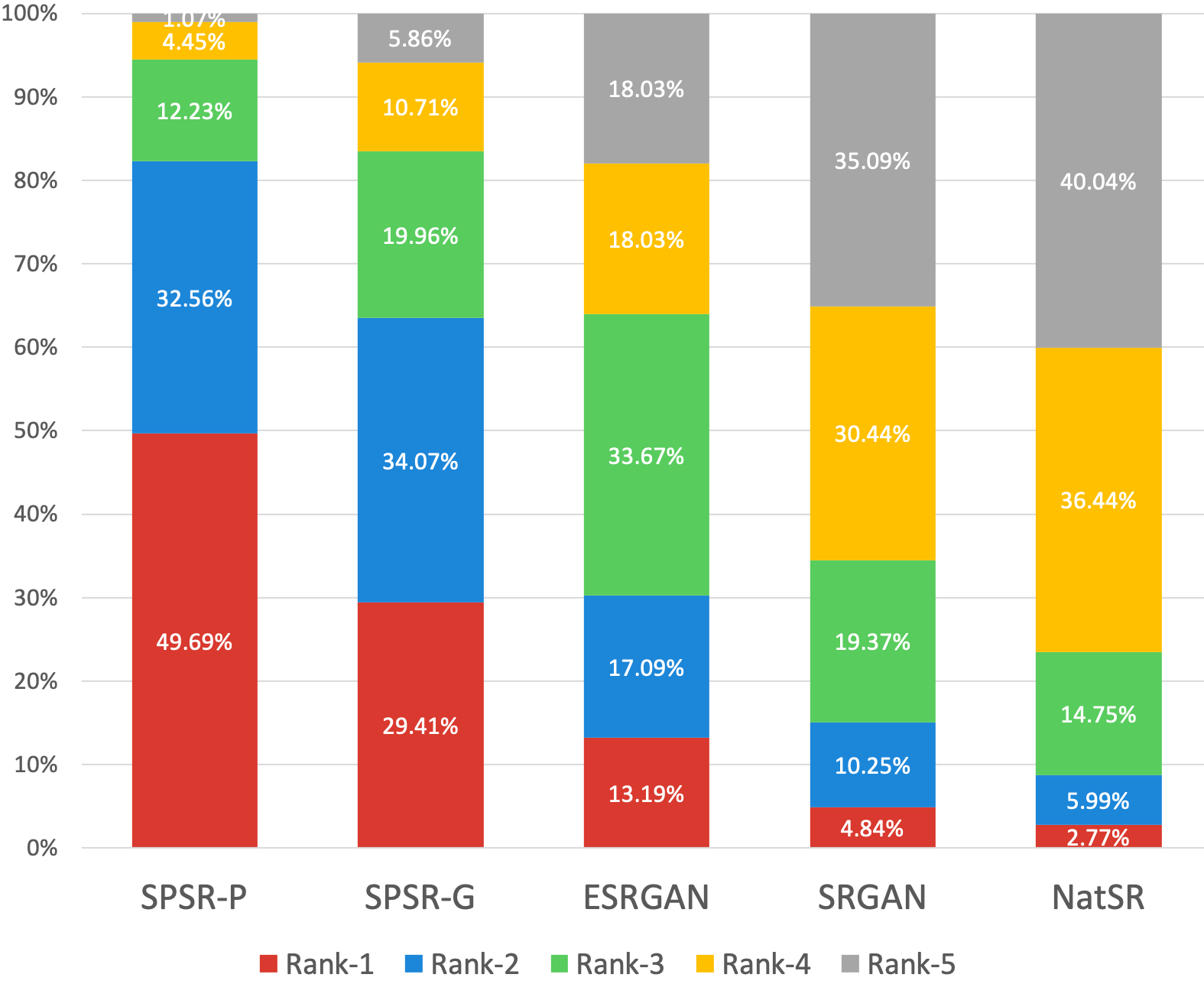}
    \vspace{-3mm}
    \caption{User study results of different GAN-based SR methods. Our SPSR-P and SPSR-G methods outperform state-of-the-art SR methods in generating high-quality images. }
    \label{fig:user-study}
    \vspace{-3mm}
\end{figure}

\begin{figure}[t]
\centering

\subfigure[HR]{
\begin{minipage}[b]{0.48\linewidth}
\includegraphics[width=1 \linewidth]{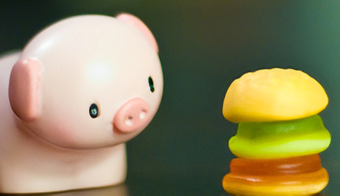}
\end{minipage}%
}%
\subfigure[HR gradiant]{
\begin{minipage}[b]{0.48\linewidth}
\includegraphics[width=1 \linewidth]{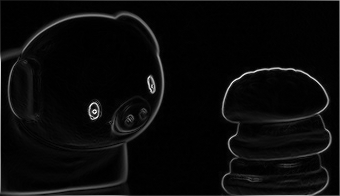}
\end{minipage}%
}%
\vspace{-3mm}

\subfigure[LR gradiant (Bicubic)]{
\begin{minipage}[b]{0.48\linewidth}
\includegraphics[width=1 \linewidth]{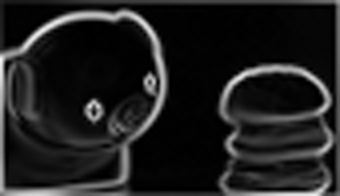}
\end{minipage}%
}%
\subfigure[Output of the gradiant branch]{
\begin{minipage}[b]{0.48\linewidth}
\includegraphics[width=1 \linewidth]{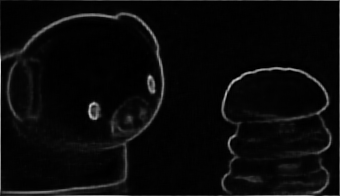}
\end{minipage}%
}%
\vspace{-3mm}

\centering
\caption{Visualization of gradient maps (`im\_073' from General100). The HR gradient map has thin outlines while those in the LR gradient map are thick. Our gradient branch is able to recover HR gradient maps with pleasant structures. }
\label{fig:branch_result_4x}
\vspace{-3mm}
\end{figure}

\begin{table*}
\centering
\caption{Comparison of SPSR-P models with different components. The best results are \textbf{highlighted}. The complete structure loss improves perceptual quality and pixelwise similarity of SR results simultaneously. } \label{table:spsrp_ablation}
 \vspace{-5px}
  \begin{tabular}{C{2.5cm}|C{1.1cm}C{1.1cm}C{1.1cm}|C{1.1cm}C{1.1cm}C{1.1cm}|C{1.1cm}C{1.1cm}C{1.1cm}}
  \Xhline{1.0pt}
  \multirow{2}*{\textbf{Method}} & 
  \multicolumn{3}{c|}{\textbf{Set14}} & 
  \multicolumn{3}{c|}{\textbf{BSD100}} & 
  \multicolumn{3}{c}{\textbf{Urban100}} \\
  &
  \textbf{LPIPS} &
  \textbf{PSNR} &
  \textbf{SSIM} &
  \textbf{LPIPS} &
  \textbf{PSNR} &
  \textbf{SSIM} &
  \textbf{LPIPS} &
  \textbf{PSNR} &
  \textbf{SSIM} \\
      \Xhline{1.0pt}
      \multirow{1}*{\textbf{SPSR-P}} & 
      \multirow{1}*{\textbf{0.1257}}  & 
      \multirow{1}*{27.067}  & 
      \multirow{1}*{0.8076}  & 
      \multirow{1}*{0.1561}  & 
      \multirow{1}*{26.048}  & 
      \multirow{1}*{0.6818}  & 
      \multirow{1}*{0.1146}   & 
      \multirow{1}*{25.228}  & 
      \multirow{1}*{0.9531}  \\ 
      \hline
      \multirow{1}*{\textbf{SPSR-P w/o ASL}} & 
      \multirow{1}*{0.1270} & 
      \multirow{1}*{\textbf{27.576}}  & 
      \multirow{1}*{\textbf{0.8232}}  & 
      \multirow{1}*{0.1675}  & 
      \multirow{1}*{\textbf{26.564}}  & 
      \multirow{1}*{\textbf{0.6982}}  & 
      \multirow{1}*{0.1171}  &
      \multirow{1}*{\textbf{25.585}}  & 
      \multirow{1}*{\textbf{0.9567}}  \\ 
      \hline
      \multirow{1}*{\textbf{SPSR-P w/o PSL}} & 
      \multirow{1}*{0.1269}  & 
      \multirow{1}*{26.924}  & 
      \multirow{1}*{0.8067}  & 
      \multirow{1}*{\textbf{0.1535}}  & 
      \multirow{1}*{25.918}  & 
      \multirow{1}*{0.6754}  & 
      \multirow{1}*{\textbf{0.1140}}  & 
      \multirow{1}*{25.163}  & 
      \multirow{1}*{0.9531}  \\ 
      \hline
      \multirow{1}*{\textbf{SPSR-P w/o GB}} & 
      \multirow{1}*{0.1261}  & 
      \multirow{1}*{27.064}  & 
      \multirow{1}*{0.8091}  & 
      \multirow{1}*{0.1591}  & 
      \multirow{1}*{26.079}  & 
      \multirow{1}*{0.6821}  & 
      \multirow{1}*{0.1173}  & 
      \multirow{1}*{25.131}  & 
      \multirow{1}*{0.9526}  \\ 
      \hline
      \multirow{1}*{\textbf{SPSR-P w/o SL}} & 
      \multirow{1}*{0.1332}  & 
      \multirow{1}*{27.354}  & 
      \multirow{1}*{0.8175}  & 
      \multirow{1}*{0.1765}  & 
      \multirow{1}*{26.458}  & 
      \multirow{1}*{0.6949}  & 
      \multirow{1}*{0.1221}  & 
      \multirow{1}*{25.383}  & 
      \multirow{1}*{0.9544}  \\ 
      \hline
\Xhline{1.0pt}
  \end{tabular}
\vspace{-3mm}
\end{table*}

\begin{table*}
\centering
\caption{Experimental results of parameter sensitivity for the SPSR-P model. For the sake of simplicity, we use $\gamma$ and $\beta$ to represent $\gamma^{SF}_{SR}$ and $\beta^{SF}_{SR}$ respectively in the following table. The original SPSR-P model is trained with $\gamma^{SF}_{SR}=10^{-1}$ and $\beta^{SF}_{SR}=10^{-7}$. The best results are \textbf{highlighted}. } \label{table:param_sens}
 \vspace{-5px}
  \begin{tabular}{C{2.5cm}|C{1.1cm}C{1.1cm}C{1.1cm}|C{1.1cm}C{1.1cm}C{1.1cm}|C{1.1cm}C{1.1cm}C{1.1cm}}
  \Xhline{1.0pt}
  \multirow{2}*{\textbf{Parameters}} & 
  \multicolumn{3}{c|}{\textbf{Set14}} & 
  \multicolumn{3}{c|}{\textbf{BSD100}} & 
  \multicolumn{3}{c}{\textbf{Urban100}} \\
  &
  \textbf{LPIPS} &
  \textbf{PSNR} &
  \textbf{SSIM} &
  \textbf{LPIPS} &
  \textbf{PSNR} &
  \textbf{SSIM} &
  \textbf{LPIPS} &
  \textbf{PSNR} &
  \textbf{SSIM} \\
      \Xhline{1.0pt}
      \multirow{1}*{\textbf{$\gamma=1$}} & 
      \multirow{1}*{0.1280}  & 
      \multirow{1}*{26.978}  & 
      \multirow{1}*{0.8063}  & 
      \multirow{1}*{0.1544}  & 
      \multirow{1}*{25.849}  & 
      \multirow{1}*{0.6743}  & 
      \multirow{1}*{0.1172}   & 
      \multirow{1}*{25.182}  & 
      \multirow{1}*{0.9523}  \\
      \hline
      \multirow{1}*{\textbf{$\gamma=10^{-2}$}} & 
      \multirow{1}*{\textbf{0.1251}}  & 
      \multirow{1}*{26.870}  & 
      \multirow{1}*{0.8054}  & 
      \multirow{1}*{0.1517}  & 
      \multirow{1}*{25.815}  & 
      \multirow{1}*{0.6720}  & 
      \multirow{1}*{0.1158}  & 
      \multirow{1}*{25.016}  & 
      \multirow{1}*{0.9515}  \\ 
      \hline
      \multirow{1}*{\textbf{$\gamma=10^{-3}$}} & 
      \multirow{1}*{0.1253}  & 
      \multirow{1}*{26.890}  & 
      \multirow{1}*{0.8034}  & 
      \multirow{1}*{\textbf{0.1498}}  & 
      \multirow{1}*{25.817}  & 
      \multirow{1}*{0.6723}  & 
      \multirow{1}*{\textbf{0.1126}}  & 
      \multirow{1}*{25.107}  & 
      \multirow{1}*{0.9526}  \\ 
      \hline
      \multirow{1}*{\textbf{$\beta=10^{-5}$}} & 
      \multirow{1}*{0.1303}  & 
      \multirow{1}*{\textbf{27.145}}  & 
      \multirow{1}*{0.8105}  & 
      \multirow{1}*{0.1605}  & 
      \multirow{1}*{26.026}  & 
      \multirow{1}*{0.6815}  & 
      \multirow{1}*{0.1187}  & 
      \multirow{1}*{25.197}  & 
      \multirow{1}*{0.9533}  \\ 
      \hline
      \multirow{1}*{\textbf{$\beta=10^{-6}$}} & 
      \multirow{1}*{0.1278}  & 
      \multirow{1}*{26.936}  & 
      \multirow{1}*{0.8071}  & 
      \multirow{1}*{0.1555}  & 
      \multirow{1}*{25.912}  & 
      \multirow{1}*{0.6782}  & 
      \multirow{1}*{0.1162}  & 
      \multirow{1}*{25.166}  & 
      \multirow{1}*{\textbf{0.9533}}  \\ 
      \hline
      \multirow{1}*{\textbf{$\beta=10^{-8}$}} & 
      \multirow{1}*{0.1298}  & 
      \multirow{1}*{27.072}  & 
      \multirow{1}*{\textbf{0.8092}}  & 
      \multirow{1}*{0.1612}  & 
      \multirow{1}*{\textbf{26.132}}  & 
      \multirow{1}*{0.6816}  & 
      \multirow{1}*{0.1178}  & 
      \multirow{1}*{25.216}  & 
      \multirow{1}*{0.9530}  \\ 
      \hline
      \multirow{1}*{\textbf{SPSR-P}} & 
      \multirow{1}*{0.1257} & 
      \multirow{1}*{27.067}  & 
      \multirow{1}*{0.8076}  & 
      \multirow{1}*{0.1561}  & 
      \multirow{1}*{26.048}  & 
      \multirow{1}*{\textbf{0.6818}}  & 
      \multirow{1}*{0.1146}  &
      \multirow{1}*{\textbf{25.228}}  & 
      \multirow{1}*{0.9531}  \\ 
      \hline
\Xhline{1.0pt}
  \end{tabular}
\end{table*}

\begin{figure}[t]
\centering

\subfigure[Only the SR branch]{
\begin{minipage}[b]{0.48\linewidth}
\includegraphics[width=1 \linewidth]{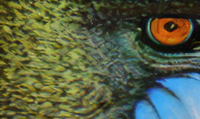}\vspace{1.8pt}
\includegraphics[width=1 \linewidth]{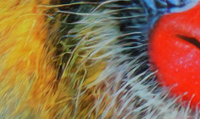}
\end{minipage}%
}%
\subfigure[Complete model]{
\begin{minipage}[b]{0.48\linewidth}
\includegraphics[width=1 \linewidth]{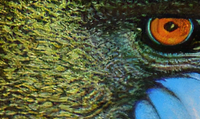}\vspace{1.8pt}
\includegraphics[width=1 \linewidth]{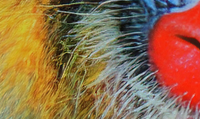}
\end{minipage}
}
\vspace{-5mm}

\centering
\caption{SR comparison of the SPSR-G models without and with the gradient branch (`baboon' from Set14). Images recovered by the complete model have clearer textures than those generated only by the features from the SR branch. }
\label{fig:branch_region_result_4x}
\vspace{-3mm}
\end{figure}

\textbf{User Study}: We conduct a user study as a subjective assessment to evaluate the visual performance of different SR methods on benchmark datasets. HR images are displayed as references while SR results of our SPSR methods, ESRGAN~\cite{wang2018esrgan}, NatSR~\cite{soh2019natural} and SRGAN~\cite{ledig2017photo} are presented in a randomized sequence. Human raters are asked to rank the five SR versions according to the perceptual quality. Finally, we collect 1590 votes from 53 human raters. The summarized results are presented in Fig.~\ref{fig:user-study}. As shown, our SPSR-P method gets more votes of rank-1 than SPSR-G, which demonstrates the effectiveness of the proposed NSE and the structure loss. Besides, both SPSR-P and SPSR-G perform better than ESRGAN, NatSR, and SRGAN. We see most SR results of ESRGAN are voted the third-best among the five methods since there are more structural distortions in the recovered images of ESRGAN than ours. NatSR and SRGAN fail to obtain satisfactory results. We think the reason is that they sometimes generate relatively blurry textures and undesirable artifacts. The comparison with the state-of-the-art GAN-based SR methods verifies the superiority of our proposed methods.

\textbf{Component Analysis of SPSR-G}: We conduct more experiments on different models to validate the necessity of each part in our proposed framework. 
Since we apply the architecture of ESRGAN~\cite{wang2018esrgan} in our SR branch, we use ESRGAN as a baseline. We compare five models with it: (1) SPSR-G w/o GB has the same architecture as ESRGAN without the gradient branch (GB) and is trained by the full loss, (2) SPSR-G w/o GL without the gradient loss (GL), (3) SPSR-G w/o AGL without the adversarial gradient loss (AGL), (4) SPSR-G w/o PGL without the pixelwise gradient loss (PGL), (5) our proposed SPSR-G model. Quantitative comparisons are presented in TABLE~\ref{table:Comparison}. It is observed that SPSR-G w/o GB has a significant enhancement on LPIPS over ESRGAN, which demonstrates the effectiveness of the proposed gradient loss in improving perceptual quality. Besides, the results of SPSR-G w/o GL also show that the gradient branch can help improve LPIPS or PSNR while relatively preserving the other one. We see SPSR-G w/o AGL performs better on PSNR and SSIM while SPSR-G w/o PGL performs  better on LPIPS. The results correspond to the fact that more pixelwise losses are imposed on SPSR-G w/o AGL while more adversarial losses are imposed on SPSR-G w/o PGL. By combining the two terms of losses, SPSR-G obtains the best SR performance among the compared models. Note that SPSR-G surpasses ESRGAN on all the measurements in all the testing sets. Therefore, the efficacy of our method is verified.

In order to validate the effectiveness of the gradient branch, we also visualize the output gradient maps as shown in Fig.~\ref{fig:branch_result_4x}. Given HR images with sharp edges, the extracted HR gradient maps may have thin and clear outlines for objects in the images. However, the gradient maps extracted from the LR counterparts commonly have thick lines after the bicubic upsampling. Our gradient branch takes LR gradient maps as inputs and produces HR gradient maps so as to provide explicit structural information as guidance for the SR branch. By treating gradient generation as an image translation problem, we can exploit the strong generative ability of the deep model. From the output gradient map in Fig.~\ref{fig:branch_result_4x} (d), we can see our gradient branch successfully recover thin and structure-pleasing gradient maps. 

We conduct another experiment to evaluate the effectiveness of the gradient branch. With a complete SPSR-G model, we remove the features from the gradient branch by setting them to $0$ and only use the SR branch for inference. The visualization results are shown in Fig.~\ref{fig:branch_region_result_4x}. From the patches, we can see the furs and whiskers super-resolved by only the SR branch are more blurry than those recovered by the complete model. The change of detailed textures reveals that the gradient branch can help produce sharp edges for better perceptual fidelity. 

\textbf{Component Analysis of SPSR-P}: In order to analyze the efficacy of each part in SPSR-P, we remove the adversarial structure loss (ASL), the pixelwise structure loss (PSL), the gradient branch (GB), and the whole structure losses (SL) to form the four models displayed in TABLE~\ref{table:spsrp_ablation}. We see that the adversarial structure loss is superior in improving the LPIPS performance while the pixelwise structure loss is mainly responsible for enhancing PSNR and SSIM results. After removing the gradient branch, we observe slight performance degradation, which indicates that the gradient branch can boost the SR capacity of SPSR-P, but the influence is not significant. When the whole structure losses are absent, the model performs poorly on LPIPS. Note that the PSNR and SSIM values of SPSR-P w/o SL are also lower than SPSR-P w/o ASL, which shows that the designed structure loss based on neural structure extractor is able to improve the performance on all the measurement metrics simultaneously. The complementary supervision of the adversarial and pixelwise structure losses can balance the trade-off between perceptual quality and pixelwise distortion. 

We also conduct experiments on parameter sensitivity for SPSR-P. The SPSR-P model is trained with $\gamma^{SF}_{SR}=10^{-1}$ and $\beta^{SF}_{SR}=10^{-7}$. Therefore, we change one of these two parameters to study their influence on SR performance. The results are shown in TABLE~\ref{table:param_sens}. 
From the table we see that the parameter values have a modest effect on the SR performance. It is inferred that our proposed model is not sensitive to the parameters.

\section{Conclusion}
In this paper, we have proposed a structure-preserving super-resolution method with gradient guidance (SPSR-G) to alleviate the issue of geometric distortions commonly existing in the SR results of perceptual-driven methods. 
We have preserved geometric structures of SR images by building a gradient branch to provide explicit structural guidance and proposing a new gradient loss to impose second-order restrictions. We have further extended SPSR-G to SPSR-P and SPSR-J by designing structure losses based on neural structure extractors trained by contrastive prediction and solving jigsaw puzzles. 
Geometric relationships can be better captured by the additional supervision provided by gradient maps and structure features. Quantitative and qualitative experimental results on five popular benchmarks have shown the effectiveness of our proposed method.

\ifCLASSOPTIONcompsoc
  \section*{Acknowledgments}
\else
  \section*{Acknowledgment}
\fi

This work was supported in part by the National Key Research and Development Program of China under Grant 2017YFA0700802, in part by the National Natural Science Foundation of China under Grant 61822603, Grant U1813218, Grant U1713214, and Grant 61672306, 
and in part by Tsinghua University Initiative Scientfic Research Program.


\appendices

\newcommand{\picwidth}{0.24}
\newcommand{\pagehspace}{-1pt}
\newcommand{\pagevspace}{2pt}
\newcommand{\picvspace}{-3pt}
\newcommand{\captionvspace}{0pt}
\newcommand{\packvspace}{6pt}

\begin{figure}
    \centering
    \includegraphics[width=0.8\linewidth]{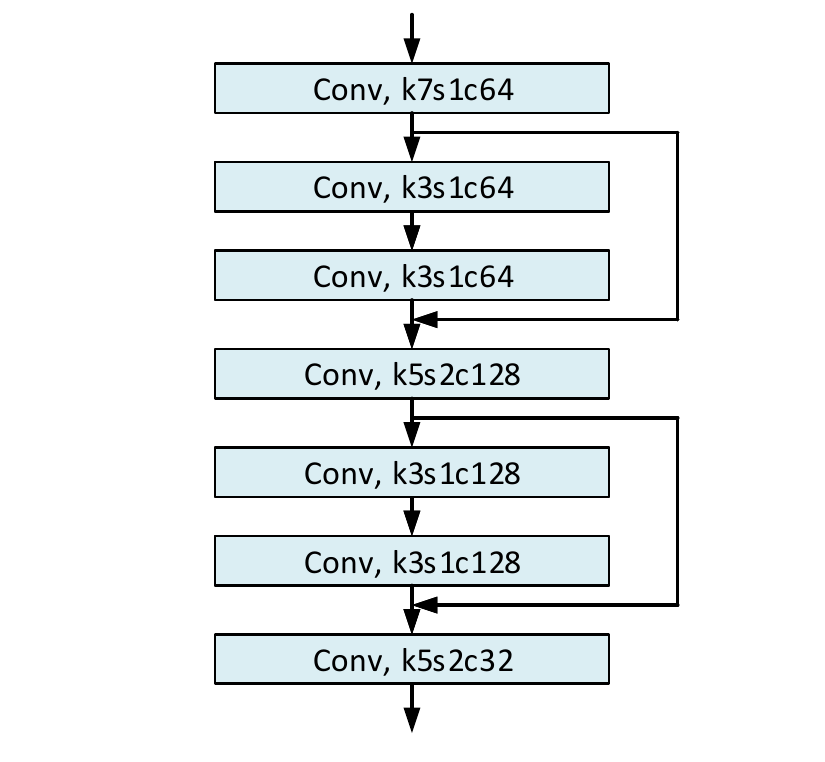}
    \vspace{-2mm}
    \caption{Network architecture of neural structure extractors. ``Conv, k3s1c64'' represents a convolutional layer with $kernel=3$, $stride=1$ and $channel\ number=64$. }
    \label{fig:network}
\end{figure}

\section{Architecture of Neural Structure Extractors}

The architecture of neural structure extractors is illustrated in Fig.~\ref{fig:network}. Since our goal is to capture local structures instead of semantic information by the extractors, we design a relatively small and shallow neural network with 6 convolutional layers followed by ReLU~\cite{nair2010rectified} layers. Each obtained feature vector has a receptive field of $31\times31$. We follow the ResNet~\cite{he2016deep} architecture and involve two skip connections to improve the representation ability of extracted features. 

\begin{figure}[t]
\centering

\subfigure[$F_{P-h}$]{
\begin{minipage}[b]{0.31\linewidth}
\includegraphics[width=1 \linewidth]{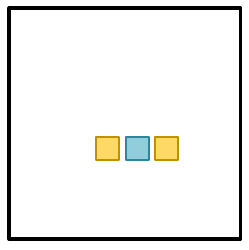}
\end{minipage}%
}
\subfigure[$F_{P-v}$]{
\begin{minipage}[b]{0.31\linewidth}
\includegraphics[width=1 \linewidth]{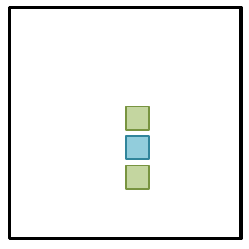}
\end{minipage}
}\hspace{-2pt}
\subfigure[$F_{P-c}$]{
\begin{minipage}[b]{0.31\linewidth}
\includegraphics[width=1 \linewidth]{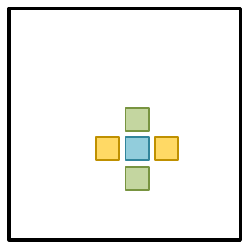}
\end{minipage}
}

\centering
\caption{Illustration of three sampling strategies for contrastive prediction: horizontal positions, vertical positions and crossing positions. The blue rectangle denotes the anchor position.  }
\label{fig:hvc}
\end{figure} 

\section{More Details of Contrasive Prediction}
\subsection{Analysis of Sampling Strategies}

\begin{figure}
    \centering
    \includegraphics[width=\linewidth]{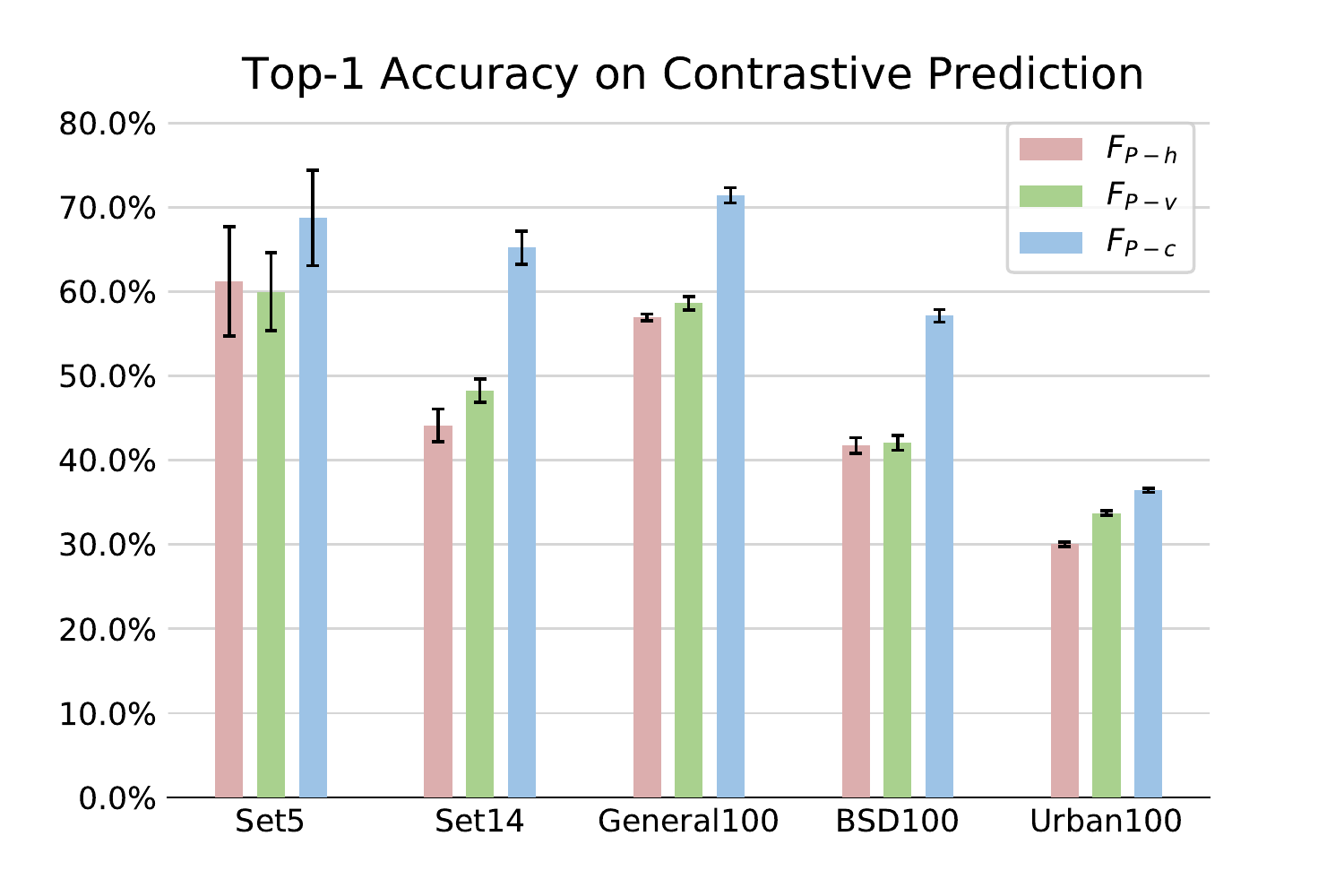}
    \vspace{-5mm}
    \caption{Validation results on the contrastive prediction task of three neural structure extractors trained by contrastive prediction with different sampling strategies. }
    \label{fig:contrastive}
\end{figure}

We conduct extensive experiments to investigate the effects of sampling strategies for training $F_P$, the neural structure extractor optimized by contrastive prediction. Besides sampling two horizontal positions near the anchor point, we also sample two vertical positions and four end positions of a neighboring cross, as illustrated in Fig.~\ref{fig:hvc}. We train three structure extractors, $F_{P-h}$, $F_{P-v}$ and $F_{P-c}$, by contrastive prediction with the above three sampling strategies and test the obtained models on five benchmark datasets. The testing results on contrastive prediction and the cross-validation accuracy for jigsaw puzzles are displayed in Fig.~\ref{fig:contrastive} and Fig.~\ref{fig:jigsaw}, respectively. From the figures, we see that $F_{P-v}$ has similar performance on structure capturing to $F_{P-h}$ while $F_{P-c}$ performs better than $F_{P-h}$ and $F_{P-v}$. Since $F_{P-h}$ and $F_{P-v}$ mainly focus on structures in one direction, $F_{P-c}$ may capture more comprehensive structural information in both horizontal and vertical directions than the other two extractors. We train three SR models, SPSR-P-h, SPSR-P-v, and SPSR-P-c by the corresponding neural structure extractors. The SR performance of the three SR models is shown in TABLE~\ref{quantitative:hvc}. We see that three models have similar SR capacity, which indicates that sampling strategy does not have a significant influence on SR performance although it is a key factor for the accuracy of the contrastive prediction and jigsaw puzzle tasks. 

\begin{table}
\centering
\caption{Performance comparison of SPSR-P models with different sampling strategies. The best results are \textbf{highlighted}. The three models present similar SR capacities on five benchmark datasets. } \label{quantitative:hvc}
\vspace{-5px}
  \begin{tabular}{C{1.4cm}L{0.9cm}|C{1.3cm}C{1.3cm}C{1.3cm}}
      \Xhline{1.0pt}
      \textbf{Dataset} & \textbf{Metric} & \textbf{SPSR-P-h} & \textbf{SPSR-P-v} & \textbf{SPSR-P-c} \\
      \Xhline{1.0pt}
      \multirow{4}*{\textbf{Set5}}
      & LPIPS & \textbf{0.0591} & 0.0646 & 0.0613 \\
      & PSNR & \textbf{31.036} & 31.027 & 30.907 \\ 
      & SSIM & 0.8772 & \textbf{0.8821} & 0.8772 \\  \hline
      \multirow{4}*{\textbf{Set14}}
      & LPIPS & \textbf{0.1257} & 0.1274 & 0.1269 \\
      & PSNR & \textbf{27.067} & 27.064 & 26.944 \\ 
      & SSIM & 0.8076 & \textbf{0.8092} & 0.8065 \\ \hline
      \multirow{4}*{\textbf{BSD100}}
      & LPIPS & 0.1561 & 0.1557 & \textbf{0.1548} \\
      & PSNR & \textbf{26.048} & 26.048 & 25.908 \\ 
      & SSIM & \textbf{0.6818} & 0.6809 & 0.6765 \\ \hline
      \multirow{4}*{\textbf{General100}}
      & LPIPS & 0.0820 & 0.0822 & \textbf{0.0817} \\
      & PSNR & 30.101 & \textbf{30.113} & 29.943 \\ 
      & SSIM & 0.8696 & \textbf{0.8705} & 0.8685 \\ \hline
      \multirow{4}*{\textbf{Urban100}}
      & LPIPS & \textbf{0.1146} & 0.1153 & 0.1166 \\
      & PSNR & \textbf{25.228} & 25.213 & 25.143 \\ 
      & SSIM & 0.9531 & \textbf{0.9534} & 0.9523 \\ \hline
      \Xhline{1.0pt}
  \end{tabular}
\end{table}

\begin{figure}
    \centering
    \includegraphics[width=\linewidth]{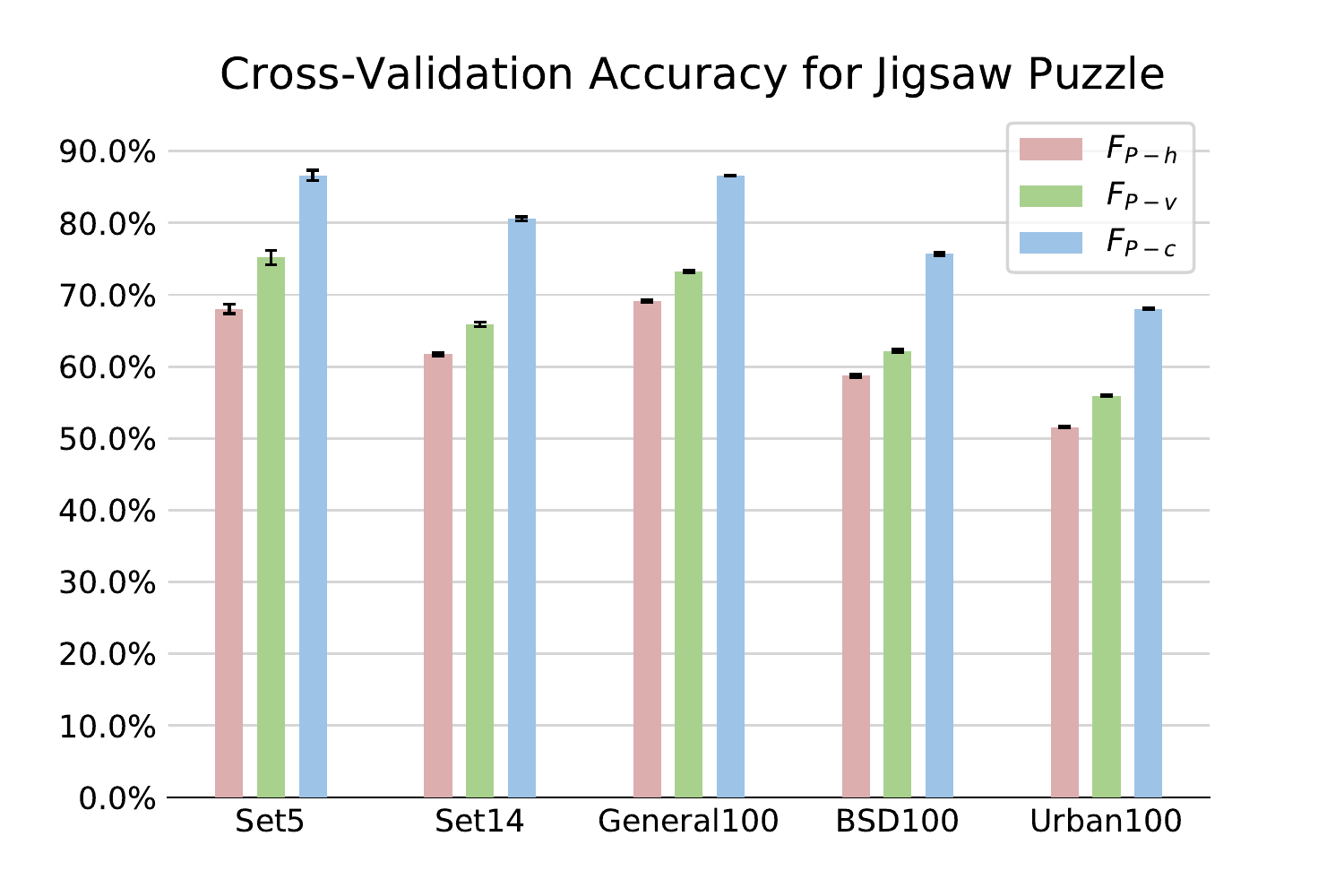}
    \vspace{-5mm}
    \caption{Cross-validation results on the jigsaw puzzle task of three neural structure extractors trained by contrastive prediction with different sampling strategies. }
    \label{fig:jigsaw}
\end{figure}

\subsection{Details for Evaluation}

Here we describe more details of evaluating NSEs by the top-1 accuracy on the contrastive prediction task. After training the NSEs, we can obtain a feature map for an input testing image by the extractors. We randomly select an anchor position $X_i$ for evaluation and feed the feature vectors of the anchor's neighboring positions into an autoregressive model $P$, similar to the operation in training. 
As mentioned in the main paper, this $P$ for evaluation is obtained by fixing the parameters of the corresponding NSE and optimizing $P$ by the contrastive prediction task on the training set. 
Since $P$ is trained to predict the feature vector of $X_i$, the output feature $p_i$ of $P$ should be the same as $f_{X_i}$. 
Hence only $f_{X_i}$ is treated as the positive vector while the other vectors on the feature map whose receptive fields have no overlap with the above-mentioned positive  feature and input features  are negative samples. If $p_i$ has the largest cosine similarity to $f_{X_i}$ among the positive and negative feature vectors, we regard the top-1 accuracy for the current anchor as 100\%. Otherwise, the accuracy is regarded as 0\%. After densely sampling image patches from the testing sets and randomly sampling anchors on the patches for multiple times, we average the obtained accuracy values and get the final top-1 accuracy.

\begin{table*}
\small
\centering
\caption{Comparison of SPSR models trained with different NSEs. The best performance is \textbf{highlighted} in \textcolor{red}{\textbf{red}} (1st best) and \textcolor{blue}{\textbf{blue}} (2nd best). } \label{quantitative:multispsr}
  \begin{tabular}{C{2cm}L{1.5cm}|C{2cm}C{2cm}C{2cm}C{2cm}C{2cm}}
      \Xhline{1.0pt}
      \textbf{Dataset} & \textbf{Metric} & \textbf{SPSR-VGG} & \textbf{SPSR-ResNet} & \textbf{SPSR-R} & \textbf{SPSR-J} & \textbf{SPSR-P} \\
      \Xhline{1.0pt}
      \multirow{3}*{\textbf{Set5}}
      & LPIPS & 0.0696 & 0.0743 & 0.06629 & \textcolor{blue}{\textbf{0.0614}} &\textcolor{red}{\textbf{0.0591}} \\
      & PSNR & 30.273 & 30.501 & 30.751 & \textcolor{blue}{\textbf{30.995}} & \textcolor{red}{\textbf{31.036}} \\ 
      & SSIM & 0.8714 & 0.8721 & 0.8706 & \textcolor{red}{\textbf{0.8773}} & \textcolor{blue}{\textbf{0.8772}} \\  \hline
      \multirow{3}*{\textbf{Set14}}
      & LPIPS & 0.1438 & 0.1489 & 0.1408 &  \textcolor{blue}{\textbf{0.1272}} & \textcolor{red}{\textbf{0.1257}} \\
      & PSNR & 26.658 & 26.894 & 26.672 & \textcolor{blue}{\textbf{27.027}} & \textcolor{red}{\textbf{27.067}} \\ 
      & SSIM  & 0.8037 & 0.8098 & 0.7983 & \textcolor{blue}{\textbf{0.8073}} & \textcolor{red}{\textbf{0.8076}} \\ \hline
      \multirow{3}*{\textbf{BSD100}}
      & LPIPS & 0.1746 & 0.1765 & 0.1631 & \textcolor{red}{\textbf{0.1544}} & \textcolor{blue}{\textbf{0.1561}} \\
      & PSNR & \textcolor{blue}{\textbf{26.005}} & 25.966 & 25.914 & 25.975 & \textcolor{red}{\textbf{26.048}} \\ 
      & SSIM  & \textcolor{blue}{\textbf{0.6805}} & 0.6801 & 0.6737 & 0.6788 & \textcolor{red}{\textbf{0.6818}} \\ \hline
      \multirow{3}*{\textbf{General100}}
      & LPIPS & 0.0896 & 0.0989 & 0.0874 &  \textcolor{blue}{\textbf{0.0830}} & \textcolor{red}{\textbf{0.0820}} \\
      & PSNR & 29.409 & 29.518 & 29.625 &  \textcolor{blue}{\textbf{30.003}} & \textcolor{red}{\textbf{30.101}} \\ 
      & SSIM & 0.8645 & 0.8677 & 0.8636 & \textcolor{blue}{\textbf{0.8680}} & \textcolor{red}{\textbf{0.8696}} \\ \hline
      \multirow{3}*{\textbf{Urban100}}
      & LPIPS & 0.1359 & 0.1271 & 0.1231 &  \textcolor{blue}{\textbf{0.1171}} & \textcolor{red}{\textbf{0.1146}} \\
      & PSNR  & 24.652 & 25.056 & 24.67 & \textcolor{blue}{\textbf{25.099}} & \textcolor{red}{\textbf{25.228}} \\ 
      & SSIM & 0.9483 & 0.9517 & 0.9481 &  \textcolor{blue}{\textbf{0.9517}} & \textcolor{red}{\textbf{0.9531}} \\ \hline
      \Xhline{1.0pt}
  \end{tabular}
\end{table*} 

\section{Experiments on Different NSEs}

In order to validate the effectiveness of the proposed self-supervised structure learning methods,  we have conducted the experiments where the proposed neural structure extractor is replaced by the VGG19~\cite{simonyan2014very} and ResNet18~\cite{he2016deep} models pretrained on the ImageNet dataset~\cite{krizhevsky2012imagenet}. 
In our experiments, we follow ESRGAN~\cite{wang2018esrgan} and use the features of the 4th convolution after the 4th max-pooling as the output of the VGG-19 model. The output of the ResNet structure is from the `Conv4\_2' layer of the ResNet18 model. In this way, the features extracted by the two models have the same spatial resolutions.
The experimental results are displayed in TABLE~\ref{quantitative:multispsr}. The SR models trained with the VGG19 model and the ResNet18 model are denoted as SPSR-VGG and SPSR-ResNet, respectively.  We can see that SPSR-VGG and SPSR-ResNet have similar  SR capacity, and both fail to perform as well as SPSR-J and SPSR-P. The reason is that both the VGG19 and the ResNet18 models are trained for large-scale image classification, which makes the models pay more attention on high-level semantic information rather than low-level texture information. Therefore, the pretrained VGG19 and ResNet18 models may filter out detailed geometric structures and mainly preserve global semantics. The features extracted by these pretrained models are less effective to provide guidance for the generator to produce perceptual-pleasant SR images with fine local structures. On the contrary, our proposed neural structure extractor is trained by the strategy of self-supervised structure learning, whose goal is to exploit local structural relationships of images. By imposing the proposed structure loss on the SR images, the generator focuses more on the recovery of details and becomes more powerful in preserving structures.

\begin{table*}
\small
\centering
\caption{Comparison of perceptual-driven SR methods on other network architectures. The best performance is \textbf{highlighted} in \textcolor{red}{\textbf{red}} (1st best) and \textcolor{blue}{\textbf{blue}} (2nd best). } \label{quantitative:other}
  \begin{tabular}{C{1.4cm}L{0.9cm}|C{1.4cm}C{1.4cm}C{1.4cm}C{1.4cm}|C{1.4cm}C{1.4cm}C{1.4cm}C{1.4cm}}
      \Xhline{1.0pt}
      \textbf{Dataset} & \textbf{Metric} & \textbf{EDSR} & \textbf{EDSR-GAN} & \textbf{EDSR-G} & \textbf{EDSR-P} & \textbf{SRResNet} & \textbf{SRResNet-GAN} & \textbf{SRResNet-G} & \textbf{SRResNet-P} \\
      \Xhline{1.0pt}
      \multirow{3}*{\textbf{Set5}}
      & LPIPS & 0.1726 & 0.0808 & \textcolor{blue}{\textbf{0.0794}} & \textcolor{red}{\textbf{0.0787}} & 0.1724 & 0.0882 & \textcolor{blue}{\textbf{0.0783}} & \textcolor{red}{\textbf{0.0731}}  \\
      & PSNR & \textcolor{red}{\textbf{32.074}} & 29.614 & 29.536 & \textcolor{blue}{\textbf{30.32}} & \textcolor{red}{\textbf{32.161}} & 29.168 & 29.632 & \textcolor{blue}{\textbf{30.410}}  \\ 
      & SSIM & \textcolor{red}{\textbf{0.9054}} & 0.8602 & 0.8556 & \textcolor{blue}{\textbf{0.8693}} & \textcolor{red}{\textbf{0.9065}} & 0.8613 & 0.8693 & \textcolor{blue}{\textbf{0.8738}}   \\  \hline
      \multirow{3}*{\textbf{Set14}} 
      & LPIPS & 0.2832 & \textcolor{blue}{\textbf{0.1473}} & \textcolor{red}{\textbf{0.1425}} & 0.1539 & 0.2837 & 0.1663 & \textcolor{blue}{\textbf{0.1418}} & \textcolor{red}{\textbf{0.1377}}   \\
      & PSNR & \textcolor{red}{\textbf{28.565}} & 26.321 & 26.337 & \textcolor{blue}{\textbf{26.881}} & \textcolor{red}{\textbf{28.589}} & 26.171 & 26.243 & \textcolor{red}{\textbf{26.781}}  \\ 
      & SSIM  & \textcolor{red}{\textbf{0.8510}} & 0.7898 & 0.7894 & \textcolor{blue}{\textbf{0.8054}} & \textcolor{red}{\textbf{0.8516}} & 0.7841 & 0.7818 & \textcolor{blue}{\textbf{0.8004}}   \\ \hline
      \multirow{3}*{\textbf{BSD100}}
      & LPIPS  & 0.3712 & \textcolor{blue}{\textbf{0.1841}} & \textcolor{red}{\textbf{0.1823}} & 0.1850 & 0.3705 & 0.1980 & \textcolor{blue}{\textbf{0.1816}} & \textcolor{red}{\textbf{0.1749}}  \\
      & PSNR & \textcolor{red}{\textbf{27.561}} & 25.182 & 25.218 & \textcolor{blue}{\textbf{26.062}} & \textcolor{red}{\textbf{27.579}} & 25.459 & 25.475 & \textcolor{blue}{\textbf{25.862}}   \\ 
      & SSIM  & \textcolor{red}{\textbf{0.7349}} & 0.6497 & 0.6531 & \textcolor{blue}{\textbf{0.6768}} & \textcolor{red}{\textbf{0.7358}} & 0.6485 & 0.6502 & \textcolor{blue}{\textbf{0.678}}  \\ \hline
      \multirow{3}*{\textbf{General100}}
      & LPIPS & 0.1785 & 0.1011 & \textcolor{blue}{\textbf{0.1008}} & \textcolor{red}{\textbf{0.0982}} & 0.1788 & 0.1055 & \textcolor{blue}{\textbf{0.0993}} & \textcolor{red}{\textbf{0.0944}}   \\
      & PSNR  & \textcolor{red}{\textbf{31.384}} & 29.003 & 28.829 & \textcolor{blue}{\textbf{29.763}} & \textcolor{red}{\textbf{31.450}} & 28.575 & 29.027 & \textcolor{blue}{\textbf{29.716}}   \\ 
      & SSIM  & \textcolor{red}{\textbf{0.8948}} & 0.8486 & 0.8458 & \textcolor{blue}{\textbf{0.8642}} & \textcolor{red}{\textbf{0.8956}} & 0.8541 & 0.8467 & \textcolor{blue}{\textbf{0.8618}}   \\ \hline
      \multirow{3}*{\textbf{Urban100}}
      & LPIPS  & 0.2281 & 0.1492 & \textcolor{red}{\textbf{0.1482}} & \textcolor{blue}{\textbf{0.1489}} & 0.2273 & 0.1551 & \textcolor{blue}{\textbf{0.1456}} & \textcolor{red}{\textbf{0.1399}}   \\
      & PSNR   & \textcolor{red}{\textbf{26.032}} & 23.865 & 23.930 & \textcolor{blue}{\textbf{24.574}} & \textcolor{red}{\textbf{26.109}} & 24.397 & 24.009 & \textcolor{blue}{\textbf{24.606}}  \\ 
      & SSIM  & \textcolor{red}{\textbf{0.9603}} & 0.9359 & 0.9368 & \textcolor{blue}{\textbf{0.9468}} & \textcolor{red}{\textbf{0.9602}} & 0.9381 & 0.9374 &  \textcolor{blue}{\textbf{0.9442}}  \\ \hline
      \Xhline{1.0pt}
  \end{tabular}
\end{table*} 

We have also trained an SPSR-R model with a randomly initialized NSE and displayed the  experimental results of SPSR-R in TABLE~\ref{quantitative:multispsr}. From the comparison, we see that SPSR-R performs better than SPSR-VGG and SPSR-ResNet, but is still inferior to SPSR-J and SPSR-P. This result reflects that the randomly initialized network has less information loss since it is not explicitly supervised to capture semantic information and filter out low-level information. Hence the structure loss based on such NSE can also be treated as a regularization term, which works better than the pretrained  VGG19 and ResNet18 models in SPSR-VGG and SPSR-ResNet. However, the randomly initialized NSE lacks the ability to extract effective structure information, which makes the supervision provided by the structure loss less helpful to enhance the SR capacity of the generator. 

\section{Experiments on Other Network Architectures}

We have conducted experiments on another two popular PSNR-oriented models, EDSR~\cite{EDSR} and SRResNet~\cite{ledig2017photo}. For EDSR, we use a pretrained EDSR model as  initialization and train an EDSR-GAN model by adding the perceptual loss and adversarial loss during training. Besides, we train an EDSR-G model with the proposed gradient loss and an EDSR-P model with the proposed structure loss. The experimental settings for SRResNet are the same, and we obtain SRResNet-GAN, SRResNet-G and, SRResNet-P, accordingly. The SR performance of these models on  five benchmark datasets is shown in TABLE~\ref{quantitative:other}. We see that EDSR and SRResNet obviously outperform the other models on PSNR and SSIM, but present much worse performance on LPIPS. With the assistant of the proposed gradient loss, EDSR-G and SRResNet-G successfully obtain better quantitative values than EDSR-GAN and SRResNet-GAN on  the  three metrics. Besides, EDSR-P and SRResNet-P both have the best SR ability among the compared perceptual-driven models. The experimental results demonstrate the effectiveness of the proposed gradient loss and structure loss.

\section{More Qualitative Results}

\subsection{SR Comparison}

\begin{figure*}[htbp]
\centering

\begin{minipage}[b]{\picwidth\linewidth}
\includegraphics[width=1 \linewidth]{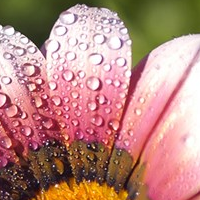}\vspace{\picvspace}
\centering{\scriptsize{HR ('im\_078' from General100)}}
\end{minipage}\hspace{\pagehspace}
\begin{minipage}[b]{\picwidth\linewidth}
\includegraphics[width=1 \linewidth]{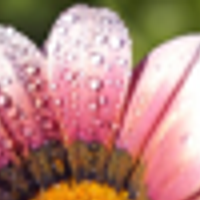}\vspace{\picvspace}
\centering{\scriptsize{LR}}
\end{minipage}\hspace{\pagehspace}
\begin{minipage}[b]{\picwidth\linewidth}
\includegraphics[width=1 \linewidth]{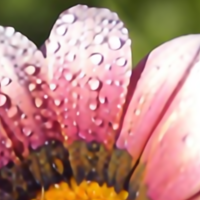}\vspace{\picvspace}
\centering{\scriptsize{RRDB}}
\end{minipage}\hspace{\pagehspace}
\begin{minipage}[b]{\picwidth\linewidth}
\includegraphics[width=1 \linewidth]{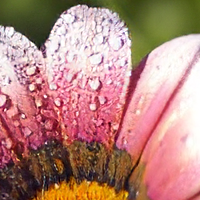}\vspace{\picvspace}
\centering{\scriptsize{SRGAN}}
\end{minipage}\vspace{\pagevspace}

\begin{minipage}[b]{\picwidth\linewidth}
\includegraphics[width=1 \linewidth]{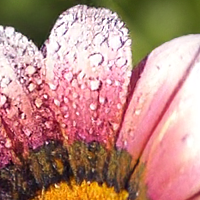}\vspace{\picvspace}
\centering{\scriptsize{ESRGAN}}
\end{minipage}\hspace{\pagehspace}
\begin{minipage}[b]{\picwidth\linewidth}
\includegraphics[width=1 \linewidth]{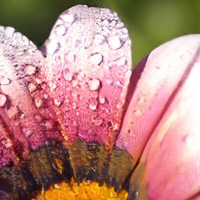}\vspace{\picvspace}
\centering{\scriptsize{NatSR}}
\end{minipage}\hspace{\pagehspace}
\begin{minipage}[b]{\picwidth\linewidth}
\includegraphics[width=1 \linewidth]{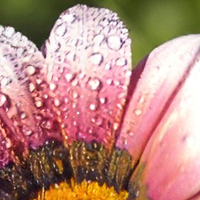}\vspace{\picvspace}
\centering{\scriptsize{SPSR-G}}
\end{minipage}\hspace{\pagehspace}
\begin{minipage}[b]{\picwidth\linewidth}
\includegraphics[width=1 \linewidth]{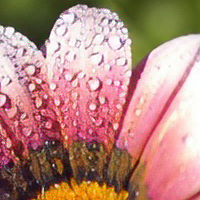}\vspace{\picvspace}
\centering{\scriptsize{SPSR-P}}
\end{minipage}\vspace{\packvspace}

\begin{minipage}[b]{\picwidth\linewidth}
\includegraphics[width=1 \linewidth]{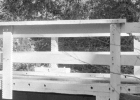}\vspace{\picvspace}
\centering{\scriptsize{HR ('bridge' from Set14)}}
\end{minipage}\hspace{\pagehspace}
\begin{minipage}[b]{\picwidth\linewidth}
\includegraphics[width=1 \linewidth]{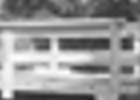}\vspace{\picvspace}
\centering{\scriptsize{LR}}
\end{minipage}\hspace{\pagehspace}
\begin{minipage}[b]{\picwidth\linewidth}
\includegraphics[width=1 \linewidth]{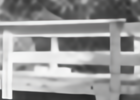}\vspace{\picvspace}
\centering{\scriptsize{RRDB}}
\end{minipage}\hspace{\pagehspace}
\begin{minipage}[b]{\picwidth\linewidth}
\includegraphics[width=1 \linewidth]{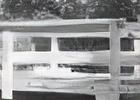}\vspace{\picvspace}
\centering{\scriptsize{SRGAN}}
\end{minipage}\vspace{\pagevspace}

\begin{minipage}[b]{\picwidth\linewidth}
\includegraphics[width=1 \linewidth]{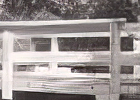}\vspace{\picvspace}
\centering{\scriptsize{ESRGAN}}
\end{minipage}\hspace{\pagehspace}
\begin{minipage}[b]{\picwidth\linewidth}
\includegraphics[width=1 \linewidth]{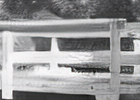}\vspace{\picvspace}
\centering{\scriptsize{NatSR}}
\end{minipage}\hspace{\pagehspace}
\begin{minipage}[b]{\picwidth\linewidth}
\includegraphics[width=1 \linewidth]{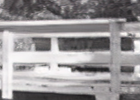}\vspace{\picvspace}
\centering{\scriptsize{SPSR-G}}
\end{minipage}\hspace{\pagehspace}
\begin{minipage}[b]{\picwidth\linewidth}
\includegraphics[width=1 \linewidth]{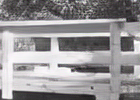}\vspace{\picvspace}
\centering{\scriptsize{SPSR-P}}
\end{minipage}\vspace{\packvspace}

\begin{minipage}[b]{\picwidth\linewidth}
\includegraphics[width=1 \linewidth]{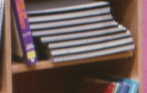}\vspace{\picvspace}
\centering{\scriptsize{HR ('barbara' from Set14)}}
\end{minipage}\hspace{\pagehspace}
\begin{minipage}[b]{\picwidth\linewidth}
\includegraphics[width=1 \linewidth]{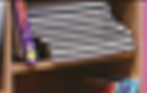}\vspace{\picvspace}
\centering{\scriptsize{LR}}
\end{minipage}\hspace{\pagehspace}
\begin{minipage}[b]{\picwidth\linewidth}
\includegraphics[width=1 \linewidth]{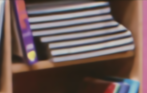}\vspace{\picvspace}
\centering{\scriptsize{RRDB}}
\end{minipage}\hspace{\pagehspace}
\begin{minipage}[b]{\picwidth\linewidth}
\includegraphics[width=1 \linewidth]{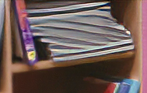}\vspace{\picvspace}
\centering{\scriptsize{SRGAN}}
\end{minipage}\vspace{\pagevspace}

\begin{minipage}[b]{\picwidth\linewidth}
\includegraphics[width=1 \linewidth]{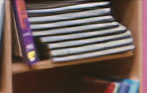}\vspace{\picvspace}
\centering{\scriptsize{ESRGAN}}
\end{minipage}\hspace{\pagehspace}
\begin{minipage}[b]{\picwidth\linewidth}
\includegraphics[width=1 \linewidth]{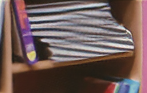}\vspace{\picvspace}
\centering{\scriptsize{NatSR}}
\end{minipage}\hspace{\pagehspace}
\begin{minipage}[b]{\picwidth\linewidth}
\includegraphics[width=1 \linewidth]{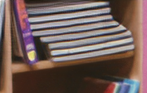}\vspace{\picvspace}
\centering{\scriptsize{SPSR-G}}
\end{minipage}\hspace{\pagehspace}
\begin{minipage}[b]{\picwidth\linewidth}
\includegraphics[width=1 \linewidth]{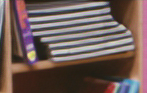}\vspace{\picvspace}
\centering{\scriptsize{SPSR-P}}
\end{minipage}\vspace{\packvspace}

\centering
\caption{Visual comparison of SR performance with state-of-the-art SR methods.}
\label{fig:supp_vis1}
\end{figure*}

\begin{figure*}[htbp]
\centering

\begin{minipage}[b]{\picwidth\linewidth}
\includegraphics[width=1 \linewidth]{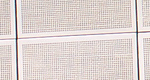}\vspace{\picvspace}
\centering{\scriptsize{HR ('img\_026' from Urban100)}}
\end{minipage}\hspace{\pagehspace}
\begin{minipage}[b]{\picwidth\linewidth}
\includegraphics[width=1 \linewidth]{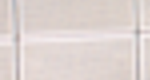}\vspace{\picvspace}
\centering{\scriptsize{LR}}
\end{minipage}\hspace{\pagehspace}
\begin{minipage}[b]{\picwidth\linewidth}
\includegraphics[width=1 \linewidth]{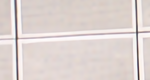}\vspace{\picvspace}
\centering{\scriptsize{RRDB}}
\end{minipage}\hspace{\pagehspace}
\begin{minipage}[b]{\picwidth\linewidth}
\includegraphics[width=1 \linewidth]{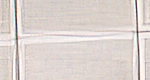}\vspace{\picvspace}
\centering{\scriptsize{SRGAN}}
\end{minipage}\vspace{\pagevspace}

\begin{minipage}[b]{\picwidth\linewidth}
\includegraphics[width=1 \linewidth]{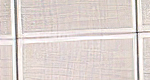}\vspace{\picvspace}
\centering{\scriptsize{ESRGAN}}
\end{minipage}\hspace{\pagehspace}
\begin{minipage}[b]{\picwidth\linewidth}
\includegraphics[width=1 \linewidth]{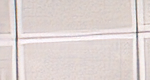}\vspace{\picvspace}
\centering{\scriptsize{NatSR}}
\end{minipage}\hspace{\pagehspace}
\begin{minipage}[b]{\picwidth\linewidth}
\includegraphics[width=1 \linewidth]{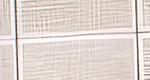}\vspace{\picvspace}
\centering{\scriptsize{SPSR-G}}
\end{minipage}\hspace{\pagehspace}
\begin{minipage}[b]{\picwidth\linewidth}
\includegraphics[width=1 \linewidth]{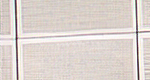}\vspace{\picvspace}
\centering{\scriptsize{SPSR-P}}
\end{minipage}\vspace{\packvspace}

\begin{minipage}[b]{\picwidth\linewidth}
\includegraphics[width=1 \linewidth]{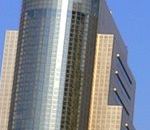}\vspace{\picvspace}
\centering{\scriptsize{HR ('img\_086' from Urban100)}}
\end{minipage}\hspace{\pagehspace}
\begin{minipage}[b]{\picwidth\linewidth}
\includegraphics[width=1 \linewidth]{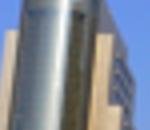}\vspace{\picvspace}
\centering{\scriptsize{LR}}
\end{minipage}\hspace{\pagehspace}
\begin{minipage}[b]{\picwidth\linewidth}
\includegraphics[width=1 \linewidth]{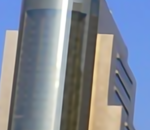}\vspace{\picvspace}
\centering{\scriptsize{RRDB}}
\end{minipage}\hspace{\pagehspace}
\begin{minipage}[b]{\picwidth\linewidth}
\includegraphics[width=1 \linewidth]{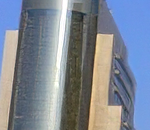}\vspace{\picvspace}
\centering{\scriptsize{SRGAN}}
\end{minipage}\vspace{\pagevspace}

\begin{minipage}[b]{\picwidth\linewidth}
\includegraphics[width=1 \linewidth]{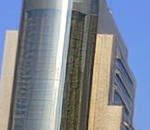}\vspace{\picvspace}
\centering{\scriptsize{ESRGAN}}
\end{minipage}\hspace{\pagehspace}
\begin{minipage}[b]{\picwidth\linewidth}
\includegraphics[width=1 \linewidth]{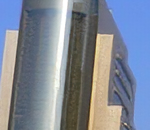}\vspace{\picvspace}
\centering{\scriptsize{NatSR}}
\end{minipage}\hspace{\pagehspace}
\begin{minipage}[b]{\picwidth\linewidth}
\includegraphics[width=1 \linewidth]{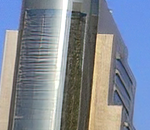}\vspace{\picvspace}
\centering{\scriptsize{SPSR-G}}
\end{minipage}\hspace{\pagehspace}
\begin{minipage}[b]{\picwidth\linewidth}
\includegraphics[width=1 \linewidth]{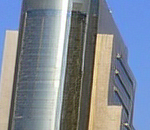}\vspace{\picvspace}
\centering{\scriptsize{SPSR-P}}
\end{minipage}\vspace{\packvspace}

\begin{minipage}[b]{\picwidth\linewidth}
\includegraphics[width=1 \linewidth]{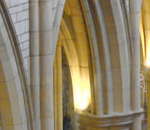}\vspace{\picvspace}
\centering{\scriptsize{HR ('img\_065' from Urban100)}}
\end{minipage}\hspace{\pagehspace}
\begin{minipage}[b]{\picwidth\linewidth}
\includegraphics[width=1 \linewidth]{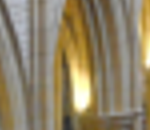}\vspace{\picvspace}
\centering{\scriptsize{LR}}
\end{minipage}\hspace{\pagehspace}
\begin{minipage}[b]{\picwidth\linewidth}
\includegraphics[width=1 \linewidth]{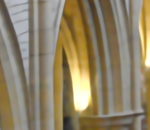}\vspace{\picvspace}
\centering{\scriptsize{RRDB}}
\end{minipage}\hspace{\pagehspace}
\begin{minipage}[b]{\picwidth\linewidth}
\includegraphics[width=1 \linewidth]{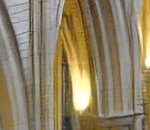}\vspace{\picvspace}
\centering{\scriptsize{SRGAN}}
\end{minipage}\vspace{\pagevspace}

\begin{minipage}[b]{\picwidth\linewidth}
\includegraphics[width=1 \linewidth]{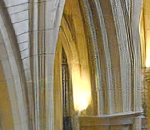}\vspace{\picvspace}
\centering{\scriptsize{ESRGAN}}
\end{minipage}\hspace{\pagehspace}
\begin{minipage}[b]{\picwidth\linewidth}
\includegraphics[width=1 \linewidth]{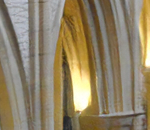}\vspace{\picvspace}
\centering{\scriptsize{NatSR}}
\end{minipage}\hspace{\pagehspace}
\begin{minipage}[b]{\picwidth\linewidth}
\includegraphics[width=1 \linewidth]{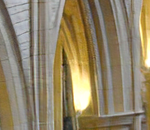}\vspace{\picvspace}
\centering{\scriptsize{SPSR-G}}
\end{minipage}\hspace{\pagehspace}
\begin{minipage}[b]{\picwidth\linewidth}
\includegraphics[width=1 \linewidth]{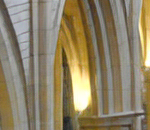}\vspace{\picvspace}
\centering{\scriptsize{SPSR-P}}
\end{minipage}\vspace{\packvspace}

\centering
\caption{Visual comparison of SR performance with state-of-the-art SR methods.}
\label{fig:supp_vis2}
\end{figure*}

\begin{figure*}[htbp]
\centering

\begin{minipage}[b]{\picwidth\linewidth}
\includegraphics[width=1 \linewidth]{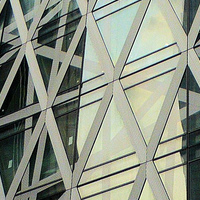}\vspace{\picvspace}
\centering{\scriptsize{HR ('img\_039' from Urban100)}}
\end{minipage}\hspace{\pagehspace}
\begin{minipage}[b]{\picwidth\linewidth}
\includegraphics[width=1 \linewidth]{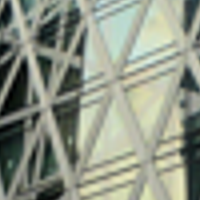}\vspace{\picvspace}
\centering{\scriptsize{LR}}
\end{minipage}\hspace{\pagehspace}
\begin{minipage}[b]{\picwidth\linewidth}
\includegraphics[width=1 \linewidth]{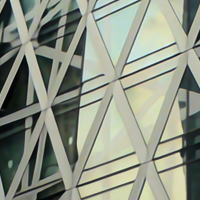}\vspace{\picvspace}
\centering{\scriptsize{RRDB}}
\end{minipage}\hspace{\pagehspace}
\begin{minipage}[b]{\picwidth\linewidth}
\includegraphics[width=1 \linewidth]{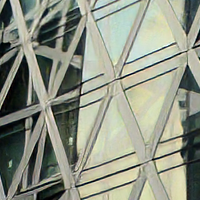}\vspace{\picvspace}
\centering{\scriptsize{SRGAN}}
\end{minipage}\vspace{\pagevspace}

\begin{minipage}[b]{\picwidth\linewidth}
\includegraphics[width=1 \linewidth]{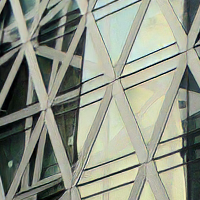}\vspace{\picvspace}
\centering{\scriptsize{ESRGAN}}
\end{minipage}\hspace{\pagehspace}
\begin{minipage}[b]{\picwidth\linewidth}
\includegraphics[width=1 \linewidth]{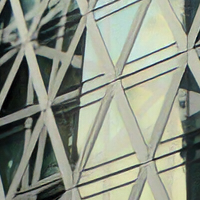}\vspace{\picvspace}
\centering{\scriptsize{NatSR}}
\end{minipage}\hspace{\pagehspace}
\begin{minipage}[b]{\picwidth\linewidth}
\includegraphics[width=1 \linewidth]{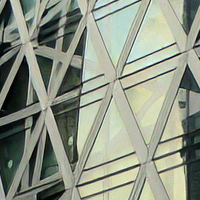}\vspace{\picvspace}
\centering{\scriptsize{SPSR-G}}
\end{minipage}\hspace{\pagehspace}
\begin{minipage}[b]{\picwidth\linewidth}
\includegraphics[width=1 \linewidth]{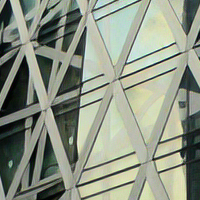}\vspace{\picvspace}
\centering{\scriptsize{SPSR-P}}
\end{minipage}\vspace{\packvspace}

\begin{minipage}[b]{\picwidth\linewidth}
\includegraphics[width=1 \linewidth]{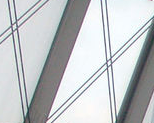}\vspace{\picvspace}
\centering{\scriptsize{HR ('img\_002' from Urban100)}}
\end{minipage}\hspace{\pagehspace}
\begin{minipage}[b]{\picwidth\linewidth}
\includegraphics[width=1 \linewidth]{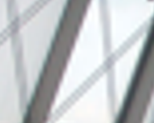}\vspace{\picvspace}
\centering{\scriptsize{LR}}
\end{minipage}\hspace{\pagehspace}
\begin{minipage}[b]{\picwidth\linewidth}
\includegraphics[width=1 \linewidth]{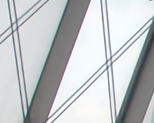}\vspace{\picvspace}
\centering{\scriptsize{RRDB}}
\end{minipage}\hspace{\pagehspace}
\begin{minipage}[b]{\picwidth\linewidth}
\includegraphics[width=1 \linewidth]{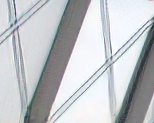}\vspace{\picvspace}
\centering{\scriptsize{SRGAN}}
\end{minipage}\vspace{\pagevspace}

\begin{minipage}[b]{\picwidth\linewidth}
\includegraphics[width=1 \linewidth]{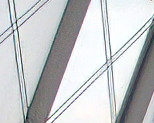}\vspace{\picvspace}
\centering{\scriptsize{ESRGAN}}
\end{minipage}\hspace{\pagehspace}
\begin{minipage}[b]{\picwidth\linewidth}
\includegraphics[width=1 \linewidth]{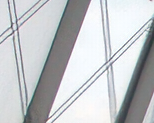}\vspace{\picvspace}
\centering{\scriptsize{NatSR}}
\end{minipage}\hspace{\pagehspace}
\begin{minipage}[b]{\picwidth\linewidth}
\includegraphics[width=1 \linewidth]{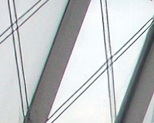}\vspace{\picvspace}
\centering{\scriptsize{SPSR-G}}
\end{minipage}\hspace{\pagehspace}
\begin{minipage}[b]{\picwidth\linewidth}
\includegraphics[width=1 \linewidth]{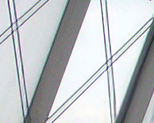}\vspace{\picvspace}
\centering{\scriptsize{SPSR-P}}
\end{minipage}\vspace{\packvspace}

\begin{minipage}[b]{\picwidth\linewidth}
\includegraphics[width=1 \linewidth]{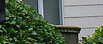}\vspace{\picvspace}
\centering{\scriptsize{HR ('img\_003' from Urban100)}}
\end{minipage}\hspace{\pagehspace}
\begin{minipage}[b]{\picwidth\linewidth}
\includegraphics[width=1 \linewidth]{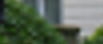}\vspace{\picvspace}
\centering{\scriptsize{LR}}
\end{minipage}\hspace{\pagehspace}
\begin{minipage}[b]{\picwidth\linewidth}
\includegraphics[width=1 \linewidth]{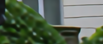}\vspace{\picvspace}
\centering{\scriptsize{RRDB}}
\end{minipage}\hspace{\pagehspace}
\begin{minipage}[b]{\picwidth\linewidth}
\includegraphics[width=1 \linewidth]{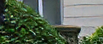}\vspace{\picvspace}
\centering{\scriptsize{SRGAN}}
\end{minipage}\vspace{\pagevspace}

\begin{minipage}[b]{\picwidth\linewidth}
\includegraphics[width=1 \linewidth]{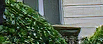}\vspace{\picvspace}
\centering{\scriptsize{ESRGAN}}
\end{minipage}\hspace{\pagehspace}
\begin{minipage}[b]{\picwidth\linewidth}
\includegraphics[width=1 \linewidth]{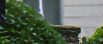}\vspace{\picvspace}
\centering{\scriptsize{NatSR}}
\end{minipage}\hspace{\pagehspace}
\begin{minipage}[b]{\picwidth\linewidth}
\includegraphics[width=1 \linewidth]{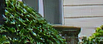}\vspace{\picvspace}
\centering{\scriptsize{SPSR-G}}
\end{minipage}\hspace{\pagehspace}
\begin{minipage}[b]{\picwidth\linewidth}
\includegraphics[width=1 \linewidth]{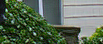}\vspace{\picvspace}
\centering{\scriptsize{SPSR-P}}
\end{minipage}\vspace{\captionvspace}

\centering
\caption{Visual comparison of SR performance with state-of-the-art SR methods.}
\label{fig:supp_vis3}
\end{figure*}

\begin{figure*}[htbp]
\centering

\begin{minipage}[b]{\picwidth\linewidth}
\includegraphics[width=1 \linewidth]{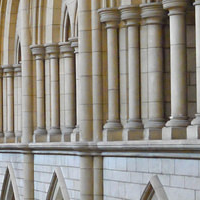}\vspace{\picvspace}
\centering{\scriptsize{HR ('img\_065' from Urban100)}}
\end{minipage}\hspace{\pagehspace}
\begin{minipage}[b]{\picwidth\linewidth}
\includegraphics[width=1 \linewidth]{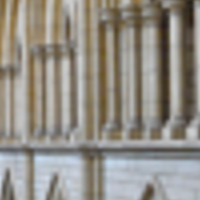}\vspace{\picvspace}
\centering{\scriptsize{LR}}
\end{minipage}\hspace{\pagehspace}
\begin{minipage}[b]{\picwidth\linewidth}
\includegraphics[width=1 \linewidth]{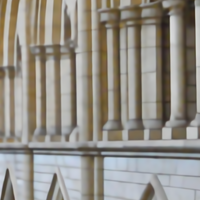}\vspace{\picvspace}
\centering{\scriptsize{RRDB}}
\end{minipage}\hspace{\pagehspace}
\begin{minipage}[b]{\picwidth\linewidth}
\includegraphics[width=1 \linewidth]{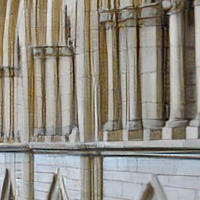}\vspace{\picvspace}
\centering{\scriptsize{SRGAN}}
\end{minipage}\vspace{\pagevspace}

\begin{minipage}[b]{\picwidth\linewidth}
\includegraphics[width=1 \linewidth]{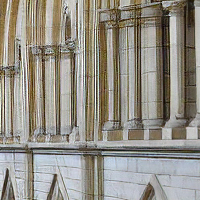}\vspace{\picvspace}
\centering{\scriptsize{ESRGAN}}
\end{minipage}\hspace{\pagehspace}
\begin{minipage}[b]{\picwidth\linewidth}
\includegraphics[width=1 \linewidth]{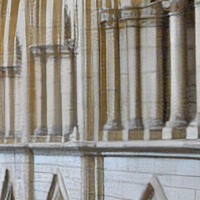}\vspace{\picvspace}
\centering{\scriptsize{NatSR}}
\end{minipage}\hspace{\pagehspace}
\begin{minipage}[b]{\picwidth\linewidth}
\includegraphics[width=1 \linewidth]{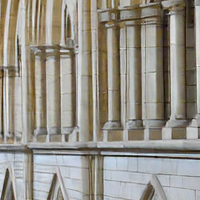}\vspace{\picvspace}
\centering{\scriptsize{SPSR-G}}
\end{minipage}\hspace{\pagehspace}
\begin{minipage}[b]{\picwidth\linewidth}
\includegraphics[width=1 \linewidth]{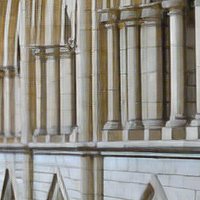}\vspace{\picvspace}
\centering{\scriptsize{SPSR-P}}
\end{minipage}\vspace{\packvspace}

\begin{minipage}[b]{\picwidth\linewidth}
\includegraphics[width=1 \linewidth]{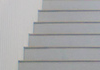}\vspace{\picvspace}
\centering{\scriptsize{HR ('img\_009' from Urban100)}}
\end{minipage}\hspace{\pagehspace}
\begin{minipage}[b]{\picwidth\linewidth}
\includegraphics[width=1 \linewidth]{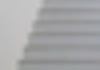}\vspace{\picvspace}
\centering{\scriptsize{LR}}
\end{minipage}\hspace{\pagehspace}
\begin{minipage}[b]{\picwidth\linewidth}
\includegraphics[width=1 \linewidth]{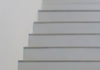}\vspace{\picvspace}
\centering{\scriptsize{RRDB}}
\end{minipage}\hspace{\pagehspace}
\begin{minipage}[b]{\picwidth\linewidth}
\includegraphics[width=1 \linewidth]{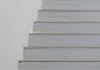}\vspace{\picvspace}
\centering{\scriptsize{SRGAN}}
\end{minipage}\vspace{\pagevspace}

\begin{minipage}[b]{\picwidth\linewidth}
\includegraphics[width=1 \linewidth]{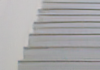}\vspace{\picvspace}
\centering{\scriptsize{ESRGAN}}
\end{minipage}\hspace{\pagehspace}
\begin{minipage}[b]{\picwidth\linewidth}
\includegraphics[width=1 \linewidth]{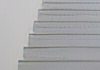}\vspace{\picvspace}
\centering{\scriptsize{NatSR}}
\end{minipage}\hspace{\pagehspace}
\begin{minipage}[b]{\picwidth\linewidth}
\includegraphics[width=1 \linewidth]{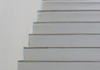}\vspace{\picvspace}
\centering{\scriptsize{SPSR-G}}
\end{minipage}\hspace{\pagehspace}
\begin{minipage}[b]{\picwidth\linewidth}
\includegraphics[width=1 \linewidth]{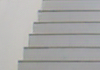}\vspace{\picvspace}
\centering{\scriptsize{SPSR-P}}
\end{minipage}\vspace{\packvspace}

\begin{minipage}[b]{\picwidth\linewidth}
\includegraphics[width=1 \linewidth]{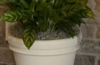}\vspace{\picvspace}
\centering{\scriptsize{HR ('im\_005' from General100)}}
\end{minipage}\hspace{\pagehspace}
\begin{minipage}[b]{\picwidth\linewidth}
\includegraphics[width=1 \linewidth]{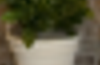}\vspace{\picvspace}
\centering{\scriptsize{LR}}
\end{minipage}\hspace{\pagehspace}
\begin{minipage}[b]{\picwidth\linewidth}
\includegraphics[width=1 \linewidth]{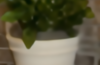}\vspace{\picvspace}
\centering{\scriptsize{RRDB}}
\end{minipage}\hspace{\pagehspace}
\begin{minipage}[b]{\picwidth\linewidth}
\includegraphics[width=1 \linewidth]{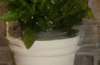}\vspace{\picvspace}
\centering{\scriptsize{SRGAN}}
\end{minipage}\vspace{\pagevspace}

\begin{minipage}[b]{\picwidth\linewidth}
\includegraphics[width=1 \linewidth]{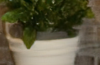}\vspace{\picvspace}
\centering{\scriptsize{ESRGAN}}
\end{minipage}\hspace{\pagehspace}
\begin{minipage}[b]{\picwidth\linewidth}
\includegraphics[width=1 \linewidth]{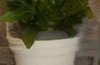}\vspace{\picvspace}
\centering{\scriptsize{NatSR}}
\end{minipage}\hspace{\pagehspace}
\begin{minipage}[b]{\picwidth\linewidth}
\includegraphics[width=1 \linewidth]{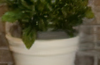}\vspace{\picvspace}
\centering{\scriptsize{SPSR-G}}
\end{minipage}\hspace{\pagehspace}
\begin{minipage}[b]{\picwidth\linewidth}
\includegraphics[width=1 \linewidth]{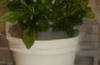}\vspace{\picvspace}
\centering{\scriptsize{SPSR-P}}
\end{minipage}\vspace{\packvspace}

\centering
\caption{Visual comparison of SR performance with state-of-the-art SR methods.}
\label{fig:supp_vis4}
\end{figure*}

We display more comparison of SR performance with state-of-the-art SR methods including SFTGAN~\cite{SFTGAN}, SRGAN~\cite{ledig2017photo}, ESRGAN~\cite{wang2018esrgan} and NatSR~\cite{soh2019natural}, as shown in Fig.~\ref{fig:supp_vis1}, Fig.~\ref{fig:supp_vis2}, Fig.~\ref{fig:supp_vis3} and Fig.~\ref{fig:supp_vis4}. Taking the first two rows in Fig.~\ref{fig:supp_vis1} as an example, our SPSR-G and SPSR-P methods present the most accurate and pleasant appearance for the drops on the flower. The SR results on other images also show the proposed methods perform better than other methods in preserving geometric structures and super-resolving photo-realistic images. By comparing SPSR-G and SPSR-P, we observe that SPSR-P recovers finer details than SPSR-G, which demonstrates the effectiveness of the proposed neural structure extractor and structure loss. 

\subsection{Visualization of Gradient Maps}

\begin{figure*}[htbp]
\centering

\begin{minipage}[b]{\picwidth\linewidth}
\includegraphics[width=1 \linewidth]{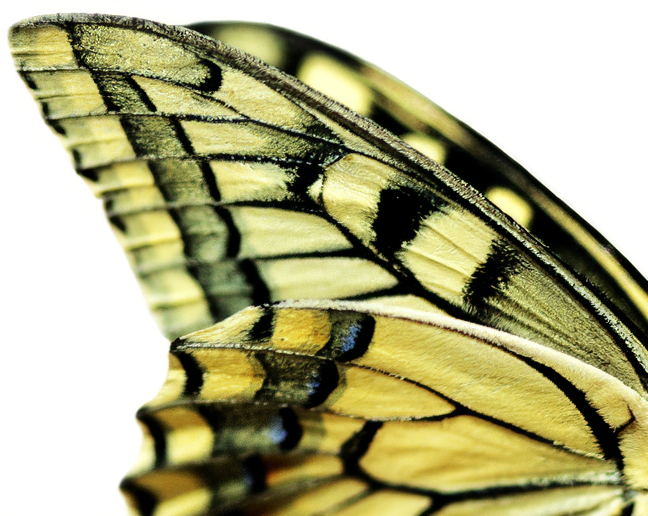}\vspace{\picvspace}
\centering{\scriptsize{HR ('im\_014' from General100)}}
\end{minipage}\hspace{\pagehspace}
\begin{minipage}[b]{\picwidth\linewidth}
\includegraphics[width=1 \linewidth]{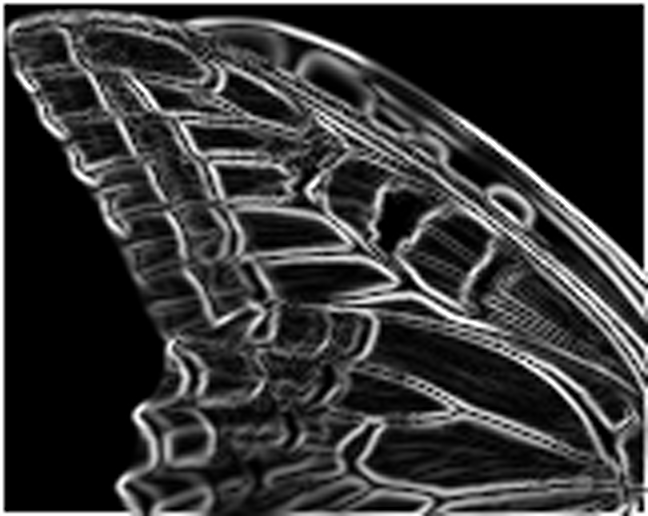}\vspace{\picvspace}
\centering{\scriptsize{LR gradient (Bicubic)}}
\end{minipage}\hspace{\pagehspace}
\begin{minipage}[b]{\picwidth\linewidth}
\includegraphics[width=1 \linewidth]{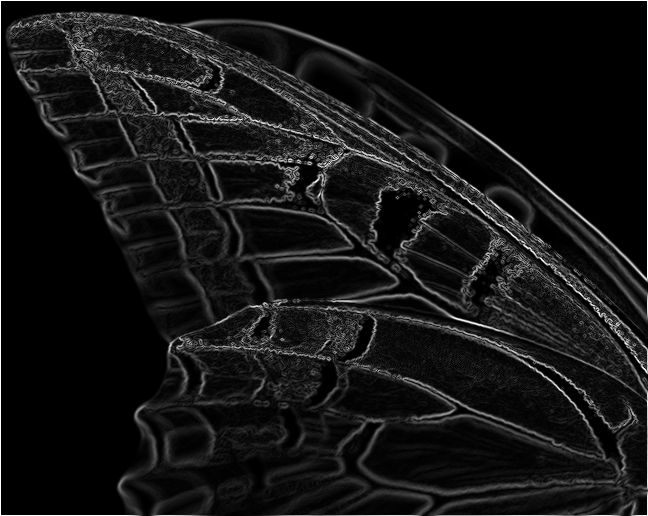}\vspace{\picvspace}
\centering{\scriptsize{HR gradient}}
\end{minipage}\hspace{\pagehspace}
\begin{minipage}[b]{\picwidth\linewidth}
\includegraphics[width=1 \linewidth]{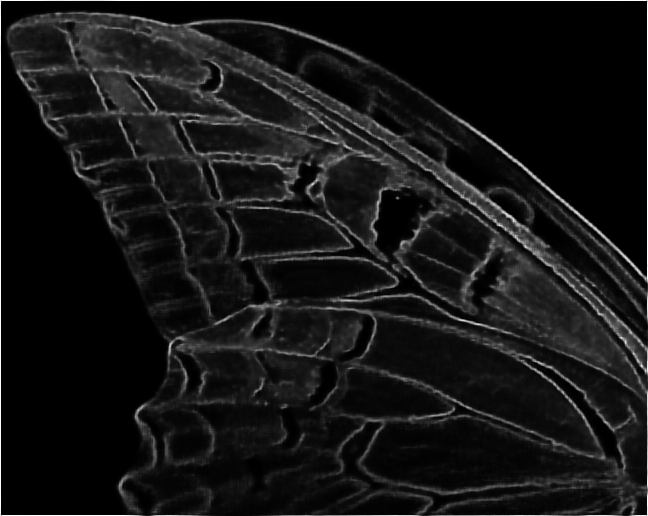}\vspace{\picvspace}
\centering{\scriptsize{Output of the gradient branch}}
\end{minipage}\vspace{\packvspace}

\begin{minipage}[b]{\picwidth\linewidth}
\includegraphics[width=1 \linewidth]{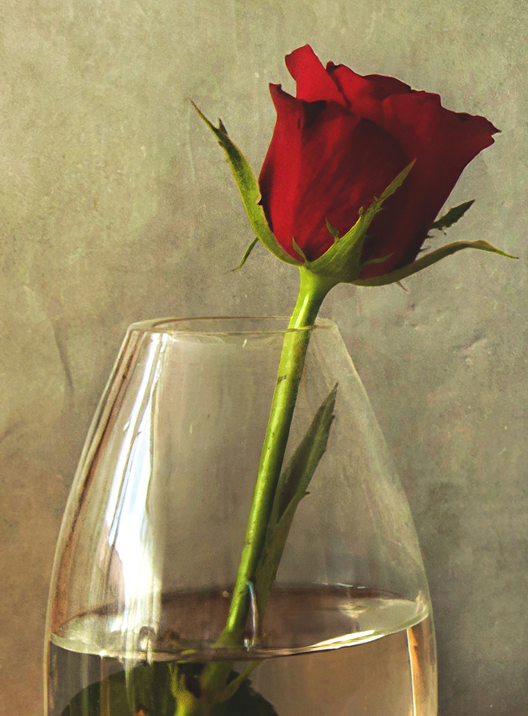}\vspace{\picvspace}
\centering{\scriptsize{HR ('im\_026' from General100)}}
\end{minipage}\hspace{\pagehspace}
\begin{minipage}[b]{\picwidth\linewidth}
\includegraphics[width=1 \linewidth]{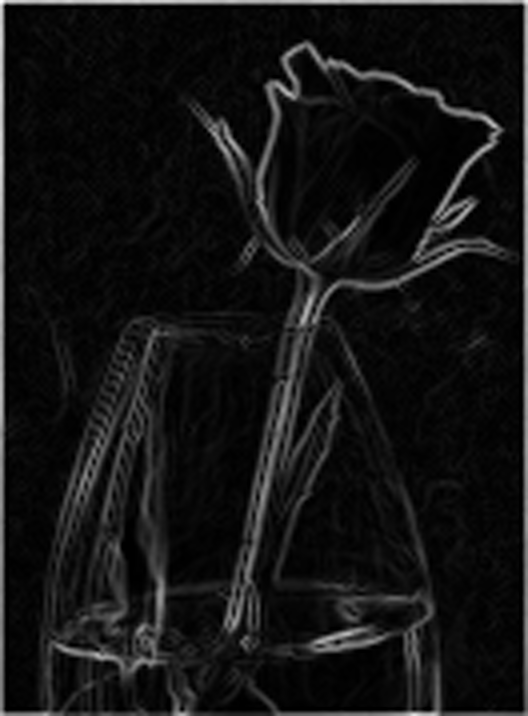}\vspace{\picvspace}
\centering{\scriptsize{LR gradient (Bicubic)}}
\end{minipage}\hspace{\pagehspace}
\begin{minipage}[b]{\picwidth\linewidth}
\includegraphics[width=1 \linewidth]{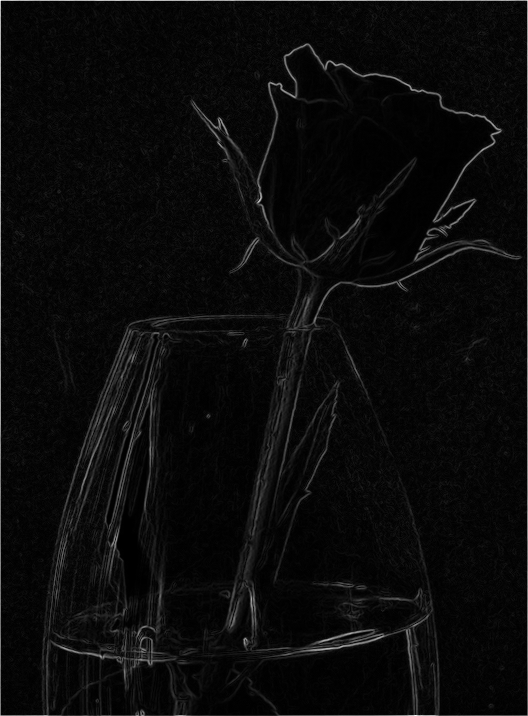}\vspace{\picvspace}
\centering{\scriptsize{HR gradient}}
\end{minipage}\hspace{\pagehspace}
\begin{minipage}[b]{\picwidth\linewidth}
\includegraphics[width=1 \linewidth]{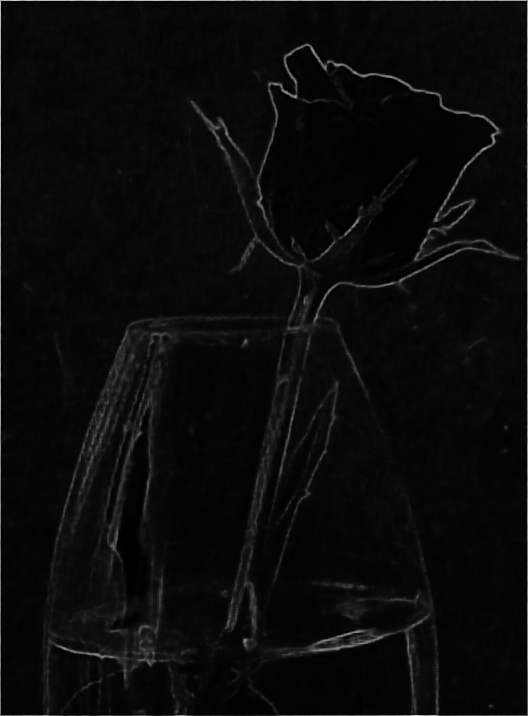}\vspace{\picvspace}
\centering{\scriptsize{Output of the gradient branch}}
\end{minipage}\vspace{\packvspace}

\begin{minipage}[b]{\picwidth\linewidth}
\includegraphics[width=1 \linewidth]{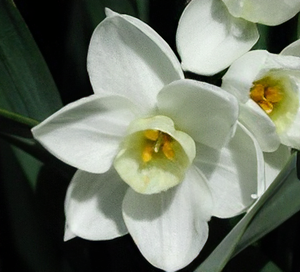}\vspace{\picvspace}
\centering{\scriptsize{HR ('im\_055' from General100)}}
\end{minipage}\hspace{\pagehspace}
\begin{minipage}[b]{\picwidth\linewidth}
\includegraphics[width=1 \linewidth]{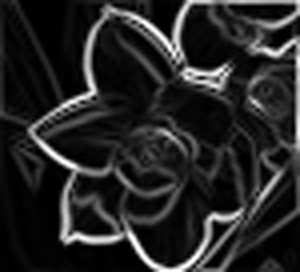}\vspace{\picvspace}
\centering{\scriptsize{LR gradient (Bicubic)}}
\end{minipage}\hspace{\pagehspace}
\begin{minipage}[b]{\picwidth\linewidth}
\includegraphics[width=1 \linewidth]{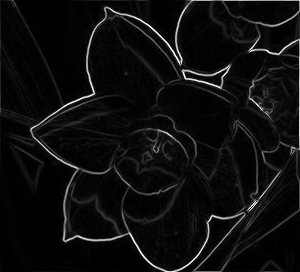}\vspace{\picvspace}
\centering{\scriptsize{HR gradient}}
\end{minipage}\hspace{\pagehspace}
\begin{minipage}[b]{\picwidth\linewidth}
\includegraphics[width=1 \linewidth]{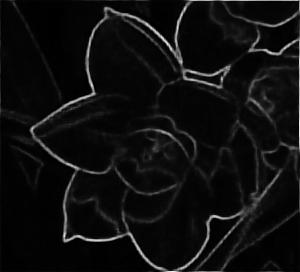}\vspace{\picvspace}
\centering{\scriptsize{Output of the gradient branch}}
\end{minipage}\vspace{\packvspace}

\begin{minipage}[b]{\picwidth\linewidth}
\includegraphics[width=1 \linewidth]{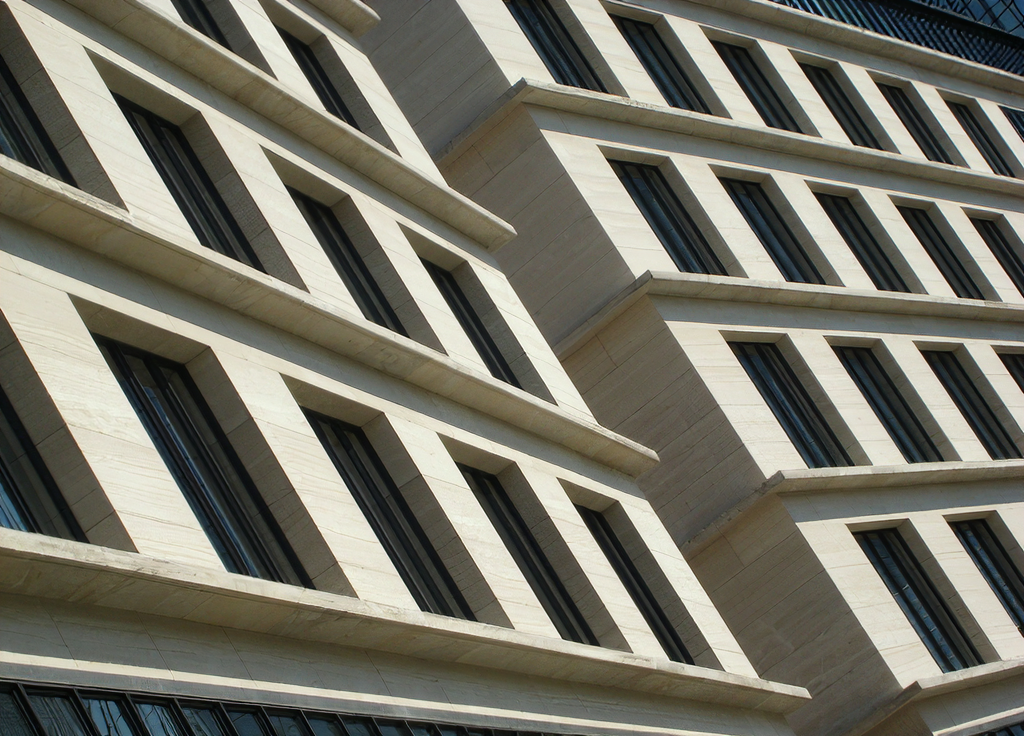}\vspace{\picvspace}
\centering{\scriptsize{HR ('img\_025' from Urban100)}}
\end{minipage}\hspace{\pagehspace}
\begin{minipage}[b]{\picwidth\linewidth}
\includegraphics[width=1 \linewidth]{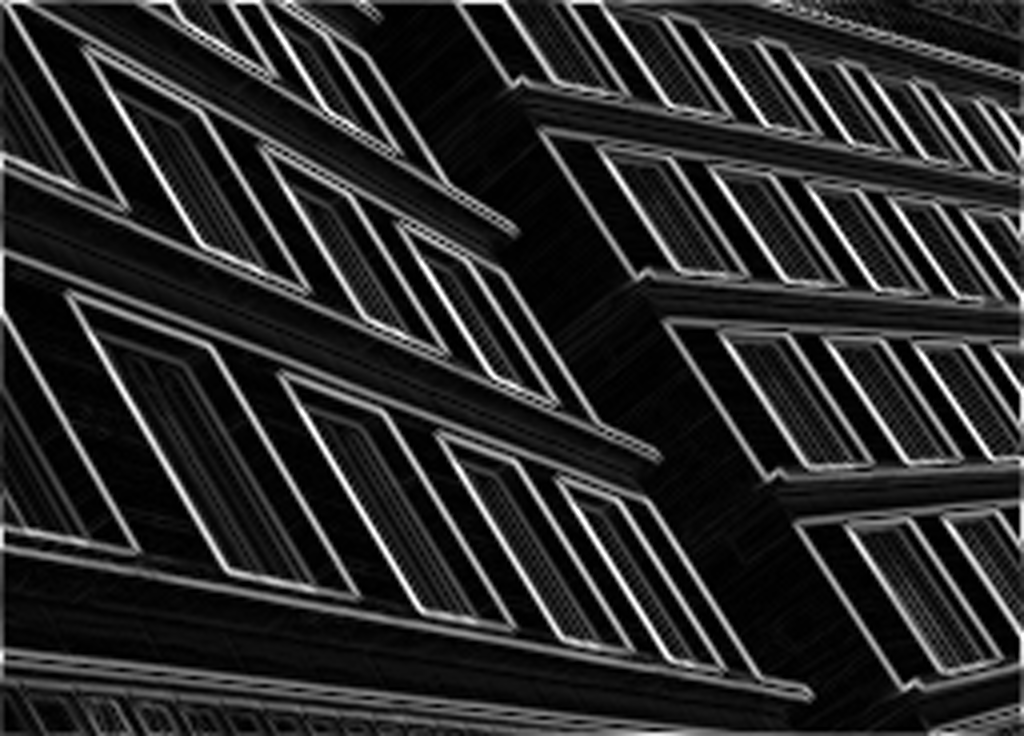}\vspace{\picvspace}
\centering{\scriptsize{LR gradient (Bicubic)}}
\end{minipage}\hspace{\pagehspace}
\begin{minipage}[b]{\picwidth\linewidth}
\includegraphics[width=1 \linewidth]{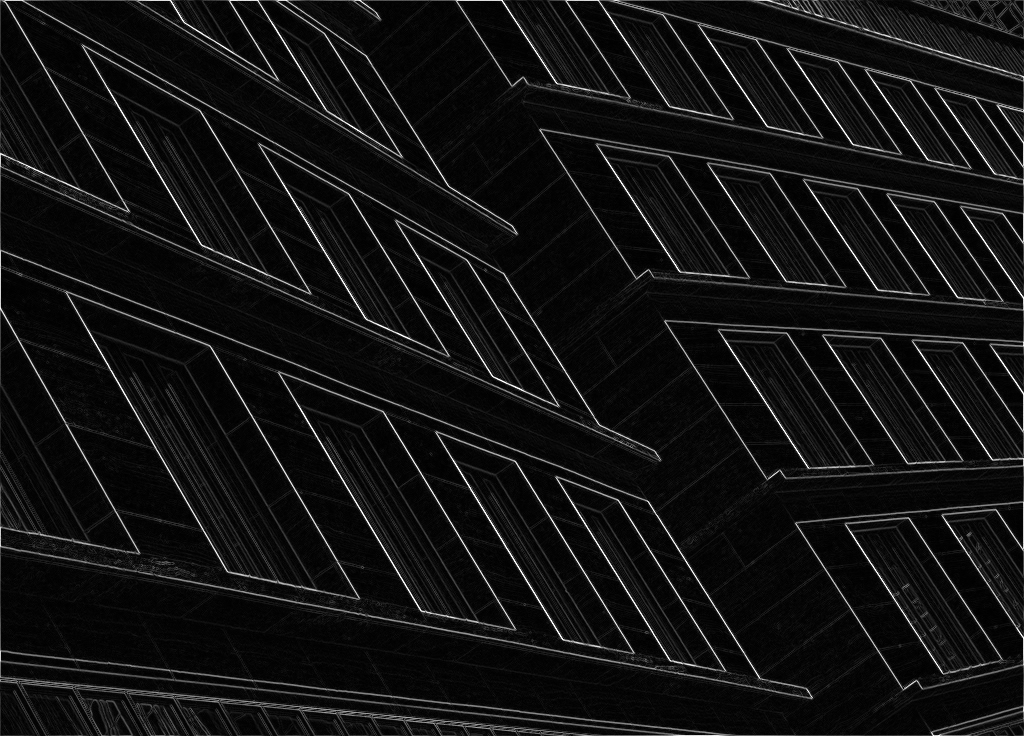}\vspace{\picvspace}
\centering{\scriptsize{HR gradient}}
\end{minipage}\hspace{\pagehspace}
\begin{minipage}[b]{\picwidth\linewidth}
\includegraphics[width=1 \linewidth]{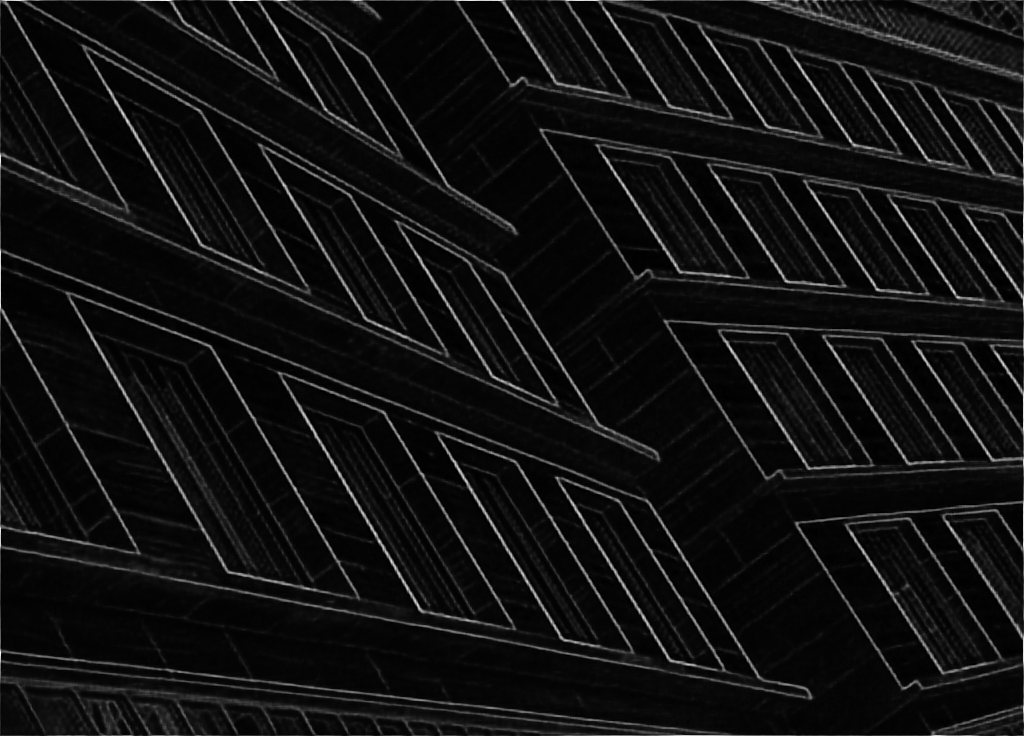}\vspace{\picvspace}
\centering{\scriptsize{Output of the gradient branch}}
\end{minipage}\vspace{\captionvspace}

\centering
\caption{Visualization of gradient maps for the SPSR-G model. }
\label{fig:supp_grad}
\end{figure*}

We visualize the outputs for the gradient branch of the SPSR-G model in Fig.~\ref{fig:supp_grad}. We see the gradient branch succeeds in converting LR gradient maps to the HR ones.

\ifCLASSOPTIONcaptionsoff
  \newpage
\fi


{\small
\bibliographystyle{IEEEtranS}
\bibliography{egbib}

\begin{thebibliography}{10}
\providecommand{\url}[1]{#1}
\csname url@samestyle\endcsname
\providecommand{\newblock}{\relax}
\providecommand{\bibinfo}[2]{#2}
\providecommand{\BIBentrySTDinterwordspacing}{\spaceskip=0pt\relax}
\providecommand{\BIBentryALTinterwordstretchfactor}{4}
\providecommand{\BIBentryALTinterwordspacing}{\spaceskip=\fontdimen2\font plus
\BIBentryALTinterwordstretchfactor\fontdimen3\font minus
  \fontdimen4\font\relax}
\providecommand{\BIBforeignlanguage}[2]{{%
\expandafter\ifx\csname l@#1\endcsname\relax
\typeout{** WARNING: IEEEtranS.bst: No hyphenation pattern has been}%
\typeout{** loaded for the language `#1'. Using the pattern for}%
\typeout{** the default language instead.}%
\else
\language=\csname l@#1\endcsname
\fi
#2}}
\providecommand{\BIBdecl}{\relax}
\BIBdecl

\bibitem{agustsson2017ntire}
E.~Agustsson and R.~Timofte, ``Ntire 2017 challenge on single image
  super-resolution: Dataset and study,'' in \emph{CVPR}, 2017, pp. 126--135.

\bibitem{anoosheh2019night}
A.~Anoosheh, T.~Sattler, R.~Timofte, M.~Pollefeys, and L.~Van~Gool,
  ``Night-to-day image translation for retrieval-based localization,'' in
  \emph{ICRA}.\hskip 1em plus 0.5em minus 0.4em\relax IEEE, 2019, pp.
  5958--5964.

\bibitem{arjovsky2017wasserstein}
M.~Arjovsky, S.~Chintala, and L.~Bottou, ``Wasserstein gan,'' \emph{arXiv
  preprint arXiv:1701.07875}, 2017.

\bibitem{bae2017beyond}
W.~Bae, J.~Yoo, and J.~Chul~Ye, ``Beyond deep residual learning for image
  restoration: Persistent homology-guided manifold simplification,'' in
  \emph{Proceedings of the IEEE conference on computer vision and pattern
  recognition workshops}, 2017, pp. 145--153.

\bibitem{bahat2020explorable}
Y.~Bahat and T.~Michaeli, ``Explorable super resolution,'' in \emph{CVPR},
  2020, pp. 2716--2725.

\bibitem{berthelot2017began}
D.~Berthelot, T.~Schumm, and L.~Metz, ``Began: Boundary equilibrium generative
  adversarial networks,'' \emph{arXiv preprint arXiv:1703.10717}, 2017.

\bibitem{bevilacqua2012low}
M.~Bevilacqua, A.~Roumy, C.~Guillemot, and M.-L. Alberi-Morel, ``Low-complexity
  single-image super-resolution based on nonnegative neighbor embedding,'' in
  \emph{BMVC}, 2012.

\bibitem{chen2020simple}
T.~Chen, S.~Kornblith, M.~Norouzi, and G.~Hinton, ``A simple framework for
  contrastive learning of visual representations,'' \emph{arXiv preprint
  arXiv:2002.05709}, 2020.

\bibitem{dong2014learning}
C.~Dong, C.~C. Loy, K.~He, and X.~Tang, ``Learning a deep convolutional network
  for image super-resolution,'' in \emph{ECCV}.\hskip 1em plus 0.5em minus
  0.4em\relax Springer, 2014, pp. 184--199.

\bibitem{dong2016accelerating}
C.~Dong, C.~C. Loy, and X.~Tang, ``Accelerating the super-resolution
  convolutional neural network,'' in \emph{ECCV}.\hskip 1em plus 0.5em minus
  0.4em\relax Springer, 2016, pp. 391--407.

\bibitem{fattal2007image}
R.~Fattal, ``Image upsampling via imposed edge statistics,'' \emph{TOG},
  vol.~26, no.~3, p.~95, 2007.

\bibitem{goodfellow2014generative}
I.~Goodfellow, J.~Pouget-Abadie, M.~Mirza, B.~Xu, D.~Warde-Farley, S.~Ozair,
  A.~Courville, and Y.~Bengio, ``Generative adversarial nets,'' in
  \emph{NeurIPS}, 2014, pp. 2672--2680.

\bibitem{gulrajani2017improved}
I.~Gulrajani, F.~Ahmed, M.~Arjovsky, V.~Dumoulin, and A.~C. Courville,
  ``Improved training of wasserstein gans,'' in \emph{NeurIPS}, 2017, pp.
  5767--5777.

\bibitem{guo2020closed}
Y.~Guo, J.~Chen, J.~Wang, Q.~Chen, J.~Cao, Z.~Deng, Y.~Xu, and M.~Tan,
  ``Closed-loop matters: Dual regression networks for single image
  super-resolution,'' in \emph{CVPR}, 2020, pp. 5407--5416.

\bibitem{he2016deep}
K.~He, X.~Zhang, S.~Ren, and J.~Sun, ``Deep residual learning for image
  recognition,'' in \emph{CVPR}, 2016, pp. 770--778.

\bibitem{hinton2006reducing}
G.~E. Hinton and R.~R. Salakhutdinov, ``Reducing the dimensionality of data
  with neural networks,'' \emph{science}, vol. 313, no. 5786, pp. 504--507,
  2006.

\bibitem{huang2017wavelet}
H.~Huang, R.~He, Z.~Sun, and T.~Tan, ``Wavelet-srnet: A wavelet-based cnn for
  multi-scale face super resolution,'' in \emph{Proceedings of the IEEE
  International Conference on Computer Vision}, 2017, pp. 1689--1697.

\bibitem{huang2015single}
J.-B. Huang, A.~Singh, and N.~Ahuja, ``Single image super-resolution from
  transformed self-exemplars,'' in \emph{CVPR}, 2015, pp. 5197--5206.

\bibitem{hyun2020varsr}
S.~Hyun and J.-P. Heo, ``Varsr: Variational super-resolution network for very
  low resolution images.''\hskip 1em plus 0.5em minus 0.4em\relax Springer,
  2020.

\bibitem{jing2020self}
L.~Jing and Y.~Tian, ``Self-supervised visual feature learning with deep neural
  networks: A survey,'' \emph{TPAMI}, 2020.

\bibitem{johnson2016perceptual}
J.~Johnson, A.~Alahi, and L.~Fei-Fei, ``Perceptual losses for real-time style
  transfer and super-resolution,'' in \emph{ECCV}.\hskip 1em plus 0.5em minus
  0.4em\relax Springer, 2016, pp. 694--711.

\bibitem{RaGAN}
A.~Jolicoeur-Martineau, ``The relativistic discriminator: a key element missing
  from standard gan,'' 2018.

\bibitem{kim2018learning}
D.~Kim, D.~Cho, D.~Yoo, and I.~S. Kweon, ``Learning image representations by
  completing damaged jigsaw puzzles,'' in \emph{WACV}.\hskip 1em plus 0.5em
  minus 0.4em\relax IEEE, 2018, pp. 793--802.

\bibitem{VDSR}
J.~Kim, J.~Kwon~Lee, and K.~Mu~Lee, ``Accurate image super-resolution using
  very deep convolutional networks,'' in \emph{CVPR}, 2016, Conference
  Proceedings, pp. 1646--1654.

\bibitem{DRCN}
------, ``Deeply-recursive convolutional network for image super-resolution,''
  in \emph{CVPR}, 2016, Conference Proceedings, pp. 1637--1645.

\bibitem{Adam}
D.~P. Kingma and J.~Ba, ``Adam: A method for stochastic optimization,''
  \emph{arXiv preprint arXiv:1412.6980}, 2014.

\bibitem{krizhevsky2012imagenet}
A.~Krizhevsky, I.~Sutskever, and G.~E. Hinton, ``Imagenet classification with
  deep convolutional neural networks,'' \emph{NeurIPS}, vol.~25, pp.
  1097--1105, 2012.

\bibitem{lai2017deep}
W.-S. Lai, J.-B. Huang, N.~Ahuja, and M.-H. Yang, ``Deep laplacian pyramid
  networks for fast and accurate super-resolution,'' in \emph{Proceedings of
  the IEEE conference on computer vision and pattern recognition}, 2017, pp.
  624--632.

\bibitem{larsson2016learning}
G.~Larsson, M.~Maire, and G.~Shakhnarovich, ``Learning representations for
  automatic colorization,'' in \emph{ECCV}.\hskip 1em plus 0.5em minus
  0.4em\relax Springer, 2016, pp. 577--593.

\bibitem{larsson2017colorization}
------, ``Colorization as a proxy task for visual understanding,'' in
  \emph{CVPR}, 2017, pp. 6874--6883.

\bibitem{ledig2017photo}
C.~Ledig, L.~Theis, F.~Husz{\'a}r, J.~Caballero, A.~Cunningham, A.~Acosta,
  A.~Aitken, A.~Tejani, J.~Totz, Z.~Wang \emph{et~al.}, ``Photo-realistic
  single image super-resolution using a generative adversarial network,'' in
  \emph{CVPR}, 2017, pp. 4681--4690.

\bibitem{EDSR}
B.~Lim, S.~Son, H.~Kim, S.~Nah, and K.~Mu~Lee, ``Enhanced deep residual
  networks for single image super-resolution,'' in \emph{CVPRW}, 2017,
  Conference Proceedings, pp. 136--144.

\bibitem{liu2018multi}
P.~Liu, H.~Zhang, K.~Zhang, L.~Lin, and W.~Zuo, ``Multi-level wavelet-cnn for
  image restoration,'' in \emph{Proceedings of the IEEE conference on computer
  vision and pattern recognition workshops}, 2018, pp. 773--782.

\bibitem{luan2017deep}
F.~Luan, S.~Paris, E.~Shechtman, and K.~Bala, ``Deep photo style transfer,'' in
  \emph{CVPR}, 2017, pp. 4990--4998.

\bibitem{lugmayr2020srflow}
A.~Lugmayr, M.~Danelljan, L.~Van~Gool, and R.~Timofte, ``Srflow: Learning the
  super-resolution space with normalizing flow,'' in \emph{ECCV}.\hskip 1em
  plus 0.5em minus 0.4em\relax Springer, 2020, pp. 715--732.

\bibitem{ma2020structure}
C.~Ma, Y.~Rao, Y.~Cheng, C.~Chen, J.~Lu, and J.~Zhou, ``Structure-preserving
  super resolution with gradient guidance,'' in \emph{CVPR}, 2020, pp.
  7769--7778.

\bibitem{maeda2020unpaired}
S.~Maeda, ``Unpaired image super-resolution using pseudo-supervision,'' in
  \emph{CVPR}, 2020, pp. 291--300.

\bibitem{BSD100}
D.~R. Martin, C.~C. Fowlkes, D.~Tal, and J.~Malik, ``A database of human
  segmented natural images and its application to evaluating segmentation
  algorithms and measuring ecological statistics,'' in \emph{ICCV}, 2001, pp.
  416--425.

\bibitem{mei2020image}
Y.~Mei, Y.~Fan, Y.~Zhou, L.~Huang, T.~S. Huang, and H.~Shi, ``Image
  super-resolution with cross-scale non-local attention and exhaustive
  self-exemplars mining,'' in \emph{CVPR}, 2020, pp. 5690--5699.

\bibitem{nair2010rectified}
V.~Nair and G.~E. Hinton, ``Rectified linear units improve restricted boltzmann
  machines,'' in \emph{ICML}, 2010.

\bibitem{niu2020single}
B.~Niu, W.~Wen, W.~Ren, X.~Zhang, L.~Yang, S.~Wang, K.~Zhang, X.~Cao, and
  H.~Shen, ``Single image super-resolution via a holistic attention network,''
  in \emph{ECCV}.\hskip 1em plus 0.5em minus 0.4em\relax Springer, 2020, pp.
  191--207.

\bibitem{noroozi2016unsupervised}
M.~Noroozi and P.~Favaro, ``Unsupervised learning of visual representations by
  solving jigsaw puzzles,'' in \emph{ECCV}.\hskip 1em plus 0.5em minus
  0.4em\relax Springer, 2016, pp. 69--84.

\bibitem{oord2016pixel}
A.~v.~d. Oord, N.~Kalchbrenner, and K.~Kavukcuoglu, ``Pixel recurrent neural
  networks,'' \emph{arXiv preprint arXiv:1601.06759}, 2016.

\bibitem{oord2018representation}
A.~v.~d. Oord, Y.~Li, and O.~Vinyals, ``Representation learning with
  contrastive predictive coding,'' \emph{arXiv preprint arXiv:1807.03748},
  2018.

\bibitem{paszke2019pytorch}
A.~Paszke, S.~Gross, F.~Massa, A.~Lerer, J.~Bradbury, G.~Chanan, T.~Killeen,
  Z.~Lin, N.~Gimelshein, L.~Antiga \emph{et~al.}, ``Pytorch: An imperative
  style, high-performance deep learning library,'' \emph{NeurIPS}, vol.~32, pp.
  8026--8037, 2019.

\bibitem{pathak2016context}
D.~Pathak, P.~Krahenbuhl, J.~Donahue, T.~Darrell, and A.~A. Efros, ``Context
  encoders: Feature learning by inpainting,'' in \emph{CVPR}, 2016, pp.
  2536--2544.

\bibitem{radford2015unsupervised}
A.~Radford, L.~Metz, and S.~Chintala, ``Unsupervised representation learning
  with deep convolutional generative adversarial networks,'' \emph{arXiv
  preprint arXiv:1511.06434}, 2015.

\bibitem{rasti2016convolutional}
P.~Rasti, T.~Uiboupin, S.~Escalera, and G.~Anbarjafari, ``Convolutional neural
  network super resolution for face recognition in surveillance monitoring,''
  in \emph{AMDO}.\hskip 1em plus 0.5em minus 0.4em\relax Springer, 2016, pp.
  175--184.

\bibitem{EnhanceNet}
M.~S. Sajjadi, B.~Scholkopf, and M.~Hirsch, ``Enhancenet: Single image
  super-resolution through automated texture synthesis,'' in \emph{ICCV}, 2017,
  Conference Proceedings, pp. 4491--4500.

\bibitem{shaham2019singan}
T.~R. Shaham, T.~Dekel, and T.~Michaeli, ``Singan: Learning a generative model
  from a single natural image,'' in \emph{ICCV}, 2019, pp. 4570--4580.

\bibitem{shi2016real}
W.~Shi, J.~Caballero, F.~Husz{\'a}r, J.~Totz, A.~P. Aitken, R.~Bishop,
  D.~Rueckert, and Z.~Wang, ``Real-time single image and video super-resolution
  using an efficient sub-pixel convolutional neural network,'' in \emph{CVPR},
  2016, pp. 1874--1883.

\bibitem{simonyan2014very}
K.~Simonyan and A.~Zisserman, ``Very deep convolutional networks for
  large-scale image recognition,'' \emph{arXiv preprint arXiv:1409.1556}, 2014.

\bibitem{soh2019natural}
J.~W. Soh, G.~Y. Park, J.~Jo, and N.~I. Cho, ``Natural and realistic single
  image super-resolution with explicit natural manifold discrimination,'' in
  \emph{CVPR}, 2019, pp. 8122--8131.

\bibitem{sun2010gradient}
J.~Sun, Z.~Xu, and H.-Y. Shum, ``Gradient profile prior and its applications in
  image super-resolution and enhancement,'' \emph{TIP}, vol.~20, no.~6, pp.
  1529--1542, 2010.

\bibitem{tatem2002super}
A.~J. Tatem, H.~G. Lewis, P.~M. Atkinson, and M.~S. Nixon, ``Super-resolution
  land cover pattern prediction using a hopfield neural network,'' \emph{Remote
  Sensing of Environment}, vol.~79, no.~1, pp. 1--14, 2002.

\bibitem{thornton2006sub}
M.~W. Thornton, P.~M. Atkinson, and D.~Holland, ``Sub-pixel mapping of rural
  land cover objects from fine spatial resolution satellite sensor imagery
  using super-resolution pixel-swapping,'' \emph{IJRS}, vol.~27, no.~3, pp.
  473--491, 2006.

\bibitem{tian2019contrastive}
Y.~Tian, D.~Krishnan, and P.~Isola, ``Contrastive multiview coding,''
  \emph{arXiv preprint arXiv:1906.05849}, 2019.

\bibitem{wang2019cfsnet}
W.~Wang, R.~Guo, Y.~Tian, and W.~Yang, ``Cfsnet: Toward a controllable feature
  space for image restoration,'' \emph{arXiv preprint arXiv:1904.00634}, 2019.

\bibitem{SFTGAN}
X.~Wang, K.~Yu, C.~Dong, and C.~Change~Loy, ``Recovering realistic texture in
  image super-resolution by deep spatial feature transform,'' in \emph{CVPR},
  2018, pp. 606--615.

\bibitem{wang2018esrgan}
X.~Wang, K.~Yu, S.~Wu, J.~Gu, Y.~Liu, C.~Dong, Y.~Qiao, and C.~C. Loy,
  ``Esrgan: Enhanced super-resolution generative adversarial networks,'' in
  \emph{ECCVW}.\hskip 1em plus 0.5em minus 0.4em\relax Springer, 2018, pp.
  63--79.

\bibitem{wang2004image}
Z.~Wang, A.~C. Bovik, H.~R. Sheikh, E.~P. Simoncelli \emph{et~al.}, ``Image
  quality assessment: from error visibility to structural similarity,''
  \emph{TIP}, vol.~13, no.~4, pp. 600--612, 2004.

\bibitem{wei2020component}
P.~Wei, Z.~Xie, H.~Lu, Z.~Zhan, Q.~Ye, W.~Zuo, and L.~Lin, ``Component
  divide-and-conquer for real-world image super-resolution,'' in
  \emph{ECCV}.\hskip 1em plus 0.5em minus 0.4em\relax Springer, 2020, pp.
  101--117.

\bibitem{yan2015single}
Q.~Yan, Y.~Xu, X.~Yang, and T.~Q. Nguyen, ``Single image superresolution based
  on gradient profile sharpness,'' \emph{TIP}, vol.~24, no.~10, pp. 3187--3202,
  2015.

\bibitem{yang2017deep}
W.~Yang, J.~Feng, J.~Yang, F.~Zhao, J.~Liu, Z.~Guo, and S.~Yan, ``Deep edge
  guided recurrent residual learning for image super-resolution,'' \emph{TIP},
  vol.~26, no.~12, pp. 5895--5907, 2017.

\bibitem{zeyde2010single}
R.~Zeyde, M.~Elad, and M.~Protter, ``On single image scale-up using
  sparse-representations,'' in \emph{ICCS}.\hskip 1em plus 0.5em minus
  0.4em\relax Springer, 2010, pp. 711--730.

\bibitem{zhang2010super}
L.~Zhang, H.~Zhang, H.~Shen, and P.~Li, ``A super-resolution reconstruction
  algorithm for surveillance images,'' \emph{SP}, vol.~90, no.~3, pp. 848--859,
  2010.

\bibitem{zhang2016colorful}
R.~Zhang, P.~Isola, and A.~A. Efros, ``Colorful image colorization,'' in
  \emph{ECCV}.\hskip 1em plus 0.5em minus 0.4em\relax Springer, 2016, pp.
  649--666.

\bibitem{zhang2018unreasonable}
R.~Zhang, P.~Isola, A.~A. Efros, E.~Shechtman, and O.~Wang, ``The unreasonable
  effectiveness of deep features as a perceptual metric,'' in \emph{CVPR},
  2018, pp. 586--595.

\bibitem{RCAN}
Y.~Zhang, K.~Li, K.~Li, L.~Wang, B.~Zhong, and Y.~Fu, ``Image super-resolution
  using very deep residual channel attention networks,'' in \emph{ECCV}, 2018,
  Conference Proceedings, pp. 286--301.

\bibitem{RDN}
Y.~Zhang, Y.~Tian, Y.~Kong, B.~Zhong, and Y.~Fu, ``Residual dense network for
  image super-resolution,'' in \emph{CVPR}, 2018, Conference Proceedings, pp.
  2472--2481.

\bibitem{zhu2015modeling}
Y.~Zhu, Y.~Zhang, B.~Bonev, and A.~L. Yuille, ``Modeling deformable gradient
  compositions for single-image super-resolution,'' in \emph{CVPR}, 2015, pp.
  5417--5425.

\end{thebibliography}
}


\end{document}